\documentclass[letterpaper]{article} 
\usepackage{aaai2026}  
\usepackage{times}  
\usepackage{helvet}  
\usepackage{courier}  
\usepackage[hyphens]{url}  
\usepackage{graphicx} 
\usepackage{natbib}  
\usepackage{caption} 
\usepackage{algorithm}
\usepackage{algorithmic}
\usepackage{xcolor}
\usepackage{newfloat}
\usepackage{listings}
\DeclareCaptionStyle{ruled}{labelfont=normalfont,labelsep=colon,strut=off} 
\floatstyle{ruled}
\newfloat{listing}{tb}{lst}{}
\floatname{listing}{Listing}
\usepackage{hyperref}
\usepackage{tabularx,colortbl,xcolor}
\hypersetup{
    colorlinks=true,
    linkcolor=blue,
    filecolor=blue,      
    urlcolor=blue,
    citecolor=cyan,
}
\usepackage{subcaption}
\usepackage{amssymb}
\usepackage{booktabs}
\usepackage{cite}
\usepackage{latexsym}
\usepackage{array}
\usepackage{epstopdf}
\usepackage{makecell}
\newcolumntype{L}[1]{>{\raggedright\let\newline\\\arraybackslash\hspace{0pt}}m{#1}}
\newcolumntype{C}[1]{>{\centering\let\newline\\\arraybackslash\hspace{0pt}}m{#1}}
\newcolumntype{R}[1]{>{\raggedleft\let\newline\\\arraybackslash\hspace{0pt}}m{#1}}
\usepackage{multirow}
\usepackage{bbding}
\usepackage{pifont}
\usepackage{amsmath}
\usepackage{arydshln}
\usepackage{cleveref}
\usepackage{enumitem}
\crefname{section}{§}{§§}
\Crefname{section}{§}{§§}
\usepackage{bm}

\newcommand{\our}{\textsc{SparK}} 
\newcommand{\think}{\textsc{ThinK}}

\newcommand\hc{ \rowcolor{teal!12}}
\newcommand\ha{ \rowcolor{teal!17}}
\newcommand{\hx}{\rowcolor{teal!22}}
\newcommand{\hd}{\rowcolor{teal!37}}
\newcommand{\hb}{\rowcolor{teal!42}}
\newcommand{\he}{\rowcolor{teal!47}}

\definecolor{amdpurple}{RGB}{75,46,131}
\definecolor{ucaspink}{HTML}{F0529C}
\definecolor{casiaorange}{HTML}{F58025}

\title{\our: Query-Aware Unstructured Sparsity with \\Recoverable KV Cache Channel Pruning}
\author{%
    Huanxuan Liao\textsuperscript{\textcolor{casiaorange}{$\tau$}, \textcolor{ucaspink}{$\mu$}}, 
    Yixing Xu\textsuperscript{\textcolor{amdpurple}{$\alpha$}}, 
    Shizhu He\textsuperscript{\textcolor{casiaorange}{$\tau$}, \textcolor{ucaspink}{$\mu$}}$\thanks{Corresponding author.}$, 
    Guanchen Li\textsuperscript{\textcolor{amdpurple}{$\alpha$}}, 
    Xuanwu Yin\textsuperscript{\textcolor{amdpurple}{$\alpha$}}, 
    Dong Li\textsuperscript{\textcolor{amdpurple}{$\alpha$}}, \\
    Emad Barsoum\textsuperscript{\textcolor{amdpurple}{$\alpha$}}, 
    Jun Zhao\textsuperscript{\textcolor{casiaorange}{$\tau$}, \textcolor{ucaspink}{$\mu$}}, 
    Kang Liu\textsuperscript{\textcolor{casiaorange}{$\tau$}, \textcolor{ucaspink}{$\mu$}}
}
\affiliations {
    \textsuperscript{\textcolor{amdpurple}{$\alpha$}}Advanced Micro Devices (china) Co., Ltd. \\
    \textsuperscript{\textcolor{casiaorange}{$\tau$}}Institute of Automation, Chinese Academy of Sciences \\ \textsuperscript{\textcolor{ucaspink}{$\mu$}}University of Chinese Academy of Sciences \\
    {\texttt{liaohuanxuan2023@ia.ac.cn}} \\%
}

\begin{document}

\maketitle

\begin{abstract}
Long-context inference in large language models (LLMs) is increasingly constrained by the KV cache bottleneck: memory usage grows linearly with sequence length, while attention computation scales quadratically. Existing approaches address this issue by compressing the KV cache along the \textit{temporal} axis through strategies such as token eviction or merging to reduce memory and computational overhead. However, these methods often neglect fine-grained importance variations across feature dimensions (i.e., the \textit{channel axis}), thereby limiting their ability to effectively balance efficiency and model accuracy.
In reality, we observe that channel saliency varies dramatically across both queries and positions: certain feature channels carry near-zero information for a given query, while others spike in relevance. 
To address this oversight, we propose \our, a training-free plug-and-play method that applies unstructured sparsity by pruning KV at the channel level, while dynamically restoring the pruned entries during attention score computation.
Notably, our approach is orthogonal to existing KV compression and quantization techniques, making it compatible for integration with them to achieve further acceleration. By reducing channel-level redundancy, \our{} enables processing of longer sequences within the same memory budget. For sequences of equal length, \our{} not only preserves or improves model accuracy but also reduces KV cache storage by over $30\%$ compared to eviction-based methods. Furthermore, even in an aggressive pruning ratio of $80\%$, \our{} maintains performance with less degradation than $5\%$ compared to the based eviction method, demonstrating its robustness and effectiveness. 
Our code will be available at \href{https://github.com/AMD-AIG-AIMA/AMD-Spark}{Spark}.


\end{abstract}


\section{Introduction}

Large language models (LLMs) are increasingly deployed in diverse and complex tasks requiring extended (even infinite) contextual understanding \citep{liu2025comprehensive, tan2025dynamic}, such as book summarization \citep{fables}, instruction following \citep{liao2024instance} and code or math reasoning \citep{liao2025neural}. To support these applications, recent models like GPT-4 \citep{gpt4}, Gemini-2.5 \citep{Gemini}, and Qwen-3 \citep{qwen3} have scaled to 100K+ token contexts. However, handling such long sequences poses serious challenges in memory and latency due to the growing Key-Value (KV) cache in inference \citep{quest}. For example, storing the KV cache for 100K tokens in LLaMA3.1-8B \citep{llama3} exceeds $50GB$, surpassing the model size itself \citep{cache, skintern}. For a hidden size of $128$, matrix multiplication latency increases from $2ms$ at $1K$ tokens to $764ms$ at $16$K, nearly $380\times$ slower.
Consequently, KV cache has become a critical bottleneck, restricting the scalability and deployment of LLMs in long-context scenarios \citep{fu2024challenges}.

\begin{figure}[t]
\centerline{\includegraphics[width=0.5\textwidth]{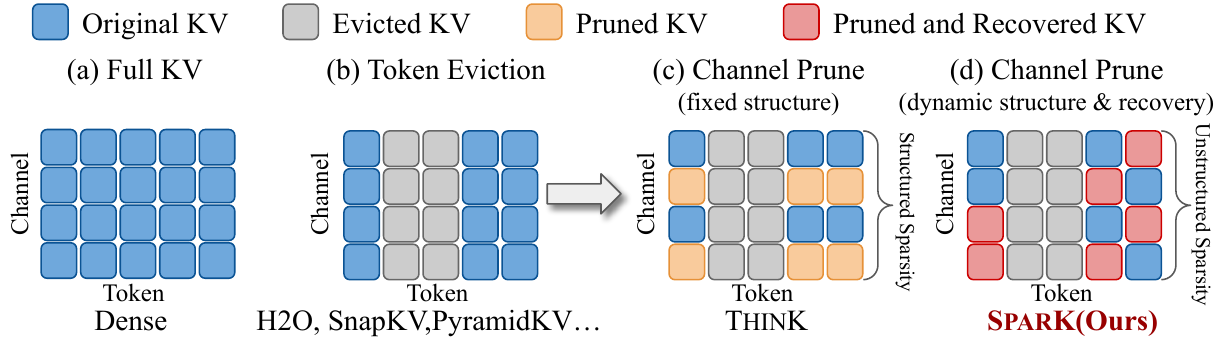}}
\caption{
\textbf{Illustrative comparisons} among (a) full KV cache, (b) eviction-based KV compression, (c) structured channel pruning-based KV reduction, and (d) our proposed \our{}, which employs unstructured channel pruning with subsequent recovery during attention score computation.
}
\label{intro}
\vspace{-0.55cm}
\end{figure}

Specifically, the total KV cache size is determined by the batch size $B$, sequence length $S$, number of layers $L$, attention heads $N$, and the head dimension $D$. Prior efforts on KV cache compression have primarily targeted the following aspects:
1) \textbf{Temporal axis ($S$)}: by evicting \citep{modeltell, h2o} or merging \citep{d2o, model} unimportant tokens using attention scores or redundancy heuristics \citep{rkv}.
2) \textbf{Spatial axis ($L$, $N$)}: by sharing KV across similar layers \citep{reducing, layer} or pruning attention heads with limited contribution to long-range dependencies \citep{xiaoduoattention}.
3) \textbf{Channel axis ($D$)}: by applying low-rank decomposition \citep{deepseek, shadowkv} or structured pruning \citep{think}.
4) \textbf{Quantization}: by applying low-bit precision storage \citep{kvquant, MIR-2023-10-223}.

However, these approaches predominantly adopt structured channel sparsity, applying uniform pruning strategies that either discard or retain entire channels, or enforce fixed pruning masks across all tokens \citep{keep}. Such methods rest on the assumption that channel importance remains consistent throughout the input sequence, which overlooks the dynamic and token-specific nature of attention in LLMs. Moreover, by applying identical pruning masks to both keys and queries, these methods fail to account for the asymmetric roles and token-wise variability in channel saliency, ultimately limiting the flexibility of the dot-product attention mechanism.
Instead of directly discarding unimportant channels, we argue that \textit{replacing unimportant channel entries with approximate or low-magnitude entries} can mitigate attention score distortion and maintain performance even under an aggressive pruning ratio.

In this paper, we propose \our, a method that introduces fine-grained query-aware unstructured sparsity to the KV cache while guaranteeing the recoverability of pruned channel entries. We reformulate channel pruning as a \textbf{critical channel set} selection problem aimed at maximizing aggregate saliency across selected channels. To this end, we introduce a lightweight metric to quantify the per-token, per-channel importance and adopt a greedy algorithm to solve the resulting optimization problem efficiently \citep{forest}. 
To mitigate information loss under high pruning ratios, we further introduce a recovery mechanism that approximates the contributions of pruned channels through a recovery function $\boldsymbol{\mathcal{F}}$ during attention computation. This approximation ensures effective information retention without incurring additional memory cost. 
We additionally explore value cache pruning via a simple norm-based heuristic, showing promising results and paving the way for future refinement. 
Furthermore, we propose two ratio-free variants: group-based (\our-g) and top-p pruning (\our-p), demonstrating the flexibility and generality of \our.

Extensive experimental evaluations demonstrate the effectiveness of \our{} across a wide range of scenarios, benchmarks \citep{longbench, ruler}, and LLMs \citep{llama3, qwen3}.
Importantly, \our{} is compatible with prior methods that optimize $S$, $L$ and $N$. When integrated with token eviction strategies, \our{} not only preserves computational efficiency and achieves comparable or superior accuracy but also reduces KV cache storage by over 30\%. Remarkably, even at high channel pruning ratio ($\geq$ 70\%) while maintaining the same sequence length via token eviction methods such as SnapKV \citep{snapkv} or PyramidKV \citep{pyramidinfer}, \our{} maintains performance degradation within 5\% compared to the based method, significantly outperforming \think, which incurs a 47.6\% accuracy loss under similar settings.
Our main contributions are listed as follows:
\begin{itemize}[leftmargin=*]
    \item We propose \our, a novel training-free plug-and-play KV cache compression approach that introduces unstructured fine-grained sparsity along the channel dimension. 
    We reformulate the pruning task as a critical channel set selection problem that aims to maximize the saliency contribution of preserved channels.
    \item We introduce an on-the-fly recovery mechanism that approximates the contribution of pruned channels during attention score computation using a lightweight function $\boldsymbol{\mathcal{F}}$ to mitigate information loss with little increasing memory footprint or computational overhead.
    \item Extensive experiments show that our method consistently achieves remarkable effectiveness in various benchmarks and LLM. Notably, even when pruning 80\% of the channels at the same sequence length, the performance degradation remains within 5\%.
\end{itemize}

\section{Related Work}

Existing KV cache compression methods can be broadly categorized into three categories based on dimensions: \textbf{temporal-axis}, \textbf{spatial-axis}, and \textbf{channel-axis} methods.

\begin{figure*}[t]  
  \centering

  \begin{subfigure}[t]{0.33\textwidth}
    \includegraphics[width=\linewidth, trim=0 -30 0 10mm, clip]{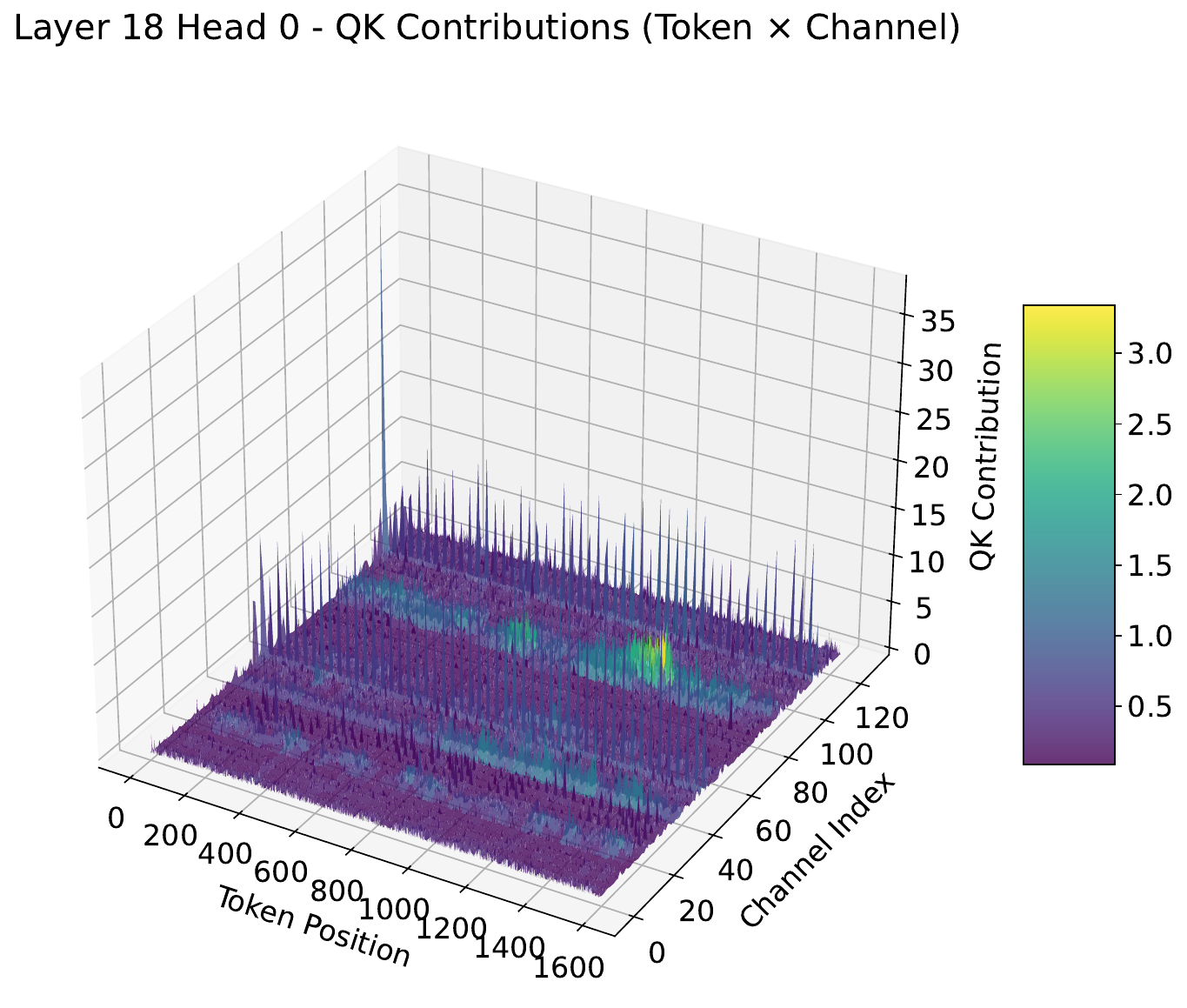}
    \caption{\textbf{3D surface visualization} of score contributions, highlighting token-wise unstructured sparsity.}
  \end{subfigure}
  \hfill
  \begin{subfigure}[t]{0.32\textwidth}
    \includegraphics[width=\linewidth, trim=190mm -20 0 9mm, clip]{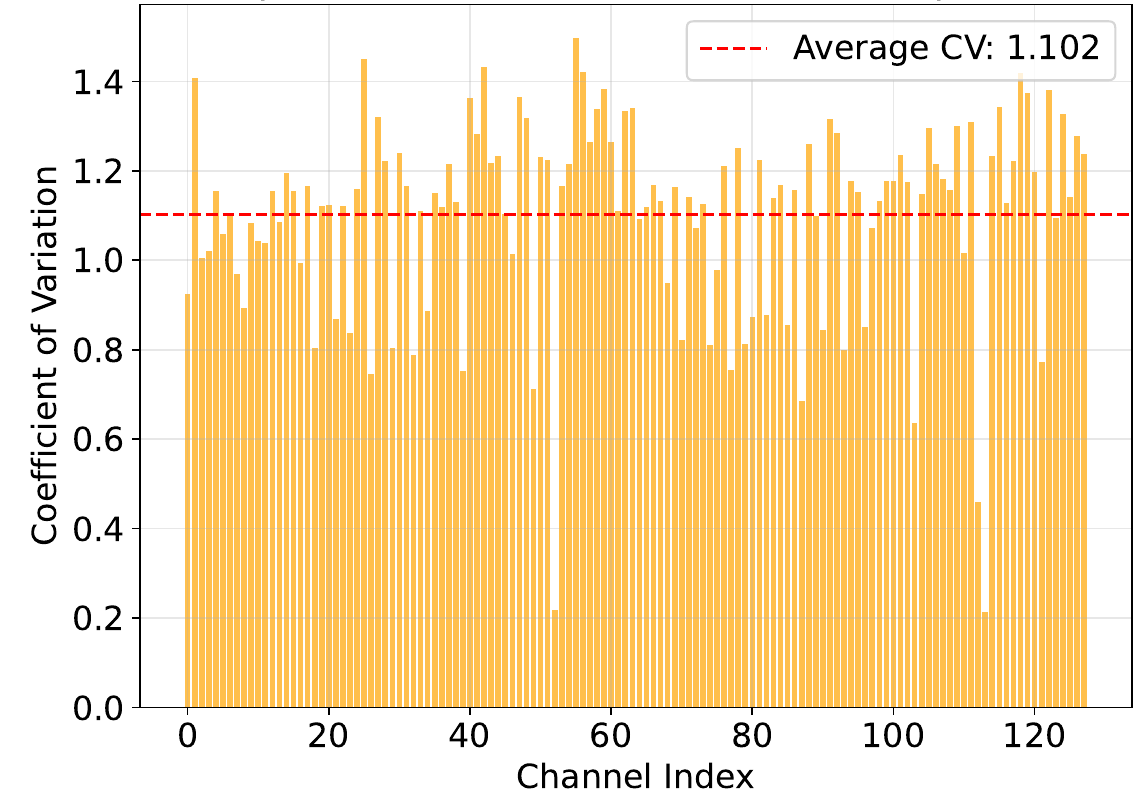}
    \caption{\textbf{Coefficient of Variation} (CV) distribution across channels, suggesting dynamic channel modulation for different contexts.}
  \end{subfigure}
  \hfill
  \begin{subfigure}[t]{0.34\textwidth}
    \includegraphics[width=\linewidth, trim=0 0 0 0, clip]{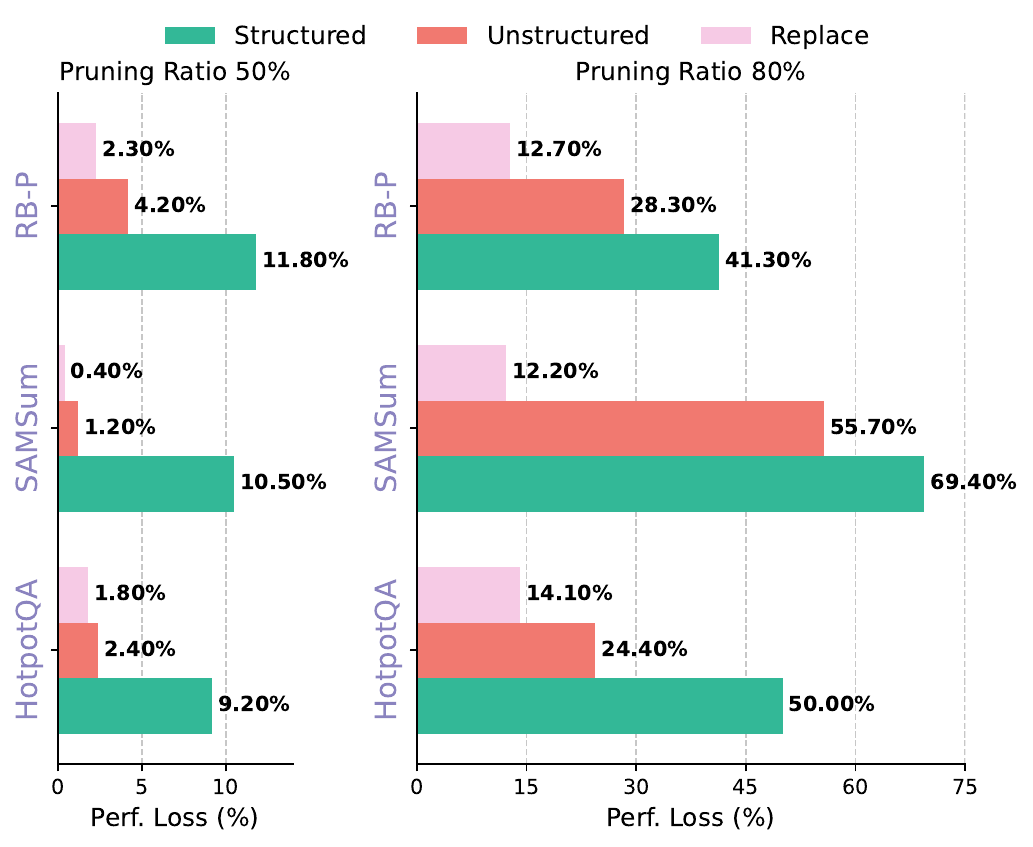}
    \caption{\textbf{Metric comparison}: performance loss caused by the method, highlighting the importance of unstructured pruning and recovery.}
  \end{subfigure}
  \hfill
  \caption{Rethinking the salience of key channels using LLaMA3.1-8B-Instruct \citep{llama3} on Longbench \citep{longbench}. All visualizations are derived from the 18th attention layer and the 0th attention head.}
  \label{fig:motivation}
\end{figure*}

\noindent \textbf{Temporal-Axis Optimization} reduces the sequence length $S$ to alleviate the linear memory growth in long-context inference \citep{liao2025beyond, updp}. \textit{Token eviction} methods selectively remove low-contributing tokens based on attention scores \citep{snapkv, modeltell, pyramidinfer, liao2025awakening} or redundancy heuristics \citep{rkv}. \textit{Token merging} techniques compress inputs by merging semantically similar tokens \citep{dynamic, d2o, model} or aggregating discarded ones \citep{squeezed, cam}. Paged KV cache architectures, such as vLLM \citep{vllm}, further enhance scalability via memory paging.

\noindent \textbf{Spatial-Axis Optimization} reduces redundancy by shrinking the number of layers $L$ or heads $N$. Cross-layer sharing \citep{you, kvsharer} enables KV reuse across layers, while MQA \citep{mqa} and GQA \citep{gqa} share KV pairs across heads. Head optimization aims to prune attention heads that are less sensitive to long-range dependencies \citep{not, razorattention,zhu2024dip}, and DuoAttention \citep{xiaoduoattention} specializes heads for retrieval or streaming to enhance efficiency.

\noindent \textbf{Channel-Axis Optimization} targets the channel dimension $D$ to reduce KV cache memory. Low-rank methods \citep{shadowkv, sampleattention} decompose KV matrices into compact representations, while MLA \citep{deepseek} learns latent heads to compress channels, requiring retraining. Closest to our work, \think{} \citep{think} performs query-guided structured pruning, but its structured strategy significantly degrades performance under high pruning ratios. In contrast, we propose unstructured, dynamic pruning with on-the-fly recovery, enabling adaptive removal and restoration of KV entries during computation.

\section{Preliminaries}

LLM inference comprises two stages \citep{liu2025comprehensive}: \textbf{prefill} and \textbf{decode}.
During prefill, the entire input sequence is processed in parallel to generate the first output token. Given a prompt embedding $\mathbf{X} \in \mathbb{R}^{S \times H}$, where $S$ is the sequence length and $H$ is the hidden dimension, the key and value matrices for each attention head $i \in [1, N]$ are computed as:
\begin{equation}
\boldsymbol{\mathcal{K}}_i = \mathbf{X} \mathbf{W}_k^{i}, \quad \boldsymbol{\mathcal{V}}_i = \mathbf{X} \mathbf{W}_v^{i},
\end{equation}
where $\mathbf{W}_k^{i}, \mathbf{W}_v^{i} \in \mathbb{R}^{H \times D}$ are the projection matrices for the $i$-th head, and $D$ is the dimensionality of each head. The resulting keys and values are stored in the KV cache.
During decode, each newly generated token embedding $\mathbf{x} \in \mathbb{R}^{1 \times H}$ is projected to obtain the corresponding query, key, and value vectors and appended to the existing KV cache:
\begin{equation}
\mathbf{q}_i = \mathbf{x} \mathbf{W}_q^{i}, \quad \mathbf{k}_i = \mathbf{x} \mathbf{W}_k^{i}, \quad \mathbf{v}_i = \mathbf{x} \mathbf{W}_v^{i}.
\end{equation}
\begin{equation}
\boldsymbol{\mathcal{K}}_i \leftarrow \text{Cat}[\boldsymbol{\mathcal{K}}_i, \mathbf{k}_i], \quad \boldsymbol{\mathcal{V}}_i \leftarrow \text{Cat}[\boldsymbol{\mathcal{V}}_i, \mathbf{v}_i].
\end{equation}
The attention output for each head is then computed as:
\begin{equation}
\mathbf{a}_i = \text{Softmax} \left( \frac{\mathbf{q}_i \boldsymbol{\mathcal{K}}_i^\top}{\sqrt{D}} \right), \quad \mathbf{o}_i = \mathbf{a}_i \boldsymbol{\mathcal{V}}_i.
\end{equation}
Finally, the outputs $\mathbf{o}_i$ from all heads are concatenated and passed to the feed-forward network (FFN).
In scenarios involving extended contexts or large batch processing, the primary bottlenecks in memory consumption and computational speed stem from the KV size. 
While existing approaches primarily focus on reducing KV size through temporal ($S$) or spatial ($L,N$) optimization, we draw inspiration from \think{} \citep{think} and propose optimizing the KV cache from channel $D$, thereby offering a complementary and orthogonal direction for KV compression.


\section{Methodology}
\label{main:Methodology}

In this section, we begin with an experimental analysis and motivation for \our{} in Sec.\ref{main:motivation}, followed by problem formulation and analysis in Sec.\ref{main:problem}. We further introduce the proposed \our{} in Sec.\ref{main:method}. 

\subsection{Motivations and Observations}
\label{main:motivation}

To understand the role of individual key channels, we conduct an empirical analysis\footnote{More analysis and metric details refer to the Appendix \ref{app:ex_observation}.} of the QK dot-product scores.
As shown in Figure \ref{fig:motivation}, we observe \textbf{unstructured, token-dependent channel} importance patterns that vary significantly across different tokens, which motivates the need for adaptive pruning strategies that can dynamically select different channels for different tokens, rather than applying uniform pruning across the entire sequence \citep{specache}. 

\noindent \textbf{Observation 1: Token-wise Unstructured Channel Sparsity.}
Empirical analysis reveals that attention heads exhibit highly unstructured channel-wise sparsity, varying significantly across tokens. As shown in Figure \ref{fig:motivation}(a), the 3D surface visualization highlights token-dependent activation patterns, where different tokens rely on distinct subsets of channels. This contradicts structured pruning assumptions where importance is globally consistent.
To quantify this variability, we compute the coefficient of variation (CV) across tokens for each channel, as illustrated in Figure~\ref{fig:motivation}(b). The average CV exceeds 1.1, indicating that token-wise fluctuations dominate. This suggests that channel importance is highly context sensitive and cannot be accurately captured through a static and structured sparsity.
Figure~\ref{fig:motivation}(c) further demonstrates that unstructured pruning, which respects token-level heterogeneity, substantially outperforms structured pruning. At 50\% pruning, unstructured pruning leads to only 1.2\% performance drop (vs. 4.2\% for structured); at 80\% pruning, it maintains a 27.4\% gap (28.3\% vs. 55.7\%). These results affirm the necessity of unstructured sparsity.

\noindent \textbf{Observation 2: Retaining Dimensional Structure Mitigates Pruning Impact.}
Figure~\ref{fig:motivation}(c) also shows that replacing pruned channel entries with minimal constant values (e.g., 0.01) during attention score computation rather than zeroing or omitting them yields substantial performance gains. This lightweight strategy preserves the structural integrity of the attention mechanism while avoiding pruning queries.
Under 80\% pruning, this approach significantly narrows the performance gap. On SAMSum, it reduces performance degradation from 55.7\% to 12.2\%; on HotpotQA, from 69.4\% to 41.3\%; and on RB-P, from 50.0\% to 24.4\%. On average, the substitution of entries reduces the loss of accuracy by 32. 4\% compared to removal. These results highlight that even a coarse query-agnostic constant of pruning channel can play a pivotal role in maintaining performance.

\subsection{Problem Formulation and Analysis}
\label{main:problem}

Let $\mathcal{C}_{i,t} = \{ \boldsymbol{c}_1, \boldsymbol{c}_2, \ldots, \boldsymbol{c}_D \}$ denote the original channel set for each head $i$ and token $t$, where $D$ is the head dimension.
We aim to select a subset $\hat{\mathcal{C}}_{i,t} \subseteq \mathcal{C}_{i,t}$ of $T$ channels ($T \ll D$) that retain the most salient attention contributions, thereby enhancing inference efficiency while minimizing performance degradation.
To formalize this, we introduce a binary mask $\mathcal{S}_{i,t} = \{\boldsymbol{z}_{i,t}^1, \ldots, \boldsymbol{z}_{i,t}^D\} \in \{0,1\}^{D}$ with $\boldsymbol{z}_{i,t}^j \in \{0, 1\}$ indicating whether channel $j$ is retained ($\boldsymbol{z}_{i,t}^j = 1$) or pruned ($\boldsymbol{z}_{i,t}^j = 0$).
Our primary goal is to \emph{minimize} the discrepancy ($\mathcal{E}$) in attention weights after pruning:
\begin{equation}
\label{eq:err}
\min_{\mathcal{S}_{i,t}}
\mathcal{E}(\mathcal{S}_{i,t}) =
\left\| \mathbf{q}_{i,t} \mathbf{k}_{i,t}^\top - (\mathbf{q}_{i,t} \mathcal{S}_{i,t}) (\mathbf{k}_{i,t} \mathcal{S}_{i,t})^\top \right\|_F,
\end{equation}
\noindent
where $\|\|_F$ denotes the Frobenius norm for vectors.
Solving this combinatorial problem exactly is intractable as it corresponds to a cardinality-constrained low-rank approximation. To derive an approximate solution, we expand the squared Frobenius norm of $\mathcal{E}$ for each token $t$:
\begin{equation}
\begin{aligned}
\mathcal{E}(\mathcal{S}_{i,t})^2
\;=\!
\sum_{j=1}^D\|\mathbf{q}_{i,t}^j\|_2^2\|\mathbf{k}_{i,t}^j\|_2^2(1-\boldsymbol{z}_{i,t}^j)^2
+ \\
2\sum_{\substack{j,r=1\\j<r}}^D
\langle\mathbf{q}_{i,t}^j,\mathbf{q}_{i,t}^r\rangle\,
\langle\mathbf{k}_{i,t}^j,\mathbf{k}_{i,t}^r\rangle(1-\boldsymbol{z}_{i,t}^j\boldsymbol{z}_{i,t}^r),
\label{eq:expansion}
\end{aligned}
\end{equation}
where $\mathbf{q}_{i,t}^j$ and $\mathbf{k}_{i,t}^j$ are the $j$-th dimensions of $\mathbf{q}_{i,t}$ and $\mathbf{k}_{i,t}$ respectively (similarly for $r$).
The first term measures individual contributions of each pruned channel, while the second reflects inter-channel redundancy.
In practice, we observe that different channels are nearly uncorrelated (i.e., $\langle\mathbf{k}_{i,t}^j,\mathbf{k}_{i,t}^r\rangle \approx 0$ for $j \ne r$), allowing us to drop the second term.
Thus, minimizing $\mathcal{E}(\mathcal{S}_{i,t})$ is well-approximated by \emph{minimizing} the sum of the norms of pruned channel contributions for each token, which is equivalent to \emph{maximizing} retained channel scores while the number of selected channels for each token is fixed: $\sum_{j}^D \boldsymbol{z}_{i,t}^j = T$.
We introduce a proxy saliency score $\boldsymbol{w}_{i,t}^j = \|\mathbf{q}_{i,t}^j\|_2\|\mathbf{k}_{i,t}^j\|_2$, which upper bounds the contribution of channel $j$ at token $t$ to the Frobenius norm. The optimization problem is reformulated as follows:
\begin{equation}
\label{eq:cal}
\max_{\boldsymbol{z}_{i,t}^j} \sum_{j=1}^{D} \boldsymbol{w}_{i,t}^j \boldsymbol{z}_{i,t}^j \quad \text{s.t.} \quad \sum_{j=1}^D \boldsymbol{z}_{i,t}^j = T, \quad \forall t,
\end{equation}
Since the objective is linear and additive in $\boldsymbol{z}_j$, the optimal solution is simply to select the $T$ channels with the highest saliency score $\boldsymbol{w}_j$, which can be efficiently solved using a greedy algorithm: $\hat{\mathcal{C}}_{i,t} = \text{Top}_T(\boldsymbol{w}_{i,t}^1, \ldots, \boldsymbol{w}_{i,t}^D)$. Given the pruning ratio $\lambda$, we only keep the $T = \lfloor(1 - \lambda)D\rfloor$ most important channels among $D$ channels of each head.

\begin{figure}[t]
\centerline{\includegraphics[width=0.5\textwidth]{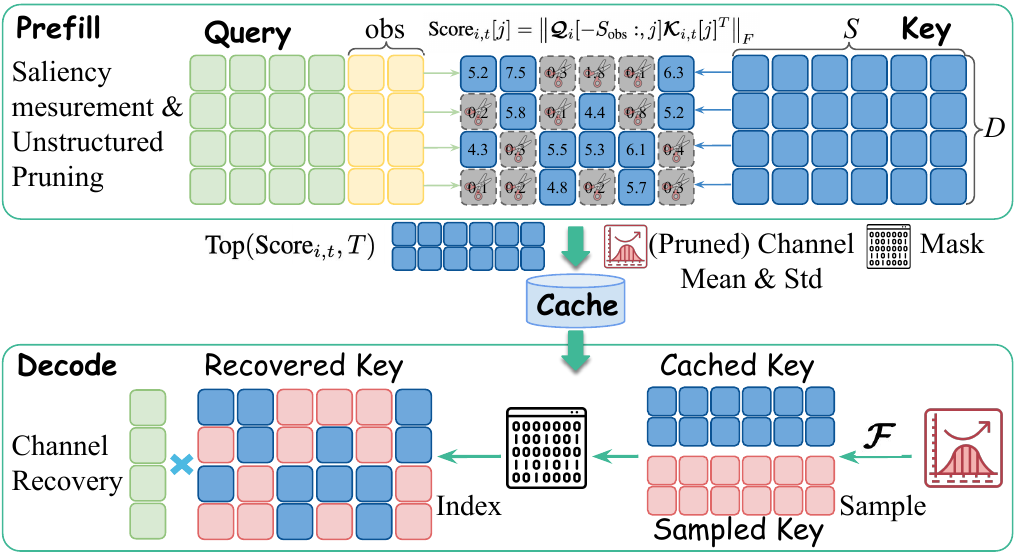}}
\caption{\textbf{An illustration of \our.}
\our{} computes channel-wise saliency scores and applies unstructured pruning during prefill.
During decoding, \our{} leverage $\boldsymbol{\mathcal{F}}$ and sampling from the cached distribution to reconstruct the pruned channels and then perform standard full attention. 
}
\label{model}
\end{figure}

\subsection{\our}
\label{main:method}

Building on above analysis, we redefine the channel pruning problem as (Eq.~\ref{eq:expansion}). Since this study focuses on efficiency in long-context inference, we employ a heuristic algorithm with relatively low computational complexity to obtain an approximate solution.
To this end, we introduce an unstructured channel pruning method (Figure \ref{model}), which selects an appropriate channel subset while ensuring that it satisfies the constraint. Our approach is training-free, plug-and-play, and model-agnostic, which makes it applicable to any LLM.

As illustrated in Figure \ref{model}, the proposed method consists of two primary phases: 1) unstructured channel pruning based on saliency measurement during \textbf{prefill}, and 2) channel recovery using stored distribution patterns during \textbf{decode}. Following previous work \citep{snapkv, think}, to reduce the computation cost, we only use the last observation window to calculate the saliency score.
Specifically, we approximate the attention interaction by replacing per-token query vectors with the mean query vector computed over a local observation window. 
Specifically, for an observation window of size \( W \), the mean query vector \(\overline{\mathbf{q}}_i\) for the head \( i \) is calculated as the average of the query vectors \(\mathbf{q}_{i,t}\) over the window: $
\overline{\mathbf{q}}_i^j = \frac{1}{W} \sum_{t=t_0}^{t_0 + W - 1} \mathbf{q}_{i,t}^j,$
where \( t_0 \) is the starting token index of the window.

\begin{table*}[t!]
\centering 
\scriptsize
\resizebox{\textwidth}{!}{\begin{tabular}
{C{1em}l@{\hspace{0.05ex}}C{3.8em}@{\hspace{0.05ex}}C{3.8em}@{\hspace{0.05ex}}C{3.8em}@{\hspace{0.05ex}}c@{\hspace{0.05ex}}c@{\hspace{0.05ex}}C{3.8em}@{\hspace{0.05ex}}c@{\hspace{0.05ex}}C{3.8em}@{\hspace{0.05ex}}c@{\hspace{0.05ex}}C{3.8em}@{\hspace{0.05ex}}c@{\hspace{0.05ex}}c@{\hspace{0.05ex}}c@{\hspace{0.6ex}}C{3.8em}@{\hspace{0.6ex}}C{3.8em}@{\hspace{0.6ex}}C{3.8em}@{\hspace{0.6ex}}c}
    \toprule
& \multirow{4}{*}{\textbf{Method}}& \multicolumn{3}{c}{\textbf{Single-Document QA}} & \multicolumn{3}{c}{\textbf{Multi-Document QA}}& \multicolumn{3}{c}{\textbf{Summarization}}& \multicolumn{3}{c}{\textbf{Few-shot Learning}}& \multicolumn{2}{c}{\textbf{Synthetic}} & \multicolumn{2}{c}{\textbf{Code}}&\multirow{4}{*}{\textbf{Avg.}} \\
& & \rotatebox[origin=c]{30}{\bf NrtvQA} & \rotatebox[origin=c]{30}{\bf Qasper} & \rotatebox[origin=c]{30}{\bf MF-en} & \rotatebox[origin=c]{30}{\bf HotpotQA} & \rotatebox[origin=c]{30}{\bf 2WikiMQA} & \rotatebox[origin=c]{30}{\bf Musique} & \rotatebox[origin=c]{30}{\bf GovReport} & \rotatebox[origin=c]{30}{\bf QMSum} & \rotatebox[origin=c]{30}{\bf MultiNews} & \rotatebox[origin=c]{30}{\bf TREC} & \rotatebox[origin=c]{30}{\bf TriviaQA} & \rotatebox[origin=c]{30}{\bf SAMSum} & \rotatebox[origin=c]{30}{\bf PCount} & \rotatebox[origin=c]{30}{~\bf PRe~~} & \rotatebox[origin=c]{30}{~\bf Lcc~~} & \rotatebox[origin=c]{30}{~\bf RB-P~} \\
\cmidrule{1-19}

 - & Vanilla & 22.48 & 44.72 & 46.23 & 48.49 & 44.71 & 24.43 & 30.7 & 22.8 & 27.28 & 72.0 & 88.35 & 42.28 & 6.5 & 72.0 & 63.61 & 51.67 & 44.27 \\
\cmidrule{1-19}
 \multirow{13}*{\raisebox{10.0em}{\rotatebox{90}{KV-size 128}}}
 &   StreamingLLM  & 13.64 & 18.03 & 17.79 & 31.36 & 27.46 & 8.67 & 17.31 & 18.99 & 17.87 & 31.0 & 31.21 & 35.71 & 1.5 & 67.5 & 56.63 & 55.16 & 28.11 \\
 &   ExpectedAttention  & 17.32 & 24.08 & 23.87 & 38.76 & 26.43 & 12.55 & 22.26 & 20.81 & 23.57 & 20.5 & 77.22 & 36.59 & 5.5 & 62.5 & 52.78 & 46.45 & 31.95 \\
 &  TOVA & 17.09 & 23.35 & 37.88 & 43.32 & 28.68 & 15.85 & 19.87 & 20.54 & 18.51 & 26.5 & 85.18 & 39.15 & 4.0 & 60.5 & 59.98 & 57.17 & 34.85 \\
  \cdashline{2-19}
 &   SnapKV &  15.29 & 20.03 & 29.2 & 39.92 & 28.26 & 15.06 & 17.74 & 19.27 & 18.05 & 21.0 & 68.64 & 36.64 & 6.0 & 66.0 & 57.86 & 59.08 & 32.38  \\
 &   +\think~(0.5)    & 13.6 & 19.2 & 31.78 & 36.24 & 25.05 & 11.92 & 16.85 & 19.17 & 16.4 & 2.0 & 50.73 & 32.8 & 6.0 & 65.0 & 52.29 & 52.11 & 28.20 \\
 &   +\think~(0.8)    & 7.6 & 7.03 & 17.45 & 19.98 & 9.8 & 6.9 & 14.37 & 14.15 & 12.5 & 0.0 & 21.36 & 11.22 & 1.02 & 63.0 & 30.42 & 34.69 & 16.97 \\
 \hc &   +\textbf{\our}~(0.5)    & 13.52 & 20.19 & 29.28 & 38.77 & 26.33 & 14.44 & 17.66 & 19.12 & 17.98 & 21.0 & 68.95 & 36.66 & 5.5 & 65.5 & 58.49 & 59.18 & 32.04 \\
 \hd &   +\textbf{\our}~(0.8)    & 13.82 & 20.28 & 28.63 & 40.84 & 26.75 & 14.25 & 17.29 & 19.06 & 17.23 & 22.0 & 57.2 & 35.41 & 7.0 & 64.0 & 57.2 & 57.61 & 31.16 \\
 \cdashline{2-19}
 &   PyramidKV        & 21.79 & 44.6 & 45.96 & 48.33 & 43.63 & 25.82 & 30.42 & 22.45 & 27.05 & 72.0 & 88.69 & 41.59 & 6.0 & 71.5 & 62.21 & 48.72 & 43.80 \\
 &   +\think~(0.5)    & 22.48 & 40.56 & 47.94 & 45.83 & 34.95 & 23.19 & 27.55 & 22.54 & 25.73 & 53.5 & 84.88 & 32.7 & 7.78 & 71.0 & 53.9 & 51.54 & 40.38 \\
 &   +\think~(0.8)    &  6.37 & 5.53 & 13.73 & 12.53 & 5.47 & 3.16 & 16.97 & 14.21 & 17.02 & 0.0 & 23.03 & 7.54 & 1.73 & 13.0 & 29.67 & 27.51 & 12.34\\
 \hc &   +\textbf{\our}~(0.5)    &  22.66 & 43.95 & 45.82 & 48.33 & 43.85 & 24.85 & 30.16 & 22.76 & 26.84 & 70.0 & 88.34 & 41.4 & 6.5 & 71.5 & 62.83 & 51.15 & 43.81 \\
 \hd &   +\textbf{\our}~(0.8)    & 22.44 & 44.2 & 44.62 & 46.29 & 40.37 & 22.68 & 27.83 & 22.56 & 25.67 & 69.0 & 84.2 & 40.17 & 5.5 & 72.0 & 60.38 & 41.98 & 41.87 \\

 \cmidrule{1-19}
  \multirow{13}*{\raisebox{10.0em}{\rotatebox{90}{KV-size 512}}}
  &   StreamingLLM  & 13.98 & 23.72 & 20.26 & 35.82 & 29.76 & 11.34 & 22.12 & 19.56 & 24.49 & 45.0 & 54.98 & 38.32 & 4.5 & 67.0 & 58.16 & 52.63 & 32.6 \\
 &   ExpectedAttention  & 19.73 & 33.41 & 30.2 & 45.06 & 32.81 & 20.43 & 25.55 & 21.45 & 26.25 & 51.0 & 85.76 & 39.57 & 6.0 & 56.0 & 62.0 & 54.84 & 38.13 \\
 &  TOVA & 18.84 & 33.46 & 44.0 & 48.36 & 36.82 & 21.47 & 23.07 & 20.72 & 24.33 & 63.0 & 88.91 & 41.01 & 6.0 & 71.0 & 64.66 & 58.33 & 41.5 \\
  \cdashline{2-19}
 &   SnapKV           & 19.24 & 36.51 & 43.61 & 46.83 & 36.62 & 23.11 & 22.62 & 21.17 & 24.03 & 45.0 & 88.59 & 40.09 & 6.0 & 71.5 & 63.75 & 58.65 & 40.46 \\
 &   +\think~(0.5)    & 18.73 & 33.83 & 41.47 & 43.72 & 27.98 & 20.91 & 20.59 & 21.56 & 22.25 & 15.5 & 84.62 & 33.82 & 7.0 & 71.5 & 57.01 & 56.97 & 36.09 \\
 &   +\think~(0.8)    & 9.48 & 6.59 & 18.62 & 18.28 & 8.32 & 9.2 & 17.11 & 15.37 & 16.46 & 0.0 & 43.94 & 8.6 & 2.21 & 34.62 & 33.43 & 35.47 & 17.36 \\
 \hc &   +\textbf{\our}~(0.5)      & 18.66 & 36.13 & 43.23 & 46.66 & 36.17 & 22.86 & 22.44 & 21.19 & 23.7 & 42.5 & 89.11 & 40.15 & 6.5 & 71.5 & 63.8 & 59.0 & 40.22 \\
 \hd &   +\textbf{\our}~(0.8)      & 18.23 & 37.34 & 42.42 & 44.71 & 34.85 & 23.14 & 21.8 & 21.26 & 23.68 & 41.5 & 87.22 & 38.88 & 5.0 & 72.5 & 62.86 & 55.01 & 39.40 \\
  \cdashline{2-19}
 &   PyramidKV        & 21.79 & 44.6 & 45.96 & 48.33 & 43.63 & 25.82 & 30.42 & 22.45 & 26.96 & 72.0 & 88.69 & 41.59 & 6.0 & 71.5 & 62.21 & 48.72 & 43.79 \\
 &   +\think~(0.5)    & 22.48 & 40.56 & 47.94 & 45.83 & 34.95 & 23.19 & 27.55 & 22.54 & 25.6 & 53.5 & 84.88 & 32.7 & 7.78 & 71.0 & 53.9 & 51.54 & 40.37 \\
 &   +\think~(0.8)    &  6.37 & 5.53 & 13.73 & 12.53 & 5.47 & 3.16 & 16.97 & 14.21 & 17.11 & 0.0 & 23.03 & 7.54 & 1.73 & 13.0 & 29.67 & 27.51 & 12.35\\
 \hc &   +\textbf{\our}~(0.5)    &  22.79 & 43.99 & 45.63 & 48.83 & 43.64 & 24.87 & 30.34 & 22.89 & 26.57 & 70.0 & 88.75 & 42.28 & 6.5 & 71.5 & 62.72 & 50.81 & 43.88 \\
 \hd &   +\textbf{\our}~(0.8)    & 22.73 & 44.1 & 47.2 & 46.47 & 40.51 & 22.81 & 26.66 & 22.72 & 24.87 & 68.0 & 88.63 & 40.44 & 5.5 & 72.0 & 59.61 & 42.44 & 42.17 \\

 \bottomrule
\end{tabular}}
\caption{Performance comparison on LLaMA-3-8B-Instruct at LongBench. \textbf{\our}~($\lambda$) and \think($\lambda$) denote the channel-wise key cache pruning ratio $\lambda$. Full results including other cache budgets and additional models are provided in Appendix~\ref{app:ex_longbench}.}
\label{tab:main}
    \end{table*}

\noindent \textbf{Saliency Measurement and Unstructured Pruning.}
We compute the proxy saliency score $\boldsymbol{w}_{j,t}$ for each channel $j$ and token $t$ to estimate per-channel contribution to the attention mechanism. We sort the scores in descending order and construct a binary pruning mask $\mathcal{S}_i \in \{0,1\}^{S \times D}$ for head $i$, retaining the top-$T$ channels. The pruned key matrix is denoted as $\boldsymbol{\hat{\mathcal{K}}}_i = \boldsymbol{\mathcal{K}}_i[\mathcal{S}_i] \in \mathbb{R}^{S \times T}$, where $\boldsymbol{\mathcal{K}}_i[\mathcal{S}_i]$ extracts the channels indexed by \(\mathcal{S}\).
To support recovery during decoding, we further compute the distributional statistics\footnote{Detailed formulations are provided in Appendix~\ref{app:cache}.} (mean $\mu_i$, standard deviation $\sigma_i$) of the saliency scores, or the mean of pruned entries $\mu_{i,\text{pruned}}$. These statistics are critical for recovering approximations of the pruned channels as our goal is to select channels with lower final attention scores, rather than those with inherently small key entries, given the non-trivial dependency of scores on query key interactions.


\noindent \textbf{Channel Recovery.}
Based on Observation 2 in Section~\ref{main:motivation}, we propose a \emph{query-aware recovery function} $\boldsymbol{\mathcal{F}}$ to reconstruct pruned key channels, addressing the limitations of discard or fixed-value replacement.
We utilize cached distributional statistics collected during the prefill stage to sample plausible score values and then back-compute the corresponding key entries. Specifically, we sample a score
$\tilde{\boldsymbol{w}}_{j,t}$
and the sampled key entry is computed as $\tilde{\mathbf{k}}_{i,t}^j = \frac{\tilde{\boldsymbol{w}}_{i,t}^j}{\| \overline{\mathbf{q}}_i^j \|_2}$,
ensuring that the inner product $\langle \overline{\mathbf{q}}_i^j, \tilde{\mathbf{k}}_{i,t}^j \rangle \approx \tilde{\boldsymbol{w}}_{i,t}^j$, consistent with the sampled score. We consider the following instantiations of the recovery function $\boldsymbol{\mathcal{F}}$:
\begin{itemize}
\item \textbf{Gaussian distribution:} \quad $\tilde{\boldsymbol{w}}_{i,t}^j \sim \mathcal{N}(\mu_i, \sigma_i^2)$
\item \textbf{Exponential distribution:} \quad $\tilde{\boldsymbol{w}}_{i,t}^j \sim \mathrm{Exp}(1/\mu_i)$
\item \textbf{Degenerate (only $\mu$) distribution:} \quad $\tilde{\boldsymbol{w}}_{i,t}^j = \mu_{i,\text{pruned}}$
\end{itemize}
The choice of distribution is flexible and can be configured per head or globally. Empirically, degenerate sampling performs robustly across tasks and layers. Overall, the $\boldsymbol{\mathcal{F}}$ is defined as:
\begin{equation}
    \tilde{\mathbf{k}}_{i,t}^j = \boldsymbol{\mathcal{F}}(\mu, \sigma) =  \frac{\text{sample}(\text{dist}(\mu_, \sigma))}{\|\overline{\mathbf{q}}_i^j\|_2},
\end{equation}
Finally, we reconstruct the full key matrix $\tilde{\boldsymbol{\mathcal{K}}}_{i}$ by combining the cached pruned keys with the sampled keys according to the mask $\mathcal{S}_{i}$, ensuring both structural completeness and numerical consistency of the attention computation.


\begin{table*}[t]
\centering
\scriptsize
\setlength{\tabcolsep}{4.5pt}{
\begin{tabular}{lc*{13}{>{\centering\arraybackslash}p{0.6cm}}}
\toprule
\textbf{Method} & \textbf{Niah1} & \textbf{Niah2} & \textbf{Niah3} & \textbf{MKey1} & \textbf{MKey2} & \textbf{MKey3} & \textbf{MValue} & \textbf{MQuery} & \textbf{VT} & \textbf{CWE} & \textbf{FWE} & \textbf{QA1} & \textbf{QA2} & \textbf{Avg.} \\
\midrule
 Vanilla    & 100.0 & 100.0 & 100.0 & 99.6 & 100 & 99.2 & 99.1 & 99.0 & 99.8 & 88.9 & 90.0 & 81.0 & 57.2 & 93.36 \\
\midrule
 StreamingLLM  & 18.8 & 17.4 & 19.0 & 20.2 & 20.0 & 18.4 & 18.25 & 18.2 & 32.84 & 0.18 & 81.33 & 31.4 & 33.6 & 25.35\\
 ExpectedAttention  &  99.2 & 42.0 & 3.4 & 33.8 & 57.0 & 0.8 & 9.35 & 21.1 & 66.12 & 54.46 & 70.6 & 72.0 & 48.2 & 44.46\\
 TOVA      &  100.0 & 100.0 & 97.8 & 99.4 & 96.8 & 0.4 & 98.9 & 99.25 & 99.76 & 54.04 & 90.8 & 77.4 & 54.6 & 82.24\\
\cdashline{1-15}
 SnapKV        & 100.0 & 100.0 & 10.0 & 99.8 & 97.2 & 63.2 & 97.7 & 99.45 & 97.36 & 53.92 & 85.73 & 80.8 & 57.2 & 80.18 \\
 +\think(0.5)  & 96.6 & 99.6 & 9.4 & 99.0 & 92.2 & 55.4 & 98.55 & 98.25 & 94.84 & 29.12 & 88.87 & 76.0 & 50.6 & 76.03 \\
 +\think(0.8)  & 0.0 & 0.0 & 0.0 & 0.0 & 0.0 & 0.0 & 0.05 & 0.0 & 0.0 & 0.32 & 0.0 & 18.8 & 20.2 & 3.03\\
 \hc +\textbf{\our}(0.5)    & 100.0 & 100.0 & 10.2 & 99.4 & 96.6 & 62.8 & 98.05 & 99.45 & 97.64 & 53.8 & 86.2 & 80.8 & 56.0 & 80.07 \\
 \hd +\textbf{\our}(0.8)    & 100.0 & 99.8 & 9.6 & 99.2 & 94.2 & 49.4 & 98.1 & 98.75 & 96.64 & 41.12 & 87.07 & 80.0 & 53.8 & 77.51\\
\cdashline{1-15}
 PyramidKV  & 100.0 & 100.0 & 5.0 & 99.8 & 98.2 & 55.0 & 98.6 & 99.35 & 98.6 & 16.88 & 87.0 & 80.0 & 57.2 & 76.59\\
 +\think(0.5)  & 97.2 & 100.0 & 4.8 & 99.4 & 93.0 & 49.2 & 98.7 & 98.75 & 96.16 & 8.46 & 88.33 & 76.2 & 52.4 & 74.05 \\
 +\think(0.8)  & 0.0 & 0.0 & 0.0 & 0.0 & 0.0 & 0.0 & 0.0 & 0.0 & 0.0 & 0.24 & 0.0 & 14.8 & 19.4 & 2.65\\
 \hc +\textbf{\our}(0.5)    & 99.2 & 99.2 & 5.2 & 99.4 & 97.6 & 54.4 & 97.95 & 98.7 & 98.16 & 16.84 & 86.27 & 79.6 & 56.8 & 76.1\\
 \hd +\textbf{\our}(0.8)    & 99.4 & 98.8 & 5.2 & 99.2 & 94.4 & 44.2 & 97.1 & 97.7 & 95.24 & 12.08 & 86.2 & 78.4 & 54.0 & 73.99\\
\bottomrule
\end{tabular}
}
\caption{RULER evaluation results on the LLaMA3.1-8B-Instruct model with \our{} under a 20\% KV cache budget and 16K input length. Additional results across varying cache budgets and input lengths are reported in  Appendix~\ref{app:ex_ruler} for completeness.
}
\label{tab:main_ruler}
\end{table*}

\section{Experiments}
\label{main:Experiments}

\subsection{Experimental Setup}
\label{main:setup}

\noindent \textbf{Benchmark Datasets.}  
We evaluate our \our{} against state-of-the-art KV cache compression methods on three widely recognized long-context understanding benchmarks: LongBench \citep{longbench} and RULER \citep{ruler} to thoroughly assess \our’s achievable performance.

\noindent \textbf{Implementation Details.} 
To validate \our’s general effectiveness, we evaluate on LLMs of varying scales and capabilities, including LLaMA-3/3.1-8/70B-Instruct \citep{llama3}, Qwen3-8B/32B \citep{qwen3}. 
To ensure a fair comparison between KV cache compression strategies and their integration with \our, we adopt consistent hyperparameter settings across all settings. Unless otherwise specified, we apply \our{} to the \textit{key cache} only and use the \textit{degenerate distribution} as the default recovery strategy.

\noindent \textbf{Baselines.} 
We benchmark \our{} against the standard full KV cache and prior KV cache compression methods, including StreamingLLM \citep{streamingllm}, PyramidKV \citep{pyramidinfer}, SnapKV \citep{snapkv} and ExpectedAttention \citep{kvpress} under various cache budgets. 

\noindent Additional experimental details can refer to Appendix~\ref{app:Implementation}.

\subsection{Benchmark on LongBench}
\label{main:benchmark_longbench}

Table \ref{tab:main} presents the performance comparison of KV compression methods and their integration with our proposed key cache channel pruning for LLaMA-3-8B-Instruct,
evaluated in various KV budgets on the LongBench. 
The pruning ratio $\lambda=0.8$ indicates that 80\% of key cache channels are removed, resulting in a 40\% reduction in the total KV cache memory.
The following observations can be drawn: 

\noindent \textbf{Compatibility with Existing Methods.} When integrated with token eviction strategies (e.g., PyramidKV), \our{} further boosts effectiveness. Comparisons between SnapKV and PyramidKV integrated with channel pruning further validate the robustness and general applicability of \our. Notably, the stronger the eviction strategy, the greater the gains observed from incorporating \our. Combining \our(0.5) outperforms the integrated eviction baseline and combining \our(0.8) maintains 95\% of accuracy while reducing cache storage by 40\%.

\noindent \textbf{Superior Performance under High Pruning Ratios.} 
\our{} consistently outperforms \think{} across all budgets and pruning ratios. 
In particular, under a high pruning ratio ($\lambda = 0.8$), we observe that integrating \think{} with either SnapKV or PyramidKV leads to substantial degradation in performance (average drop of \textbf{65\%}). In contrast, combining \textbf{\our} with the same baselines incurs less than \textbf{5\%} average performance loss. 
\our's recoverable pruning preserves both expressivity and stability even at 80\% sparsity, while \think{} suffers catastrophic degradation.


\begin{figure*}[t]
\centerline{\includegraphics[width=1.0\textwidth]{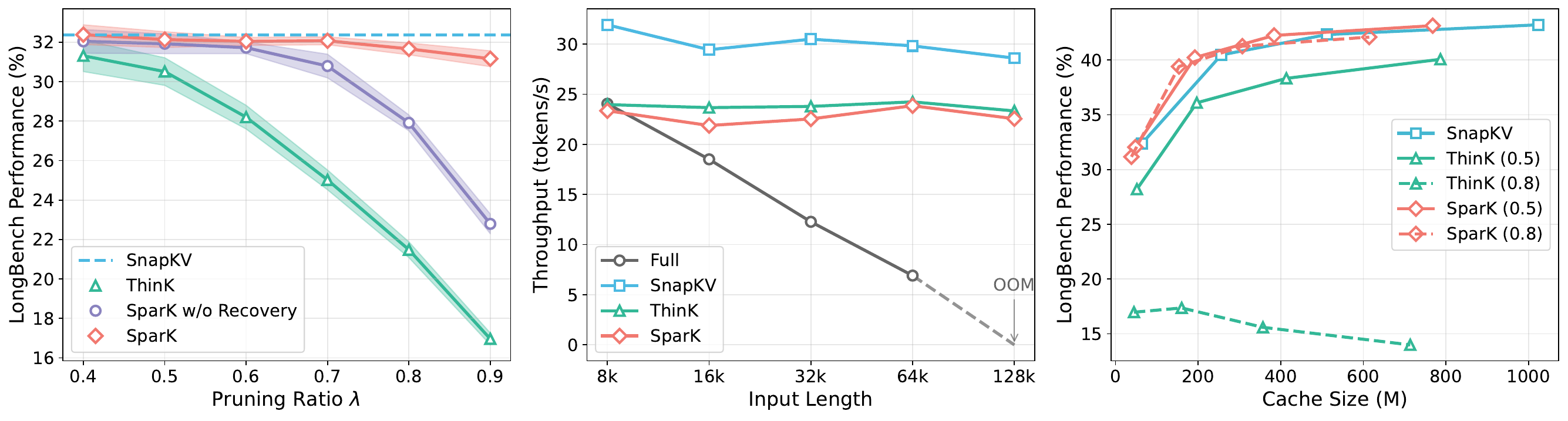}}
\caption{
Performance–Efficiency analysis of \our{} on LLaMA3-8B-Instruct.
(a) LongBench average performance under varying pruning ratios ($\lambda$). \our{} significantly outperforms \think{} across all compression levels.
(b) Throughput (tokens/s) with increasing input length. \our{} maintains stable decoding speed across long sequences (up to 128k)
(c) Cache size vs. performance trade-off. \our{} achieves favorable efficiency–performance balance compared to \think{} and SnapKV.
}
\label{fig:analysis}
\end{figure*}

\subsection{Benchmark on RULER}
\label{main:benchmark_RULER}


Table \ref{tab:main_ruler} presents the results of RULER under 20\% cache budget. \our{} consistently outperforms \think{} while preserving competitive accuracy under all settings. 
Notably, under a stringent cache budget (20\% or 50\%) with 8K and 16K inputs, \think{} (0.8) suffers drastic degradation with performance dropping below 3\%, while \our{} (0.8) retains accuracy within 3\% of baseline eviction methods, highlighting the effectiveness of our recovery mechanism. Even at moderate pruning (e.g., 0.5), \our{} consistently outperforms \think{} and matches or surpasses baseline strategies, demonstrating both accuracy preservation and general applicability of our method \our.


\subsection{Analysis}
\label{main:analysis}

We conduct a comprehensive evaluation of \our{} across three key dimensions: pruning ratio, input length, and cache size. Results are summarized in Figure~\ref{fig:analysis}.

\noindent \textbf{Impact of Pruning Ratio.}
Figure~\ref{fig:analysis}(a) shows that \our{} consistently outperforms \think{} and the unrecovered variant, particularly under high compression. At $\lambda=0.8$, \think{} incurs a performance drop exceeding 35\%, whereas \our{} maintains a degradation within 5\%. This highlights the effectiveness of channel-aware pruning and query-aware recovery in preserving attention quality.

\noindent \textbf{Throughput under Long Inputs.}
Figure~\ref{fig:analysis}(b) illustrates the decoding throughput across varying input lengths with KV budget of 128. While the full-cache baseline fails beyond 64k due to memory overflow, \our{} sustains high throughput across all lengths. Notably, \our{} achieves comparable throughput to \think, despite the added recovery step. This indicates that the recovery mechanism introduces negligible overhead in decoding latency.

\noindent \textbf{Cache Size vs. Performance.}
As shown in Figure~\ref{fig:analysis}(c), \our{} achieves superior performance under the same or smaller cache budgets. 
By pruning key channels, both \our{} and \think{} achieve lower memory usage than SnapKV under the same KV size. Compared to \think, \our{} consistently delivers performance closer to SnapKV across varying compression ratios. Under equal memory budgets, \our{} outperforms all baselines, underscoring its effectiveness in complementing KV compression methods for improved memory efficiency.


\noindent \textbf{Pruning Value Cache Channels.} 
We further extend \our{} to support simultaneous pruning of both key and value cache channels ($\lambda_k$ + $\lambda_v$) in the Appendix \ref{app:value}. As shown in Table~\ref{tab:app_kv_llama3-8b}, \our{} maintains strong robustness under joint pruning. For example, under the (0.5+0.5) configuration with SnapKV in 128 KV-size, the average performance drops marginally from 32.38 to 32.03, despite a further reduction in memory footprint. 
Notably, the results of (0.5 + 0.3) and (0.5 + 0.5) configuration achieve \textbf{comparable or even superior} performance to the (0.5) configuration. Although extreme compression (0.8+0.8) leads to more noticeable accuracy drops, the recovery mechanism ensures that the additional loss remains within 5\% on average. These results demonstrate that \our{} generalizes effectively to joint KV pruning, enabling greater memory savings under moderate settings while preserving task performance, and highlight the flexibility of our channel-wise sparsity and the critical role of recovery in maintaining accuracy.

\begin{table}[t]
\centering
\scriptsize
\setlength{\tabcolsep}{2.8pt}{
\begin{tabular}{lcccccccc}
\toprule
\multirow{2}{*}{\textbf{Dist.}} & \multicolumn{4}{c}{$\lambda=0.5$} & \multicolumn{4}{c}{$\lambda=0.8$}  \\
\cmidrule(r){2-5}\cmidrule(r){6-9}
& \textbf{128} & \textbf{512} & \textbf{1024} & \textbf{2048} & \textbf{128} & \textbf{512} & \textbf{1024} & \textbf{2048} \\
\midrule
Norm. & \textbf{32.71} & 39.76 & 41.37 & 42.18 & 31.25 & 38.78 & 41.03 & 41.68 \\
Exp.  & 32.56 & 40.16 & 42.18 & 43.04 & \textbf{31.43} & 39.04 & 41.21 & 41.87 \\
\hd Deg.  & 32.04 & \textbf{40.22} & \textbf{42.24} & \textbf{43.13} & 31.16 & \textbf{39.41} & \textbf{41.26} & \textbf{42.09} \\
\bottomrule
\end{tabular}
}
\caption{Ablation study on recovery distribution.}
\label{tab:ab_dist}
\end{table}

\subsection{Ablation Studies}
\label{main:ab}

Unless stated otherwise, all ablation experiments are conducted on the LongBench benchmark using the LLaMA3-8B-Instruct model with various KV budgets.

\noindent \textbf{Recovery Distributions.}
We investigate the impact of different recovery distributions under two pruning ratios ($\lambda = 0.5$ and $0.8$). As shown in Table~\ref{tab:ab_dist}, all three strategies Degenerate, Gaussian (Normal) and Exponential perform comparably, indicating that \our{} is robust to the choice of statistical modeling.
Degenerate recovery outperforms other strategies, particularly on long inputs, suggesting its stability under aggressive pruning. While Gaussian and Exponential offer moderate flexibility, they tend to introduce slight noise that may not always benefit attention approximation when key is highly limited.
The exponential distribution yields slightly better results at short sequences, likely due to its heavier tail offering greater diversity in sampled keys.

\begin{table}[t]
\centering
\scriptsize
\setlength{\tabcolsep}{2.8pt}{
\begin{tabular}{lccccccccc}
\toprule
\multirow{2}{*}{\textbf{Variants}} & \textbf{Pruning} & \textbf{Threshold} & \textbf{Group} & \multicolumn{4}{c}{\textbf{KV-Size}} & \textbf{Overall}\\
\cmidrule{5-8}
& \textbf{Ratio ($\lambda$)} & \textbf{(p)} & \textbf{(g)} & \textbf{128} & \textbf{512} & \textbf{1024} & \textbf{2048} & \textbf{Ratio}\\
\midrule
\hd \our   & 0.5 & - & - & 32.04 & 40.22 & 42.24 & 43.13 & 0.50 \\ 
\our-p & - & 99\% & - & 32.11 & 40.13 & 42.18 & 42.95 & 0.58 \\ 
\our-g & - & -  & 5 & 32.06 & 40.17 & 42.17 & 42.76 & 0.55 \\ 
\our-g & - & -  & 4 & 32.11 & 40.11 & 42.45 & 43.27 & 0.44 \\ 
\bottomrule
\end{tabular}
}
\caption{Ablation study on variants.}
\label{tab:ab_var}
\end{table}

\noindent \textbf{Adaptive Variants of \our.} 
We further explore two adaptive variants of \our{} that remove the need for a predefined pruning ratio. The first variant, \our-p, applies a top-$p$ thresholding strategy by greedily selecting the minimum number of salient channels per token that cumulatively account for 99\% of the total saliency. The second variant, \our-g, groups the $D$ channels into $g$ disjoint segments with ascending importance and assigns progressively larger pruning ratios to less salient groups. Specifically, for $g=4$, we assign pruning ratios of $(0.25, 0.5, 0.75, 1.0)$; for $g=5$, we use $(0.1, 0.3, 0.5, 0.7, 0.9)$.
As shown in Table~\ref{tab:ab_var}, both variants achieve comparable accuracy to the fixed-ratio baseline, while offering greater flexibility. Notably, the grouped variant with $g=4$ achieves the highest overall performance (43.27 at 2048 input length) with a lower average pruning ratio (0.44), suggesting that fine-grained structured sparsity can lead to better trade-offs between compression and performance. These results underscore the potential of \our{} as a flexible and extensible framework for KV compression.

\section{Conclusion}

In this paper, we introduce \our, a novel channel-wise pruning that leverages unstructured sparsity alongside a lightweight statistical recovery mechanism. Unlike prior methods that suffer from significant degradation under high pruning ratios, \our{} preserves attention fidelity by selectively retaining salient channels and reconstructing pruned entries using cached statistics. Extensive experiments demonstrate that \our{} significantly reduces memory consumption and maintains competitive performance, highlighting the importance of channel recovery in mitigating the adverse effects of aggressive pruning.


\bibliography{aaai2026}

@inproceedings{liao2025neural,
	title={Neural-symbolic collaborative distillation: Advancing small language models for complex reasoning tasks},
	author={Liao, Huanxuan and He, Shizhu and Xu, Yao and Zhang, Yuanzhe and Liu, Kang and Zhao, Jun},
	booktitle={Proceedings of the AAAI Conference on Artificial Intelligence},
	volume={39},
	number={23},
	pages={24567--24575},
	year={2025}
}

@inproceedings{liao2025awakening,
	title={Awakening Augmented Generation: Learning to Awaken Internal Knowledge of Large Language Models for Question Answering},
	author={Liao, Huanxuan and He, Shizhu and Xu, Yao and Zhang, Yuanzhe and Liu, Shengping and Liu, Kang and Zhao, Jun},
	booktitle={Proceedings of the 31st International Conference on Computational Linguistics},
	pages={1333--1352},
	year={2025}
}

@article{liao2024instance,
	title={From instance training to instruction learning: Task adapters generation from instructions},
	author={Liao, Huanxuan and He, Shizhu and Xu, Yao and Zhang, Yuanzhe and Hao, Yanchao and Liu, Shengping and Liu, Kang and Zhao, Jun},
	journal={Advances in Neural Information Processing Systems},
	volume={37},
	pages={45552--45577},
	year={2024}
}

@article{tan2025dynamic,
  title={Dynamic parametric retrieval augmented generation for test-time knowledge enhancement},
  author={Tan, Yuqiao and He, Shizhu and Liao, Huanxuan and Zhao, Jun and Liu, Kang},
  journal={arXiv preprint arXiv:2503.23895},
  year={2025}
}

@article{specache,
  title={SpeCache: Speculative Key-Value Caching for Efficient Generation of LLMs},
  author={Jie, Shibo and Tang, Yehui and Han, Kai and Deng, Zhi-Hong and Han, Jing},
  journal={arXiv preprint arXiv:2503.16163},
  year={2025}
}

@article{forest,
  title={Forest-of-thought: Scaling test-time compute for enhancing llm reasoning},
  author={Bi, Zhenni and Han, Kai and Liu, Chuanjian and Tang, Yehui and Wang, Yunhe},
  journal={arXiv preprint arXiv:2412.09078},
  year={2024}
}

@inproceedings{updp,
  title={Updp: A unified progressive depth pruner for cnn and vision transformer},
  author={Liu, Ji and Tang, Dehua and Huang, Yuanxian and Zhang, Li and Zeng, Xiaocheng and Li, Dong and Lu, Mingjie and Peng, Jinzhang and Wang, Yu and Jiang, Fan and others},
  booktitle={Proceedings of the AAAI conference on artificial intelligence},
  volume={38},
  number={12},
  pages={13891--13899},
  year={2024}
}

@Article {MIR-2023-10-223,
author = {Zhao Zhang and  Yangcheng Gao and  Jicong Fan and  Zhongqiu Zhao and  Yi Yang and  Shuicheng Yan },
journal = {Machine Intelligence Research},
title = {SelectQ: Calibration Data Selection for Post-training Quantization},
year = {2025},
volume = {22},
issue = {3},
pages = {499-510},
doi = {10.1007/s11633-024-1518-0} 
}

@article{shadowkv,
  title={Shadowkv: Kv cache in shadows for high-throughput long-context llm inference},
  author={Sun, Hanshi and Chang, Li-Wen and Bao, Wenlei and Zheng, Size and Zheng, Ningxin and Liu, Xin and Dong, Harry and Chi, Yuejie and Chen, Beidi},
  journal={arXiv preprint arXiv:2410.21465},
  year={2024}
}

@article{sampleattention,
  title={Sampleattention: Near-lossless acceleration of long context llm inference with adaptive structured sparse attention},
  author={Zhu, Qianchao and Duan, Jiangfei and Chen, Chang and Liu, Siran and Li, Xiuhong and Feng, Guanyu and Lv, Xin and Cao, Huanqi and Chuanfu, Xiao and Zhang, Xingcheng and others},
  journal={arXiv preprint arXiv:2406.15486},
  year={2024}
}

@article{razorattention,
  title={Razorattention: Efficient kv cache compression through retrieval heads},
  author={Tang, Hanlin and Lin, Yang and Lin, Jing and Han, Qingsen and Hong, Shikuan and Yao, Yiwu and Wang, Gongyi},
  journal={arXiv preprint arXiv:2407.15891},
  year={2024}
}

@inproceedings{cam,
  title={Cam: Cache merging for memory-efficient llms inference},
  author={Zhang, Yuxin and Du, Yuxuan and Luo, Gen and Zhong, Yunshan and Zhang, Zhenyu and Liu, Shiwei and Ji, Rongrong},
  booktitle={Forty-first international conference on machine learning},
  year={2024}
}

@inproceedings{xiaoduoattention,
  title={DuoAttention: Efficient Long-Context LLM Inference with Retrieval and Streaming Heads},
  author={Xiao, Guangxuan and Tang, Jiaming and Zuo, Jingwei and Yang, Shang and Tang, Haotian and Fu, Yao and Han, Song and others},
  booktitle={The Thirteenth International Conference on Learning Representations},
  year={2025}
}

@article{layer,
  title={Layer-condensed kv cache for efficient inference of large language models},
  author={Wu, Haoyi and Tu, Kewei},
  journal={arXiv preprint arXiv:2405.10637},
  year={2024}
}

@article{reducing,
  title={Reducing transformer key-value cache size with cross-layer attention},
  author={Brandon, William and Mishra, Mayank and Nrusimha, Aniruddha and Panda, Rameswar and Ragan-Kelley, Jonathan},
  journal={Advances in Neural Information Processing Systems},
  volume={37},
  pages={86927--86957},
  year={2024}
}

@inproceedings{vllm,
  title={Efficient Memory Management for Large Language Model Serving with PagedAttention},
  author={Woosuk Kwon and Zhuohan Li and Siyuan Zhuang and Ying Sheng and Lianmin Zheng and Cody Hao Yu and Joseph E. Gonzalez and Hao Zhang and Ion Stoica},
  booktitle={Proceedings of the ACM SIGOPS 29th Symposium on Operating Systems Principles},
  year={2023}
}

@article{squeezed,
  title={Squeezed attention: Accelerating long context length llm inference},
  author={Hooper, Coleman and Kim, Sehoon and Mohammadzadeh, Hiva and Maheswaran, Monishwaran and Paik, June and Mahoney, Michael W and Keutzer, Kurt and Gholami, Amir},
  journal={arXiv preprint arXiv:2411.09688},
  year={2024}
}

@article{dynamic,
  title={Dynamic memory compression: Retrofitting llms for accelerated inference},
  author={Nawrot, Piotr and {\L}a{\'n}cucki, Adrian and Chochowski, Marcin and Tarjan, David and Ponti, Edoardo M},
  journal={arXiv preprint arXiv:2403.09636},
  year={2024}
}

@article{h2o,
  title={H2o: Heavy-hitter oracle for efficient generative inference of large language models},
  author={Zhang, Zhenyu and Sheng, Ying and Zhou, Tianyi and Chen, Tianlong and Zheng, Lianmin and Cai, Ruisi and Song, Zhao and Tian, Yuandong and R{\'e}, Christopher and Barrett, Clark and others},
  journal={Advances in Neural Information Processing Systems},
  volume={36},
  pages={34661--34710},
  year={2023}
}

@article{model,
  title={Model tells you where to merge: Adaptive kv cache merging for llms on long-context tasks},
  author={Wang, Zheng and Jin, Boxiao and Yu, Zhongzhi and Zhang, Minjia},
  journal={arXiv preprint arXiv:2407.08454},
  year={2024}
}

@article{modeltell,
  title={Model tells you what to discard: Adaptive kv cache compression for llms},
  author={Ge, Suyu and Zhang, Yunan and Liu, Liyuan and Zhang, Minjia and Han, Jiawei and Gao, Jianfeng},
  journal={arXiv preprint arXiv:2310.01801},
  year={2023}
}

@article{fu2024challenges,
  title={Challenges in deploying long-context transformers: A theoretical peak performance analysis},
  author={Fu, Yao},
  journal={arXiv preprint arXiv:2405.08944},
  year={2024}
}

@article{d2o,
  title={D2o: Dynamic discriminative operations for efficient generative inference of large language models},
  author={Wan, Zhongwei and Wu, Xinjian and Zhang, Yu and Xin, Yi and Tao, Chaofan and Zhu, Zhihong and Wang, Xin and Luo, Siqi and Xiong, Jing and Zhang, Mi},
  journal={arXiv preprint arXiv:2406.13035},
  year={2024}
}

@article{rkv,
  title={R-KV: Redundancy-aware KV Cache Compression for Training-Free Reasoning Models Acceleration},
  author={Cai, Zefan and Xiao, Wen and Sun, Hanshi and Luo, Cheng and Zhang, Yikai and Wan, Ke and Li, Yucheng and Zhou, Yeyang and Chang, Li-Wen and Gu, Jiuxiang and others},
  journal={arXiv preprint arXiv:2505.24133},
  year={2025}
}

@article{q,
  title={Q-hitter: A better token oracle for efficient llm inference via sparse-quantized kv cache},
  author={Zhang, Zhenyu and Liu, Shiwei and Chen, Runjin and Kailkhura, Bhavya and Chen, Beidi and Wang, Zhangyang},
  journal={Proceedings of Machine Learning and Systems},
  volume={6},
  pages={381--394},
  year={2024}
}

@article{kvquant,
  title={Kvquant: Towards 10 million context length llm inference with kv cache quantization},
  author={Hooper, Coleman and Kim, Sehoon and Mohammadzadeh, Hiva and Mahoney, Michael W and Shao, Yakun S and Keutzer, Kurt and Gholami, Amir},
  journal={Advances in Neural Information Processing Systems},
  volume={37},
  pages={1270--1303},
  year={2024}
}

@article{cache,
  title={Cache Me If You Must: Adaptive Key-Value Quantization for Large Language Models},
  author={Shutova, Alina and Malinovskii, Vladimir and Egiazarian, Vage and Kuznedelev, Denis and Mazur, Denis and Surkov, Nikita and Ermakov, Ivan and Alistarh, Dan},
  journal={arXiv preprint arXiv:2501.19392},
  year={2025}
}

@inproceedings{longbench,
  title={LongBench: A Bilingual, Multitask Benchmark for Long Context Understanding},
  author={Bai, Yushi and Lv, Xin and Zhang, Jiajie and Lyu, Hongchang and Tang, Jiankai and Huang, Zhidian and Du, Zhengxiao and Liu, Xiao and Zeng, Aohan and Hou, Lei and others},
  booktitle={Proceedings of the 62nd Annual Meeting of the Association for Computational Linguistics (Volume 1: Long Papers)},
  pages={3119--3137},
  year={2024}
}

@article{ruler,
  title={RULER: What's the Real Context Size of Your Long-Context Language Models?},
  author={Hsieh, Cheng-Ping and Sun, Simeng and Kriman, Samuel and Acharya, Shantanu and Rekesh, Dima and Jia, Fei and Zhang, Yang and Ginsburg, Boris},
  journal={arXiv preprint arXiv:2404.06654},
  year={2024}
}

@article{quest,
  title={Quest: Query-aware sparsity for efficient long-context llm inference},
  author={Tang, Jiaming and Zhao, Yilong and Zhu, Kan and Xiao, Guangxuan and Kasikci, Baris and Han, Song},
  journal={arXiv preprint arXiv:2406.10774},
  year={2024}
}

@article{streamingllm,
  title={Efficient streaming language models with attention sinks},
  author={Xiao, Guangxuan and Tian, Yuandong and Chen, Beidi and Han, Song and Lewis, Mike},
  journal={arXiv preprint arXiv:2309.17453},
  year={2023}
}

@article{think,
  title={Think: Thinner key cache by query-driven pruning},
  author={Xu, Yuhui and Jie, Zhanming and Dong, Hanze and Wang, Lei and Lu, Xudong and Zhou, Aojun and Saha, Amrita and Xiong, Caiming and Sahoo, Doyen},
  journal={arXiv preprint arXiv:2407.21018},
  year={2024}
}

@inproceedings{pyramidinfer,
  title={PyramidInfer: Pyramid KV Cache Compression for High-throughput LLM Inference},
  author={Yang, Dongjie and Han, Xiaodong and Gao, Yan and Hu, Yao and Zhang, Shilin and Zhao, Hai},
  booktitle={Findings of the Association for Computational Linguistics ACL 2024},
  pages={3258--3270},
  year={2024}
}

@article{snapkv,
  title={Snapkv: Llm knows what you are looking for before generation},
  author={Li, Yuhong and Huang, Yingbing and Yang, Bowen and Venkitesh, Bharat and Locatelli, Acyr and Ye, Hanchen and Cai, Tianle and Lewis, Patrick and Chen, Deming},
  journal={Advances in Neural Information Processing Systems},
  volume={37},
  pages={22947--22970},
  year={2024}
}

@article{keep,
  title={Keep the cost down: A review on methods to optimize LLM's KV-cache consumption},
  author={Shi, Luohe and Zhang, Hongyi and Yao, Yao and Li, Zuchao and Zhao, Hai},
  journal={arXiv preprint arXiv:2407.18003},
  year={2024}
}

@article{fables,
  title={Fables: Evaluating faithfulness and content selection in book-length summarization},
  author={Kim, Yekyung and Chang, Yapei and Karpinska, Marzena and Garimella, Aparna and Manjunatha, Varun and Lo, Kyle and Goyal, Tanya and Iyyer, Mohit},
  journal={arXiv preprint arXiv:2404.01261},
  year={2024}
}

@article{qwen3,
  title={Qwen3 technical report},
  author={Yang, An and Li, Anfeng and Yang, Baosong and Zhang, Beichen and Hui, Binyuan and Zheng, Bo and Yu, Bowen and Gao, Chang and Huang, Chengen and Lv, Chenxu and others},
  journal={arXiv preprint arXiv:2505.09388},
  year={2025}
}

@misc{Gemini,
      title={Gemini 2.5: Pushing the Frontier with Advanced Reasoning, Multimodality, Long Context, and Next Generation Agentic Capabilities}, 
      author={Gheorghe Comanici and Eric Bieber and Mike Schaekermann and Ice Pasupat and Noveen Sachdeva and Inderjit Dhillon and Marcel Blistein and Ori Ram and Dan Zhang and others},
      year={2025},
      eprint={2507.06261},
      archivePrefix={arXiv},
      primaryClass={cs.CL},
      url={https://arxiv.org/abs/2507.06261}, 
}

@article{deepseek,
  title={Deepseek-v2: A strong, economical, and efficient mixture-of-experts language model},
  author={Liu, Aixin and Feng, Bei and Wang, Bin and Wang, Bingxuan and Liu, Bo and Zhao, Chenggang and Dengr, Chengqi and Ruan, Chong and Dai, Damai and Guo, Daya and others},
  journal={arXiv preprint arXiv:2405.04434},
  year={2024}
}

@article{mqa,
  title={Fast transformer decoding: One write-head is all you need},
  author={Shazeer, Noam},
  journal={arXiv preprint arXiv:1911.02150},
  year={2019}
}

@article{gqa,
  title={Gqa: Training generalized multi-query transformer models from multi-head checkpoints},
  author={Ainslie, Joshua and Lee-Thorp, James and De Jong, Michiel and Zemlyanskiy, Yury and Lebr{\'o}n, Federico and Sanghai, Sumit},
  journal={arXiv preprint arXiv:2305.13245},
  year={2023}
}

@article{kvsharer,
  title={Kvsharer: Efficient inference via layer-wise dissimilar kv cache sharing},
  author={Yang, Yifei and Cao, Zouying and Chen, Qiguang and Qin, Libo and Yang, Dongjie and Zhao, Hai and Chen, Zhi},
  journal={arXiv preprint arXiv:2410.18517},
  year={2024}
}

@article{not,
  title={Not all heads matter: A head-level kv cache compression method with integrated retrieval and reasoning},
  author={Fu, Yu and Cai, Zefan and Asi, Abedelkadir and Xiong, Wayne and Dong, Yue and Xiao, Wen},
  journal={arXiv preprint arXiv:2410.19258},
  year={2024}
}

@article{you,
  title={You only cache once: Decoder-decoder architectures for language models},
  author={Sun, Yutao and Dong, Li and Zhu, Yi and Huang, Shaohan and Wang, Wenhui and Ma, Shuming and Zhang, Quanlu and Wang, Jianyong and Wei, Furu},
  journal={Advances in Neural Information Processing Systems},
  volume={37},
  pages={7339--7361},
  year={2024}
}

@article{liao2025beyond,
  title={Beyond Hard and Soft: Hybrid Context Compression for Balancing Local and Global Information Retention},
  author={Liao, Huanxuan and Hu, Wen and Xu, Yao and He, Shizhu and Zhao, Jun and Liu, Kang},
  journal={arXiv preprint arXiv:2505.15774},
  year={2025}
}

@inproceedings{skintern,
  title={SKIntern: Internalizing Symbolic Knowledge for Distilling Better CoT Capabilities into Small Language Models},
  author={Liao, Huanxuan and He, Shizhu and Hao, Yupu and Li, Xiang and Zhang, Yuanzhe and Zhao, Jun and Liu, Kang},
  booktitle={Proceedings of the 31st International Conference on Computational Linguistics},
  pages={3203--3221},
  year={2025}
}

@article{liu2025comprehensive,
  title={A Comprehensive Survey on Long Context Language Modeling},
  author={Liu, Jiaheng and Zhu, Dawei and Bai, Zhiqi and He, Yancheng and Liao, Huanxuan and Que, Haoran and Wang, Zekun and Zhang, Chenchen and Zhang, Ge and Zhang, Jiebin and others},
  journal={arXiv preprint arXiv:2503.17407},
  year={2025}
}

@article{gpt4,
  title={Gpt-4 technical report},
  author={Achiam, Josh and Adler, Steven and Agarwal, Sandhini and Ahmad, Lama and Akkaya, Ilge and Aleman, Florencia Leoni and Almeida, Diogo and Altenschmidt, Janko and Altman, Sam and Anadkat, Shyamal and others},
  journal={arXiv preprint arXiv:2303.08774},
  year={2023}
}

@article{pytorch,
  title={Pytorch: An imperative style, high-performance deep learning library},
  author={Paszke, A},
  journal={arXiv preprint arXiv:1912.01703},
  year={2019}
}

@article{llama3,
  title={The Llama 3 Herd of Models},
  author={Abhimanyu Dubey and Abhinav Jauhri and Abhinav Pandey and Abhishek Kadian and Ahmad Al-Dahle and Aiesha Letman and Akhil Mathur and Alan Schelten and Amy Yang and Angela Fan and Anirudh Goyal and Anthony S. Hartshorn and Aobo Yang and Archi Mitra and Archie Sravankumar and Artem Korenev and Arthur Hinsvark and Arun Rao and Aston Zhang and et al.},
  journal={ArXiv},
  year={2024},
  volume={abs/2407.21783},
  url={https://api.semanticscholar.org/CorpusID:271571434}
}

@inproceedings{dao2023flashattention2,
  title={Flash{A}ttention-2: Faster Attention with Better Parallelism and Work Partitioning},
  author={Dao, Tri},
  booktitle={International Conference on Learning Representations (ICLR)},
  year={2024}
}

@misc{kvpress,
  title = {kvpress},
  author = {Simon Jegou and Maximilian Jeblick and David Austin and Alessio Devoto},
  year = {2024},
  month = {October},
  url = {https://github.com/NVIDIA/kvpress}
}

@article{zhu2024dip,
  title={Dip-go: A diffusion pruner via few-step gradient optimization},
  author={Zhu, Haowei and Tang, Dehua and Liu, Ji and Lu, Mingjie and Zheng, Jintu and Peng, Jinzhang and Li, Dong and Wang, Yu and Jiang, Fan and Tian, Lu and others},
  journal={Advances in Neural Information Processing Systems},
  volume={37},
  pages={92581--92604},
  year={2024}
}

\newpage

\appendix

\section{\our}

\subsection{Error Objective Expansion}
\label{app:err-derivation}

Our goal is to minimize the attention discrepancy after pruning, measured by the Frobenius norm between the original and pruned attention matrices:
\begin{equation}
\label{eq:err-obj}
\min_{\mathcal{S}_{i,t}} \; 
\mathcal{E}(\mathcal{S}_{i,t}) = 
\left\| \mathbf{q}_{i,t} \mathbf{k}_{i,t}^\top 
- (\mathbf{q}_{i,t} \mathcal{S}_{i,t}) (\mathbf{k}_{i,t} \mathcal{S}_{i,t})^\top \right\|_F.
\end{equation}

This objective is combinatorial and difficult to solve exactly. To enable efficient channel selection, we expand the squared Frobenius norm. Let $\mathbf{q}_{i,t}[j]$ and $\mathbf{k}_{i,t}[j]$ denote the $j$-th channel vector of query and key, respectively. Using the identity
\[
\|A - B\|_F^2 = \|A\|_F^2 + \|B\|_F^2 - 2 \langle A, B \rangle,
\]
we can rewrite the squared error as:
\begin{align}
& \mathcal{E}^2(\mathcal{S}_{i,t}) \notag \\
&= \sum_{j=1}^D \sum_{r=1}^D 
\langle \mathbf{q}_{i,t}[j], \mathbf{q}_{i,t}[r] \rangle 
\langle \mathbf{k}_{i,t}[j], \mathbf{k}_{i,t}[r] \rangle 
(1 - \boldsymbol{z}_{j,t} \boldsymbol{z}_{r,t}) \notag \\
&= \sum_{j=1}^D \| \mathbf{q}_{i,t}[j] \|_2^2 
\| \mathbf{k}_{i,t}[j] \|_2^2 (1 - \boldsymbol{z}_{j,t}) \notag \\
&\quad + 2 \sum_{\substack{j,r=1 \\ j<r}}^D 
\langle \mathbf{q}_{i,t}[j], \mathbf{q}_{i,t}[r] \rangle 
\langle \mathbf{k}_{i,t}[j], \mathbf{k}_{i,t}[r] \rangle 
(1 - \boldsymbol{z}_{j,t} \boldsymbol{z}_{r,t}).
\label{eq:err-expanded}
\end{align}

\subsection{Caching Pruned Channel Statistics}
\label{app:cache}

Specifically, for each attention head $i$, first, we identify the set of channels that were pruned after the Top-$T$ selection. Let this set of pruned channel indices be $\mathcal{C}_{i,\text{pruned}} = \mathcal{C}_i \setminus \hat{\mathcal{C}}_i$. We then compute the distribution statistics for the saliency scores $\boldsymbol{w}_{i,j}$ of all channels within the pruned set $\mathcal{C}_{i,\text{pruned}}$:

\noindent Mean ($\mu_{i,\text{pruned}}$):
\begin{equation}
    \mu_{i,\text{pruned}} = \frac{1}{|\mathcal{C}_{i,\text{pruned}}|} \sum_{j \in \mathcal{C}_{i,\text{pruned}}} \boldsymbol{w}_{i,j} \in \mathbb{R}^{S}
\end{equation}

\noindent Mean ($\mu_i$):
\begin{equation}
    \mu_i = \frac{1}{|\mathcal{C}_i|} \sum_{j \in \mathcal{C}_i} \boldsymbol{w}_{i,j} \in \mathbb{R}^{S}
\end{equation}
     
\noindent Standard Deviation ($\sigma_i$):
\begin{equation}
    \sigma_{i} = \sqrt{\frac{1}{|\mathcal{C}_{i}|} \sum_{j \in \mathcal{C}_{i}} (\boldsymbol{w}_{i,j} - \mu_{i})^2}
\end{equation}
Then these calculated statistics ($\mu_{i}$ and $\sigma_{i}$, or possibly just the mean of the pruned channels $\mu_{i, \text{pruned}}$) are cached. In later stages, when it's necessary to recover or compensate for the impact of pruned channels, these statistics enable the generation of more reasonable compensation values, mitigating performance degradation that would result from simple zero or constant padding.

\section{Observations}
\label{app:ex_observation}

\subsection{Coefficient of Variation (CV)}

The Coefficient of Variation (CV) is a standardized statistical measure that quantifies the relative variability of a dataset by expressing the standard deviation as a proportion of the mean. Formally, for a random variable $X$ with mean $\mu$ and standard deviation $\sigma$, the CV is defined as:
\begin{equation}
    CV = \frac{\sigma}{\mu} = \frac{\sqrt{\mathbb{E}[(X - \mu)^2]}}{\mathbb{E}[X]}
\end{equation}
This dimensionless metric enables direct comparison of variability across datasets with different scales and units, making it particularly suitable for analyzing heterogeneous patterns in neural network activations.

The CV analysis is particularly necessary for key channel pattern analysis because: (1) it captures the context-sensitivity of individual channels by measuring how much their contributions vary across different input tokens; (2) it provides a scale-invariant measure that allows comparison across channels with different activation magnitudes; and (3) it enables systematic categorization of channels based on their behavioral patterns, informing adaptive compression strategies.

In our context, we employ CV to quantify the variability of channel-wise attention key activations across tokens. High CV values indicate that the importance of a given channel varies significantly with the input context, suggesting that a globally fixed importance ranking may be insufficient. This motivates the use of token-dependent, dynamic channel pruning strategies over static, globally ranked pruning. Therefore, CV provides a principled criterion for evaluating the necessity of fine-grained, context-aware channel selection in our method.

\subsection{Token-Specific Channel Activation Patterns}

\begin{figure*}[t]
\centerline{\includegraphics[width=1.0\textwidth]{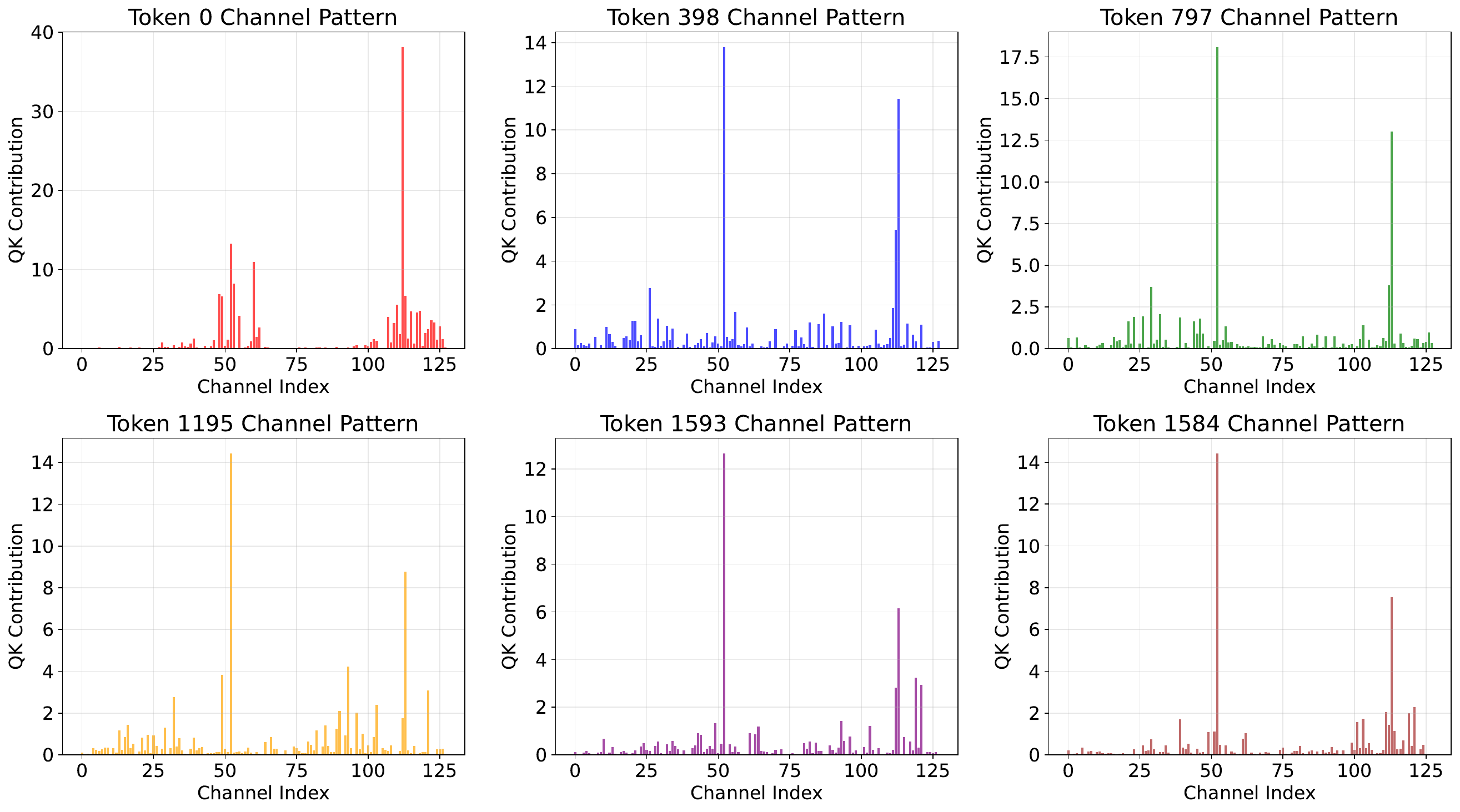}}
\caption{
Visualization of QK-score distributions across channel indices for 6 representative tokens. Brighter hues indicate higher attention contributions, revealing:
(1) Position-dependent sparsity (e.g., Token 0 vs 1195),
(2) Task-critical channel clustering,
(3) High variance in salient channel indices.
}
\label{fig:app_token_pattern}
\end{figure*}

\begin{figure*}[t]  
  \centering

  \begin{subfigure}[t]{0.48\textwidth}
    \includegraphics[width=\linewidth, trim=0 0 0 10mm, clip]{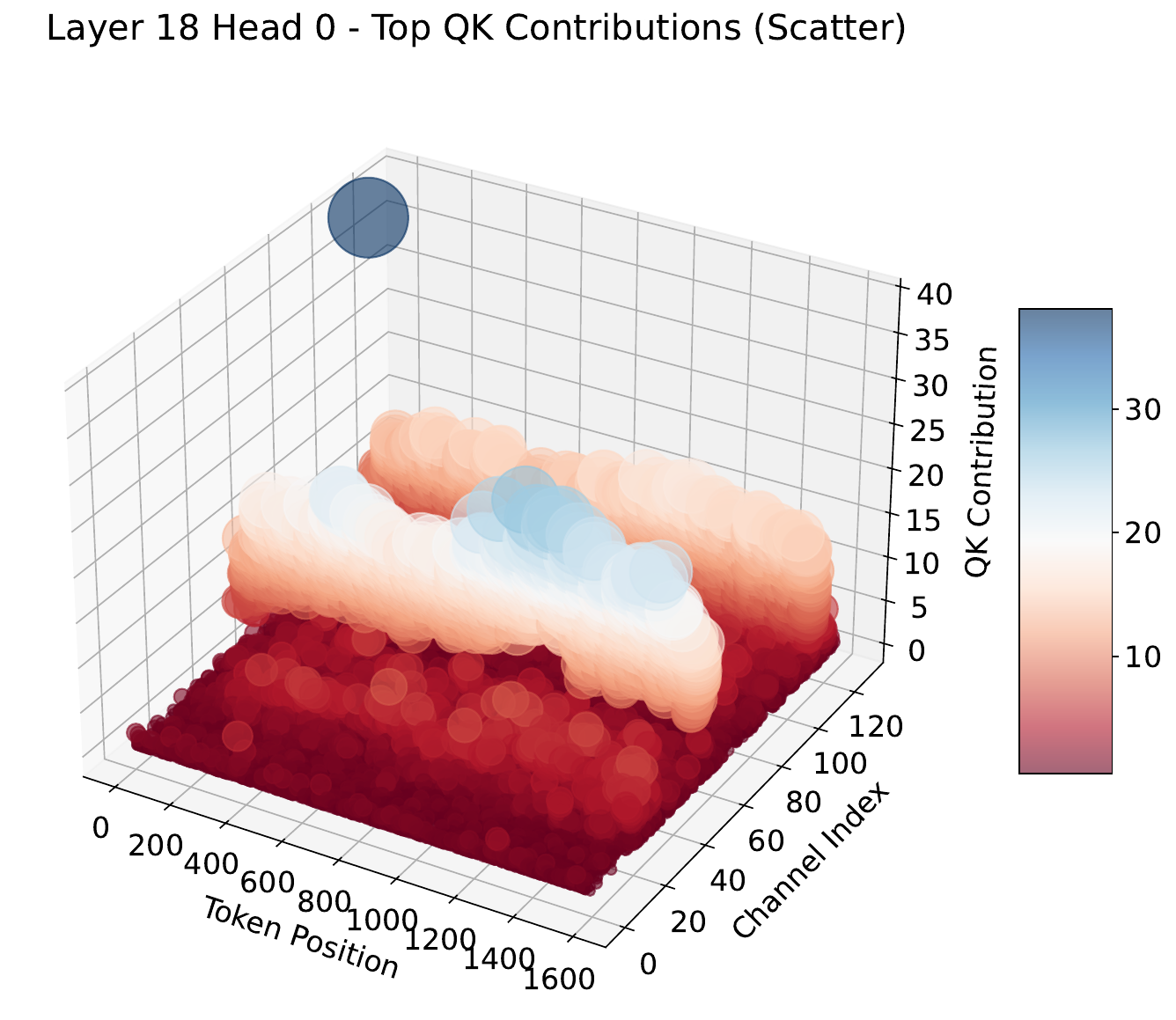}
    \caption{\textbf{3D scatter visualization} of score contributions, highlighting token-wise unstructured sparsity.}
  \end{subfigure}
  \hfill
  \begin{subfigure}[t]{0.48\textwidth}
    \includegraphics[width=\linewidth, trim=0 -80 0 10mm, clip]{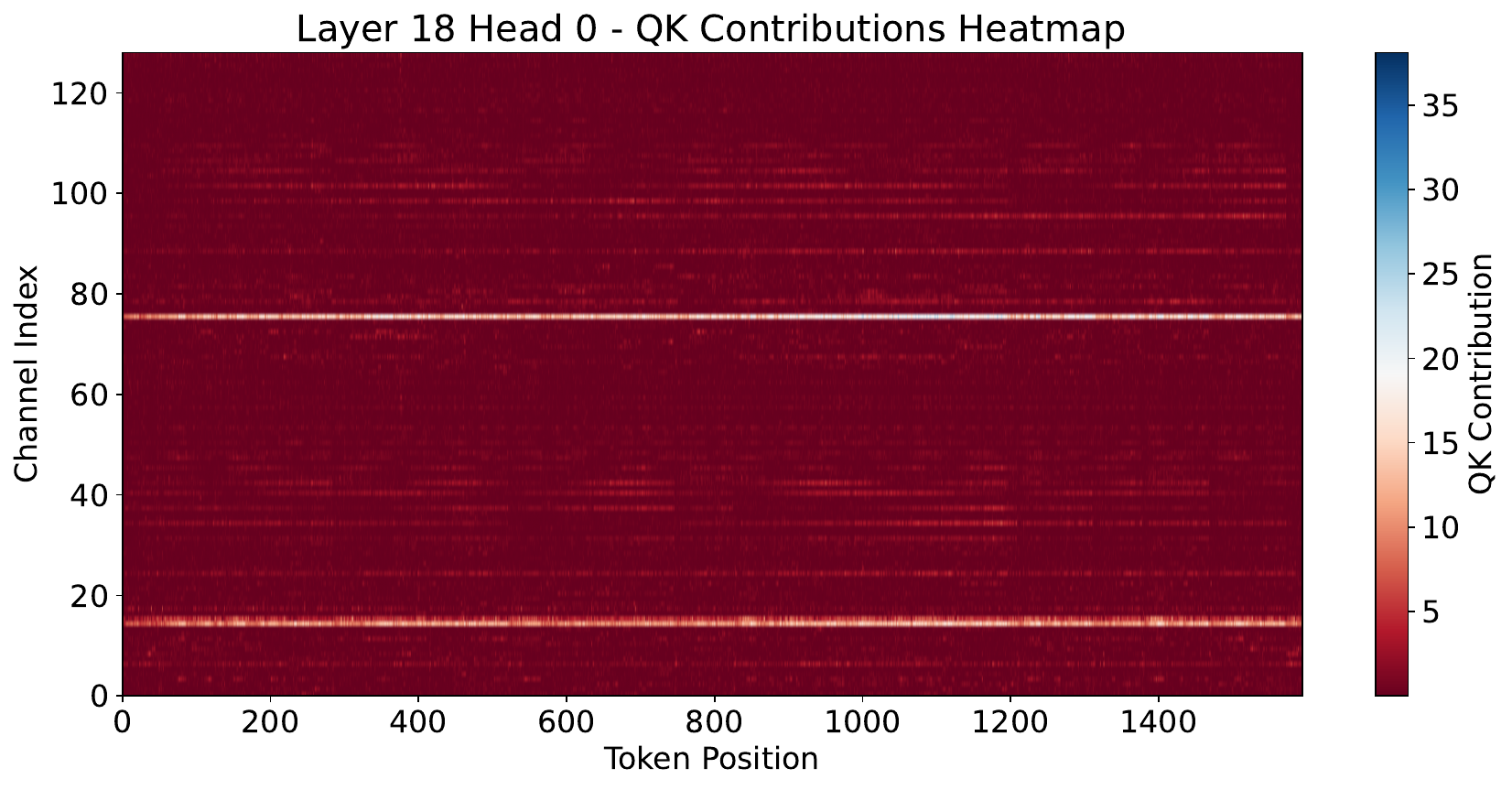}
    \caption{\textbf{Heatmap visualization} of channel distribution across channels.}
  \end{subfigure}
  \caption{More visualizations for motivations of layer 18 and head 0.}
  \label{fig:app_heat}
\end{figure*}

To gain deeper insight into how different channels contribute to the attention computation, we visualize the QK scores across channel indices for representative tokens in Figure \ref{fig:app_token_pattern} and heatmap of channels in Figure \ref{fig:app_heat}.

\section{Implementation}
\label{app:Implementation}

For all tasks, we use a batch size of 1 for evaluation and follow the settings of the based eviction method. For instance, when comparing SnapKV and SnapKV integrated with THINK, we used a maximum pooling kernel size of 7 and an observation window size of 32, maintaining the same KV size for both configurations.
For the RULER benchmark \citep{ruler}, we adopt 10 repetitions for each test unit and use context lengths of 16k and 8k.
We implement all experiments in PyTorch \citep{pytorch} and Flash Attention \citep{dao2023flashattention2}.

\section{Pruning Value Cache Channels}
\label{app:value}

\begin{table*}[t!]
    \centering \scriptsize

\resizebox{\textwidth}{!}{\begin{tabular}
{L{9.0em}@{\hspace{0.05ex}}C{3.8em}@{\hspace{0.05ex}}C{3.8em}@{\hspace{0.05ex}}C{3.8em}@{\hspace{0.05ex}}c@{\hspace{0.05ex}}c@{\hspace{0.05ex}}C{3.8em}@{\hspace{0.05ex}}c@{\hspace{0.05ex}}C{3.8em}@{\hspace{0.05ex}}c@{\hspace{0.05ex}}C{3.8em}@{\hspace{0.05ex}}c@{\hspace{0.05ex}}c@{\hspace{0.05ex}}c@{\hspace{0.6ex}}C{3.8em}@{\hspace{0.6ex}}C{3.8em}@{\hspace{0.6ex}}C{3.8em}@{\hspace{0.6ex}}c}
    \toprule
 \multirow{4}{*}{\textbf{Method}}& \multicolumn{3}{c}{\textbf{Single-Document QA}} & \multicolumn{3}{c}{\textbf{Multi-Document QA}}& \multicolumn{3}{c}{\textbf{Summarization}}& \multicolumn{3}{c}{\textbf{Few-shot Learning}}& \multicolumn{2}{c}{\textbf{Synthetic}} & \multicolumn{2}{c}{\textbf{Code}}&\multirow{4}{*}{\textbf{Avg.}} \\
& & \rotatebox[origin=c]{30}{\bf NrtvQA} & \rotatebox[origin=c]{30}{\bf Qasper} & \rotatebox[origin=c]{30}{\bf MF-en} & \rotatebox[origin=c]{30}{\bf HotpotQA} & \rotatebox[origin=c]{30}{\bf 2WikiMQA} & \rotatebox[origin=c]{30}{\bf Musique} & \rotatebox[origin=c]{30}{\bf GovReport} & \rotatebox[origin=c]{30}{\bf QMSum} & \rotatebox[origin=c]{30}{\bf MultiNews} & \rotatebox[origin=c]{30}{\bf TREC} & \rotatebox[origin=c]{30}{\bf TriviaQA} & \rotatebox[origin=c]{30}{\bf SAMSum} & \rotatebox[origin=c]{30}{\bf PCount} & \rotatebox[origin=c]{30}{~\bf PRe~~} & \rotatebox[origin=c]{30}{~\bf Lcc~~} & \rotatebox[origin=c]{30}{~\bf RB-P~} \\
\cmidrule{1-18}

 \multicolumn{18}{c}{LLaMA-3-8B-Instruct, KV-size 128}\\
 SnapKV &  15.29 & 20.03 & 29.2 & 39.92 & 28.26 & 15.06 & 17.74 & 19.27 & 18.05 & 21.0 & 68.64 & 36.64 & 6.0 & 66.0 & 57.86 & 59.08 & 32.38  \\
 \hc +\textbf{\our}~(0.5)    & 13.52 & 20.19 & 29.28 & 38.77 & 26.33 & 14.44 & 17.66 & 19.12 & 17.98 & 21.0 & 68.95 & 36.66 & 5.5 & 65.5 & 58.49 & 59.18 & 32.04 \\
  \ha +\textbf{\our}~(0.5 + 0.3)      & 13.54 & 20.59 & 28.24 & 39.69 & 26.01 & 14.07 & 17.63 & 19.36 & 17.87 & 21.0 & 67.69 & 36.12 & 7.5 & 67.5 & 56.02 & 56.04 & 31.8 \\
 \hx +\textbf{\our}~(0.5 + 0.5)      & 13.68 & 20.28 & 29.95 & 40.44 & 26.34 & 13.16 & 17.4 & 19.38 & 17.43 & 21.0 & 71.53 & 36.23 & 7.5 & 66.5 & 55.77 & 55.88 & 32.03 \\
 \hd +\textbf{\our}~(0.8)    & 13.82 & 20.28 & 28.63 & 40.84 & 26.75 & 14.25 & 17.29 & 19.06 & 17.23 & 22.0 & 57.2 & 35.41 & 7.0 & 64.0 & 57.2 & 57.61 & 31.16 \\
   \hb +\textbf{\our}~(0.8 + 0.3)      & 13.81 & 19.48 & 29.32 & 41.51 & 24.07 & 14.96 & 17.19 & 19.05 & 17.72 & 21.0 & 59.72 & 35.69 & 6.5 & 65.0 & 54.43 & 55.17 & 30.91 \\
 \he +\textbf{\our}~(0.8 + 0.8)      & 13.39 & 19.41 & 29.28 & 37.07 & 24.53 & 11.78 & 16.06 & 18.68 & 15.61 & 18.0 & 63.54 & 31.52 & 6.09 & 63.5 & 49.85 & 52.04 & 29.4 \\
 \cdashline{2-18}
 PyramidKV        &  21.79 & 44.6 & 45.96 & 48.33 & 43.63 & 25.82 & 30.42 & 22.45 & 27.05 & 72.0 & 88.69 & 41.59 & 6.0 & 71.5 & 62.21 & 48.72 & 43.8 \\
 \hc +\textbf{\our}~(0.5)    &  22.66 & 43.95 & 45.82 & 48.33 & 43.85 & 24.85 & 30.16 & 22.76 & 26.84 & 70.0 & 88.34 & 41.4 & 6.5 & 71.5 & 62.83 & 51.15 & 43.81 \\
 \ha +\textbf{\our}~(0.5 + 0.3)      & 21.97 & 43.67 & 45.69 & 48.83 & 44.04 & 26.49 & 30.09 & 22.76 & 26.98 & 71.5 & 88.13 & 41.86 & 5.5 & 72.0 & 61.99 & 51.83 & 43.96 \\
 \hd +\textbf{\our}~(0.8)    & 22.44 & 44.2 & 44.62 & 46.29 & 40.37 & 22.68 & 27.83 & 22.56 & 25.67 & 69.0 & 84.2 & 40.17 & 5.5 & 72.0 & 60.38 & 41.98 & 41.87 \\
 \hb +\textbf{\our}~(0.8 + 0.3)      & 23.59 & 42.82 & 46.72 & 46.89 & 40.38 & 22.66 & 28.4 & 22.82 & 26.38 & 67.5 & 88.49 & 39.96 & 6.0 & 72.0 & 60.89 & 44.66 & 42.51 \\
 \he +\textbf{\our}~(0.8 + 0.8)      & 19.0 & 37.38 & 45.83 & 42.05 & 34.14 & 19.39 & 23.04 & 21.73 & 24.01 & 64.0 & 87.98 & 36.42 & 3.83 & 70.5 & 56.49 & 57.01 & 40.17 \\

 \cmidrule{1-18}
  \multicolumn{18}{c}{LLaMA-3-8B-Instruct, KV-size 512}\\
 SnapKV           & 19.24 & 36.51 & 43.61 & 46.83 & 36.62 & 23.11 & 22.62 & 21.17 & 24.03 & 45.0 & 88.59 & 40.09 & 6.0 & 71.5 & 63.75 & 58.65 & 40.46 \\
 \hc +\textbf{\our}~(0.5)      & 18.66 & 36.13 & 43.23 & 46.66 & 36.17 & 22.86 & 22.44 & 21.19 & 23.7 & 42.5 & 89.11 & 40.15 & 6.5 & 71.5 & 63.8 & 59.0 & 40.22 \\
 \ha +\textbf{\our}~(0.5 + 0.3)      & 18.89 & 36.42 & 42.27 & 46.3 & 37.1 & 21.89 & 22.33 & 21.36 & 23.54 & 43.0 & 88.46 & 40.23 & 5.5 & 71.5 & 61.89 & 56.65 & 39.83 \\
 \hx +\textbf{\our}~(0.5 + 0.5)      & 17.66 & 36.29 & 44.12 & 48.07 & 36.33 & 22.72 & 21.7 & 21.48 & 23.07 & 42.5 & 88.53 & 39.29 & 5.0 & 71.5 & 62.74 & 59.51 & 40.03 \\
 \hd +\textbf{\our}~(0.8)      & 18.23 & 37.34 & 42.42 & 44.71 & 34.85 & 23.14 & 21.8 & 21.26 & 23.68 & 41.5 & 87.22 & 38.88 & 5.0 & 72.5 & 62.86 & 55.01 & 39.40 \\
  \hb +\textbf{\our}~(0.8 + 0.3)      & 18.02 & 35.92 & 42.88 & 44.84 & 33.93 & 23.64 & 21.14 & 21.34 & 23.43 & 41.5 & 87.4 & 38.57 & 5.0 & 72.0 & 61.4 & 53.22 & 39.01 \\
 \he +\textbf{\our}~(0.8 + 0.8)      & 16.08 & 28.99 & 41.35 & 40.58 & 30.7 & 20.85 & 19.55 & 20.88 & 21.15 & 34.5 & 85.91 & 35.8 & 3.9 & 68.5 & 54.49 & 57.92 & 36.32 \\
  \cdashline{2-18}
 PyramidKV        & 21.79 & 44.6 & 45.96 & 48.33 & 43.63 & 25.82 & 30.42 & 22.45 & 26.96 & 72.0 & 88.69 & 41.59 & 6.0 & 71.5 & 62.21 & 48.72 & 43.79 \\
 \hc +\textbf{\our}~(0.5)    &  22.79 & 43.99 & 45.63 & 48.83 & 43.64 & 24.87 & 30.34 & 22.89 & 26.57 & 70.0 & 88.75 & 42.28 & 6.5 & 71.5 & 62.72 & 50.81 & 43.88 \\
 \ha +\textbf{\our}~(0.5 + 0.3)      & 21.92 & 43.78 & 45.89 & 49.33 & 43.54 & 26.29 & 29.92 & 22.73 & 26.85 & 71.5 & 88.08 & 41.57 & 5.5 & 71.5 & 62.03 & 52.26 & 43.92 \\
 \hd +\textbf{\our}~(0.8)    & 22.73 & 44.1 & 47.2 & 46.47 & 40.51 & 22.81 & 26.66 & 22.72 & 24.87 & 68.0 & 88.63 & 40.44 & 5.5 & 72.0 & 59.61 & 42.44 & 42.17 \\
 \hb +\textbf{\our}~(0.8 + 0.3)      & 22.87 & 42.99 & 46.23 & 46.99 & 40.03 & 23.15 & 28.06 & 22.76 & 26.0 & 67.5 & 88.8 & 39.92 & 5.5 & 72.5 & 60.89 & 44.56 & 42.42 \\
 \he +\textbf{\our}~(0.8 + 0.8)      & 19.01 & 36.61 & 46.44 & 41.91 & 34.64 & 19.24 & 23.13 & 21.66 & 24.02 & 64.0 & 87.92 & 36.19 & 3.33 & 70.5 & 56.29 & 55.75 & 40.04 \\

 \cmidrule{1-18}
  \multicolumn{18}{c}{LLaMA-3-8B-Instruct, KV-size 1024}\\
 SnapKV                  & 21.39 & 39.89 & 44.54 & 48.78 & 43.51 & 23.76 & 24.61 & 21.92 & 25.64 & 55.5 & 88.51 & 40.79 & 6.0 & 72.5 & 63.76 & 56.05 & 42.32 \\
 \hc +\textbf{\our}~(0.5)    & 21.9 & 38.92 & 45.22 & 48.69 & 41.27 & 24.25 & 24.65 & 21.92 & 25.88 & 55.0 & 88.8 & 41.22 & 6.5 & 72.0 & 63.43 & 56.22 & 42.24 \\
  \ha +\textbf{\our}~(0.5 + 0.3)      & 20.13 & 38.5 & 43.06 & 47.54 & 41.82 & 24.36 & 24.13 & 21.03 & 25.62 & 54.5 & 85.73 & 40.45 & 6.0 & 71.5 & 61.4 & 53.17 & 41.18 \\
 \hx +\textbf{\our}~(0.5 + 0.5)      & 21.28 & 38.17 & 45.04 & 46.17 & 37.97 & 24.34 & 22.81 & 20.15 & 24.96 & 50.5 & 85.45 & 39.3 & 6.0 & 70.0 & 59.58 & 52.92 & 40.29 \\
 \hd +\textbf{\our}~(0.8)    & 21.26 & 39.65 & 45.48 & 46.93 & 38.85 & 22.84 & 23.98 & 21.94 & 25.37 & 54.0 & 87.93 & 39.34 & 5.0 & 72.0 & 63.66 & 51.97 & 41.26 \\
 \hb +\textbf{\our}~(0.8 + 0.3)      &  21.02 & 37.3 & 46.12 & 44.15 & 36.79 & 22.97 & 22.31 & 20.59 & 24.54 & 51.0 & 85.41 & 38.89 & 4.5 & 73.5 & 58.04 & 48.65 & 39.74 \\
 \he +\textbf{\our}~(0.8 + 0.8)      & 15.74 & 32.07 & 40.23 & 34.54 & 33.89 & 19.33 & 20.08 & 19.54 & 21.92 & 46.5 & 79.37 & 33.1 & 3.9 & 61.5 & 52.68 & 53.42 & 35.49 \\
   \cdashline{2-18}
 PyramidKV        & 21.79 & 44.6 & 46.0 & 48.33 & 43.63 & 25.82 & 30.42 & 22.45 & 26.53 & 72.0 & 88.69 & 41.59 & 6.0 & 71.5 & 61.87 & 48.72 & 43.75 \\
 \hc +\textbf{\our}~(0.5)    &  22.53 & 43.84 & 45.97 & 47.83 & 43.64 & 24.87 & 30.06 & 22.9 & 26.82 & 70.0 & 89.28 & 41.87 & 6.5 & 71.5 & 61.4 & 50.84 & 43.74 \\
 \ha +\textbf{\our}~(0.5 + 0.3)      & 21.78 & 43.49 & 45.99 & 49.66 & 43.46 & 26.32 & 30.07 & 22.6 & 26.49 & 71.5 & 88.14 & 41.92 & 5.5 & 72.0 & 61.96 & 51.76 & 43.92 \\
 \hd +\textbf{\our}~(0.8)    & 22.59 & 44.35 & 47.66 & 47.13 & 39.96 & 22.94 & 28.04 & 22.68 & 25.37 & 68.5 & 88.65 & 40.62 & 5.5 & 72.5 & 57.89 & 43.28 & 42.35 \\
 \hb +\textbf{\our}~(0.8 + 0.3)      & 23.19 & 43.03 & 47.13 & 46.21 & 40.3 & 23.47 & 28.54 & 22.59 & 25.94 & 68.0 & 87.99 & 40.34 & 6.0 & 72.5 & 60.24 & 44.3 & 42.49 \\
 \he +\textbf{\our}~(0.8 + 0.8)      & 19.36 & 36.94 & 45.03 & 41.86 & 33.47 & 20.29 & 23.05 & 21.75 & 23.47 & 64.5 & 88.05 & 36.09 & 3.33 & 69.0 & 55.15 & 56.56 & 39.87 \\

  \cmidrule{1-18}
  \multicolumn{18}{c}{LLaMA-3-8B-Instruct, KV-size 2048}\\
 SnapKV                  & 22.66 & 41.71 & 46.74 & 48.86 & 43.68 & 23.76 & 27.09 & 22.39 & 27.28 & 62.0 & 88.3 & 41.45 & 6.0 & 72.0 & 63.64 & 53.8 & 43.21 \\
 \hc +\textbf{\our}~(0.5)    & 22.98 & 40.11 & 46.65 & 48.86 & 42.48 & 23.97 & 27.24 & 22.27 & 26.99 & 61.5 & 88.69 & 41.45 & 6.5 & 72.0 & 63.57 & 54.77 & 43.13 \\
 \ha +\textbf{\our}~(0.5 + 0.3)      & 23.06 & 40.75 & 45.87 & 49.11 & 43.43 & 24.92 & 26.91 & 22.15 & 27.05 & 60.0 & 87.56 & 40.94 & 6.0 & 71.5 & 61.9 & 53.97 & 42.82 \\
 \hx +\textbf{\our}~(0.5 + 0.5)      & 22.82 & 41.78 & 46.32 & 47.99 & 40.56 & 23.86 & 26.16 & 22.32 & 26.79 & 60.5 & 88.99 & 40.94 & 7.0 & 71.5 & 62.52 & 56.11 & 42.88 \\
 \hd +\textbf{\our}~(0.8)    & 23.65 & 41.91 & 46.59 & 47.13 & 41.33 & 21.84 & 25.99 & 22.58 & 26.81 & 59.0 & 88.04 & 39.62 & 5.0 & 72.5 & 62.74 & 48.69 & 42.09 \\
 \hb +\textbf{\our}~(0.8 + 0.3)      &  21.88 & 41.94 & 46.18 & 47.17 & 38.95 & 22.62 & 25.41 & 22.29 & 25.74 & 59.0 & 88.51 & 39.96 & 4.5 & 72.5 & 61.22 & 47.0 & 41.55 \\
 \he +\textbf{\our}~(0.8 + 0.8)      & 18.68 & 34.6 & 45.04 & 42.42 & 35.95 & 20.52 & 22.27 & 21.58 & 23.69 & 61.0 & 88.54 & 34.3 & 3.33 & 70.0 & 57.18 & 57.6 & 39.79 \\

    \cdashline{2-18}
 PyramidKV        & 23.7 & 42.37 & 45.43 & 48.7 & 43.73 & 22.86 & 26.65 & 22.16 & 26.73 & 60.5 & 88.44 & 41.36 & 6.0 & 72.0 & 61.91 & 50.23 & 42.67 \\
 \hc +\textbf{\our}~(0.5)    &  23.3 & 40.47 & 41.47 & 47.72 & 43.38 & 23.49 & 25.5 & 21.85 & 25.49 & 59.5 & 88.02 & 41.41 & 6.0 & 71.0 & 61.79 & 51.52 & 41.99 \\
 \ha +\textbf{\our}~(0.5 + 0.3)      & 22.89 & 41.54 & 46.6 & 48.37 & 42.98 & 21.75 & 26.11 & 22.14 & 26.78 & 60.0 & 87.85 & 41.38 & 5.5 & 71.0 & 62.14 & 53.62 & 42.54 \\
 \hd +\textbf{\our}~(0.8)    & 21.93 & 41.48 & 45.24 & 46.2 & 41.5 & 22.38 & 23.08 & 21.74 & 25.83 & 55.5 & 86.79 & 39.72 & 5.0 & 72.5 & 59.55 & 44.35 & 40.8 \\
 \hb +\textbf{\our}~(0.8 + 0.3)      &  21.0 & 40.12 & 46.12 & 47.48 & 39.06 & 19.7 & 24.9 & 22.09 & 26.38 & 58.5 & 87.91 & 40.17 & 4.5 & 71.5 & 61.23 & 47.19 & 41.12 \\
 \he +\textbf{\our}~(0.8 + 0.8)      & 18.15 & 34.74 & 46.88 & 42.72 & 35.85 & 18.39 & 21.42 & 21.32 & 23.48 & 57.5 & 86.82 & 35.43 & 3.33 & 68.5 & 56.39 & 56.29 & 39.2 \\
 \bottomrule
    \end{tabular}}
     \caption{Performance comparison of pruning both K and V cache on LLaMA-3-8B-Instruct at LongBench. \textbf{\our}~($\lambda_k$+ $\lambda_v$) denote the channel-wise key cache pruning ratio $\lambda_k$ and value cache pruning ratio $\lambda_v$. \textbf{\our}~($\lambda$) denote the channel-wise key cache pruning ratio $\lambda$.}
     \label{tab:app_kv_llama3-8b}
    \end{table*}

Unlike keys, value vectors cannot be assessed using the query for their relative importance, which makes structured pruning strategies such as those used in \think{} \citep{think} less suitable. To address this limitation, we adopt an unstructured sparsity approach that better aligns with the distributional characteristics of value channels.

Specifically, for each token $t$, we estimate the importance score of each value channel $\boldsymbol{v}_{i,t}^j$ denotes the $j$-th channel at head $i$. This norm-based scoring captures per-channel activation strength, allowing us to identify and prune the least informative dimensions in a fine-grained manner.
We then apply the same masking and recovery mechanism as in key pruning: pruned channels are removed from the cache, and only the top-$T$ channels (according to the norm) are retained. 
Unlike key recovery, value recovery requires no additional operations such as scaling or recombination, as values are directly consumed in the final weighted sum. This greatly simplifies the recovery process and reduces runtime overhead. The full results on LLaMA-3-8B-Instruct are in the Table \ref{tab:app_kv_llama3-8b}.

While this norm-based criterion offers a practical and lightweight solution, it does not fully capture the semantics of value representations. We leave the exploration of more sophisticated pruning strategies—potentially leveraging attention weights, value-token correlations, or dynamic token importance—for future work.

\section{Limitations}

\noindent \textbf{Increased Computational Overhead.}
Although our recovery mechanism enables accurate reconstruction of pruned channels, it inevitably introduces additional computations during attention score estimation. This overhead, while lightweight in steady-state throughput, contributes to increased Time-To-First-Token (TTFT), particularly in low-latency applications or systems with stringent serving constraints.

\noindent \textbf{Limited Gains on Short Inputs.}
Our method is primarily designed to improve efficiency under long input sequences and large KV cache budgets. In contrast, for short inputs (e.g., $\ leq$4k tokens), the memory footprint is already minimal, and the overhead introduced by dynamic channel scoring and recovery may outweigh the benefits. In such cases, static caching or lightweight token-eviction strategies may offer better latency-efficiency trade-offs.

\noindent \textbf{Heuristic-Based Value Pruning.}
While our channel-wise pruning for the key cache is guided by query-aware saliency, the value cache pruning currently relies on simple norm-based heuristics. This limits its ability to fully exploit the semantic structure of value representations. Future work could explore task- or position-adaptive value pruning strategies.

\section{Extended Results}
\label{app:extended}

\begin{figure}[t]
\centerline{\includegraphics[width=0.5\textwidth]{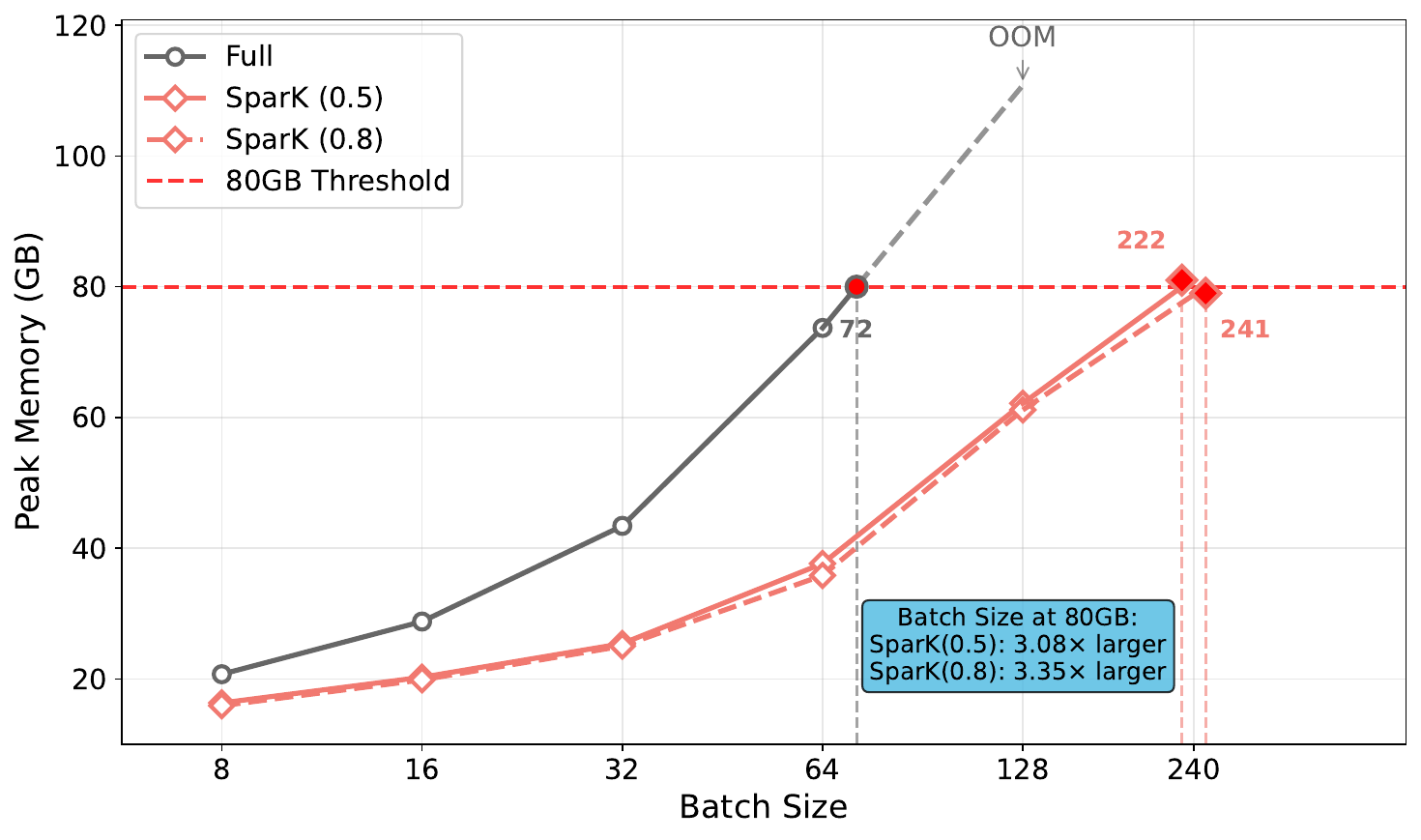}}
\caption{\textbf{Batch Size and Memory.} \our{} enables a 300\% larger batch size, saving more than 20GB memory.
}
\label{fig:app_bsz_mem}
\end{figure}

\subsection{Memory Efficiency Analysis}

To further assess the memory efficiency of our method, we conduct a peak memory usage analysis under varying batch sizes using the LLaMA-3.1-8B-Instruct model. We compare the full KV cache baseline with \our{} under different pruning ratios (0.5 and 0.8). Results are summarized in Figure~\ref{fig:app_bsz_mem}.

\noindent \textbf{Substantial Memory Reduction.} Across all batch sizes, \our{} consistently reduces peak memory consumption compared to the full KV cache.
At batch size 32, the full cache consumes 43.41 GB, while \our{} (0.5) and \our{} (0.8) reduce it to 25.41 GB and 25.02 GB, respectively.
At batch size 64, memory drops from 73.69 GB to 37.65 GB (\our{}-0.5) and 35.83 GB (\our{}-0.8), indicating a >50\% reduction.

\noindent \textbf{Scalability under Memory Constraints.}
We additionally measure the maximum supported batch size under an 80GB memory cap:
Full KV cache supports only 72 batch sizes,
\our{} (0.5) supports up to 222 batch sizes,
\our{} (0.8) supports up to 241 batch sizes.
This highlights \our{}’s effectiveness in enabling larger batch inference under fixed hardware budgets, improving throughput by over 3× without sacrificing quality.

\subsection{Longbench}
\label{app:ex_longbench}

To further validate the generality and robustness of our method, we conduct extensive experiments on the LongBench benchmark across multiple open-source LLMs with varying model scales and instruction-following capabilities. Specifically, Table~\ref{tab:full_llama3-8b} presents results on LLaMA3-8B-Instruct, while Tables~\ref{tab:full_llama3.1-8b}, \ref{tab:full_llama3.1-70b}, \ref{tab:full_qwen3-8b}, and\ref{tab:full_qwen3-32b} extend the evaluation to LLaMA3.1-8B, LLaMA3.1-70B,Qwen3-8B, and Qwen3-32B, respectively.

\begin{table*}[t!]
    \centering \scriptsize

\resizebox{\textwidth}{!}{\begin{tabular}
{C{1em}l@{\hspace{0.05ex}}C{3.8em}@{\hspace{0.05ex}}C{3.8em}@{\hspace{0.05ex}}C{3.8em}@{\hspace{0.05ex}}c@{\hspace{0.05ex}}c@{\hspace{0.05ex}}C{3.8em}@{\hspace{0.05ex}}c@{\hspace{0.05ex}}C{3.8em}@{\hspace{0.05ex}}c@{\hspace{0.05ex}}C{3.8em}@{\hspace{0.05ex}}c@{\hspace{0.05ex}}c@{\hspace{0.05ex}}c@{\hspace{0.6ex}}C{3.8em}@{\hspace{0.6ex}}C{3.8em}@{\hspace{0.6ex}}C{3.8em}@{\hspace{0.6ex}}c}
    \toprule
& \multirow{4}{*}{\textbf{Method}}& \multicolumn{3}{c}{\textbf{Single-Document QA}} & \multicolumn{3}{c}{\textbf{Multi-Document QA}}& \multicolumn{3}{c}{\textbf{Summarization}}& \multicolumn{3}{c}{\textbf{Few-shot Learning}}& \multicolumn{2}{c}{\textbf{Synthetic}} & \multicolumn{2}{c}{\textbf{Code}}&\multirow{4}{*}{\textbf{Avg.}} \\
& & \rotatebox[origin=c]{30}{\bf NrtvQA} & \rotatebox[origin=c]{30}{\bf Qasper} & \rotatebox[origin=c]{30}{\bf MF-en} & \rotatebox[origin=c]{30}{\bf HotpotQA} & \rotatebox[origin=c]{30}{\bf 2WikiMQA} & \rotatebox[origin=c]{30}{\bf Musique} & \rotatebox[origin=c]{30}{\bf GovReport} & \rotatebox[origin=c]{30}{\bf QMSum} & \rotatebox[origin=c]{30}{\bf MultiNews} & \rotatebox[origin=c]{30}{\bf TREC} & \rotatebox[origin=c]{30}{\bf TriviaQA} & \rotatebox[origin=c]{30}{\bf SAMSum} & \rotatebox[origin=c]{30}{\bf PCount} & \rotatebox[origin=c]{30}{~\bf PRe~~} & \rotatebox[origin=c]{30}{~\bf Lcc~~} & \rotatebox[origin=c]{30}{~\bf RB-P~} \\
\cmidrule{1-19}

 - & Vanilla & 22.48 & 44.72 & 46.23 & 48.49 & 44.71 & 24.43 & 30.7 & 22.8 & 27.28 & 72.0 & 88.35 & 42.28 & 6.5 & 72.0 & 63.61 & 51.67 & 44.27 \\
\cmidrule{1-19}
 \multirow{13}*{\raisebox{10.0em}{\rotatebox{90}{KV-size 128}}}
 &   StreamingLLM  & 13.64 & 18.03 & 17.79 & 31.36 & 27.46 & 8.67 & 17.31 & 18.99 & 17.87 & 31.0 & 31.21 & 35.71 & 1.5 & 67.5 & 56.63 & 55.16 & 28.11 \\
 &   ExpectedAttention  & 17.32 & 24.08 & 23.87 & 38.76 & 26.43 & 12.55 & 22.26 & 20.81 & 23.57 & 20.5 & 77.22 & 36.59 & 5.5 & 62.5 & 52.78 & 46.45 & 31.95 \\
 &  TOVA & 17.09 & 23.35 & 37.88 & 43.32 & 28.68 & 15.85 & 19.87 & 20.54 & 18.51 & 26.5 & 85.18 & 39.15 & 4.0 & 60.5 & 59.98 & 57.17 & 34.85 \\
  \cdashline{2-19}
 &   SnapKV &  15.29 & 20.03 & 29.2 & 39.92 & 28.26 & 15.06 & 17.74 & 19.27 & 18.05 & 21.0 & 68.64 & 36.64 & 6.0 & 66.0 & 57.86 & 59.08 & 32.38  \\
 &   +\think~(0.5)    & 13.6 & 19.2 & 31.78 & 36.24 & 25.05 & 11.92 & 16.85 & 19.17 & 16.4 & 2.0 & 50.73 & 32.8 & 6.0 & 65.0 & 52.29 & 52.11 & 28.20 \\
 &   +\think~(0.8)    & 7.6 & 7.03 & 17.45 & 19.98 & 9.8 & 6.9 & 14.37 & 14.15 & 12.5 & 0.0 & 21.36 & 11.22 & 1.02 & 63.0 & 30.42 & 34.69 & 16.97 \\
 \hc &   +\textbf{\our}~(0.5)    & 13.52 & 20.19 & 29.28 & 38.77 & 26.33 & 14.44 & 17.66 & 19.12 & 17.98 & 21.0 & 68.95 & 36.66 & 5.5 & 65.5 & 58.49 & 59.18 & 32.04 \\
 \hd &   +\textbf{\our}~(0.8)    & 13.82 & 20.28 & 28.63 & 40.84 & 26.75 & 14.25 & 17.29 & 19.06 & 17.23 & 22.0 & 57.2 & 35.41 & 7.0 & 64.0 & 57.2 & 57.61 & 31.16 \\
 \cdashline{2-19}
 &   PyramidKV        &  21.79 & 44.6 & 45.96 & 48.33 & 43.63 & 25.82 & 30.42 & 22.45 & 27.05 & 72.0 & 88.69 & 41.59 & 6.0 & 71.5 & 62.21 & 48.72 & 43.8 \\
 &   +\think~(0.5)    & 22.48 & 40.56 & 47.94 & 45.83 & 34.95 & 23.19 & 27.55 & 22.54 & 25.73 & 53.5 & 84.88 & 32.7 & 7.78 & 71.0 & 53.9 & 51.54 & 40.38 \\
 &   +\think~(0.8)    &  6.37 & 5.53 & 13.73 & 12.53 & 5.47 & 3.16 & 16.97 & 14.21 & 17.02 & 0.0 & 23.03 & 7.54 & 1.73 & 13.0 & 29.67 & 27.51 & 12.34\\
 \hc &   +\textbf{\our}~(0.5)    &  22.66 & 43.95 & 45.82 & 48.33 & 43.85 & 24.85 & 30.16 & 22.76 & 26.84 & 70.0 & 88.34 & 41.4 & 6.5 & 71.5 & 62.83 & 51.15 & 43.81 \\
 \hd &   +\textbf{\our}~(0.8)    & 22.44 & 44.2 & 44.62 & 46.29 & 40.37 & 22.68 & 27.83 & 22.56 & 25.67 & 69.0 & 84.2 & 40.17 & 5.5 & 72.0 & 60.38 & 41.98 & 41.87 \\

 \cmidrule{1-19}
  \multirow{13}*{\raisebox{10.0em}{\rotatebox{90}{KV-size 512}}}
  &   StreamingLLM  & 13.98 & 23.72 & 20.26 & 35.82 & 29.76 & 11.34 & 22.12 & 19.56 & 24.49 & 45.0 & 54.98 & 38.32 & 4.5 & 67.0 & 58.16 & 52.63 & 32.6 \\
 &   ExpectedAttention  & 19.73 & 33.41 & 30.2 & 45.06 & 32.81 & 20.43 & 25.55 & 21.45 & 26.25 & 51.0 & 85.76 & 39.57 & 6.0 & 56.0 & 62.0 & 54.84 & 38.13 \\
 &  TOVA & 18.84 & 33.46 & 44.0 & 48.36 & 36.82 & 21.47 & 23.07 & 20.72 & 24.33 & 63.0 & 88.91 & 41.01 & 6.0 & 71.0 & 64.66 & 58.33 & 41.5 \\
  \cdashline{2-19}
 &   SnapKV           & 19.24 & 36.51 & 43.61 & 46.83 & 36.62 & 23.11 & 22.62 & 21.17 & 24.03 & 45.0 & 88.59 & 40.09 & 6.0 & 71.5 & 63.75 & 58.65 & 40.46 \\
 &   +\think~(0.5)    & 18.73 & 33.83 & 41.47 & 43.72 & 27.98 & 20.91 & 20.59 & 21.56 & 22.25 & 15.5 & 84.62 & 33.82 & 7.0 & 71.5 & 57.01 & 56.97 & 36.09 \\
 &   +\think~(0.8)    & 9.48 & 6.59 & 18.62 & 18.28 & 8.32 & 9.2 & 17.11 & 15.37 & 16.46 & 0.0 & 43.94 & 8.6 & 2.21 & 34.62 & 33.43 & 35.47 & 17.36 \\
 \hc &   +\textbf{\our}~(0.5)      & 18.66 & 36.13 & 43.23 & 46.66 & 36.17 & 22.86 & 22.44 & 21.19 & 23.7 & 42.5 & 89.11 & 40.15 & 6.5 & 71.5 & 63.8 & 59.0 & 40.22 \\
 \hd &   +\textbf{\our}~(0.8)      & 18.23 & 37.34 & 42.42 & 44.71 & 34.85 & 23.14 & 21.8 & 21.26 & 23.68 & 41.5 & 87.22 & 38.88 & 5.0 & 72.5 & 62.86 & 55.01 & 39.40 \\
  \cdashline{2-19}
 &   PyramidKV        & 21.79 & 44.6 & 45.96 & 48.33 & 43.63 & 25.82 & 30.42 & 22.45 & 26.96 & 72.0 & 88.69 & 41.59 & 6.0 & 71.5 & 62.21 & 48.72 & 43.79 \\
 &   +\think~(0.5)    & 22.48 & 40.56 & 47.94 & 45.83 & 34.95 & 23.19 & 27.55 & 22.54 & 25.6 & 53.5 & 84.88 & 32.7 & 7.78 & 71.0 & 53.9 & 51.54 & 40.37 \\
 &   +\think~(0.8)    &  6.37 & 5.53 & 13.73 & 12.53 & 5.47 & 3.16 & 16.97 & 14.21 & 17.11 & 0.0 & 23.03 & 7.54 & 1.73 & 13.0 & 29.67 & 27.51 & 12.35\\
 \hc &   +\textbf{\our}~(0.5)    &  22.79 & 43.99 & 45.63 & 48.83 & 43.64 & 24.87 & 30.34 & 22.89 & 26.57 & 70.0 & 88.75 & 42.28 & 6.5 & 71.5 & 62.72 & 50.81 & 43.88 \\
 \hd &   +\textbf{\our}~(0.8)    & 22.73 & 44.1 & 47.2 & 46.47 & 40.51 & 22.81 & 26.66 & 22.72 & 24.87 & 68.0 & 88.63 & 40.44 & 5.5 & 72.0 & 59.61 & 42.44 & 42.17 \\

 \cmidrule{1-19}
  \multirow{13}*{\raisebox{10.0em}{\rotatebox{90}{KV-size 1024}}} 
  &   StreamingLLM  & 18.05 & 28.35 & 25.3 & 38.35 & 31.0 & 12.31 & 24.1 & 20.26 & 25.92 & 52.5 & 71.87 & 38.91 & 5.5 & 61.5 & 55.89 & 48.71 & 34.91 \\
 &   ExpectedAttention  & 21.06 & 36.69 & 37.86 & 45.76 & 35.36 & 22.08 & 26.59 & 21.62 & 26.76 & 64.5 & 89.64 & 40.36 & 5.5 & 62.0 & 63.79 & 55.67 & 40.95 \\
 &  TOVA & 20.78 & 37.49 & 46.34 & 48.92 & 41.96 & 21.91 & 25.15 & 21.72 & 26.36 & 69.0 & 89.33 & 41.83 & 7.0 & 71.5 & 64.13 & 57.03 & 43.15 \\
  \cdashline{2-19}
 &   SnapKV                  & 21.39 & 39.89 & 44.54 & 48.78 & 43.51 & 23.76 & 24.61 & 21.92 & 25.64 & 55.5 & 88.51 & 40.79 & 6.0 & 72.5 & 63.76 & 56.05 & 42.32 \\
 &   +\think~(0.5)           & 19.44 & 38.4 & 45.16 & 46.3 & 32.01 & 21.18 & 22.4 & 21.88 & 24.43 & 30.5 & 85.45 & 33.98 & 7.0 & 72.0 & 57.09 & 55.86 & 38.32 \\
 &   +\think~(0.8)           & 7.97 & 6.08 & 17.09 & 16.7 & 6.41 & 7.23 & 17.41 & 15.33 & 16.75 & 0.0 & 38.44 & 8.12 & 1.64 & 22.14 & 33.76 & 34.42 & 15.59 \\
 \hc &   +\textbf{\our}~(0.5)    & 21.9 & 38.92 & 45.22 & 48.69 & 41.27 & 24.25 & 24.65 & 21.92 & 25.88 & 55.0 & 88.8 & 41.22 & 6.5 & 72.0 & 63.43 & 56.22 & 42.24 \\
 \hd &   +\textbf{\our}~(0.8)    & 21.26 & 39.65 & 45.48 & 46.93 & 38.85 & 22.84 & 23.98 & 21.94 & 25.37 & 54.0 & 87.93 & 39.34 & 5.0 & 72.0 & 63.66 & 51.97 & 41.26 \\
   \cdashline{2-19}
 &   PyramidKV        & 21.79 & 44.6 & 46.0 & 48.33 & 43.63 & 25.82 & 30.42 & 22.45 & 26.53 & 72.0 & 88.69 & 41.59 & 6.0 & 71.5 & 61.87 & 48.72 & 43.75 \\
 &   +\think~(0.5)    & 22.48 & 40.56 & 47.78 & 45.83 & 34.95 & 23.19 & 27.55 & 22.54 & 25.25 & 53.5 & 84.49 & 32.58 & 7.78 & 71.0 & 54.33 & 51.54 & 40.33 \\
 &   +\think~(0.8)    &  6.37 & 5.53 & 13.75 & 12.53 & 5.44 & 3.16 & 16.97 & 14.21 & 16.88 & 0.0 & 23.03 & 7.55 & 1.73 & 13.0 & 29.59 & 27.51 & 12.33\\
 \hc &   +\textbf{\our}~(0.5)    &  22.53 & 43.84 & 45.97 & 47.83 & 43.64 & 24.87 & 30.06 & 22.9 & 26.82 & 70.0 & 89.28 & 41.87 & 6.5 & 71.5 & 61.4 & 50.84 & 43.74 \\
 \hd &   +\textbf{\our}~(0.8)    & 22.59 & 44.35 & 47.66 & 47.13 & 39.96 & 22.94 & 28.04 & 22.68 & 25.37 & 68.5 & 88.65 & 40.62 & 5.5 & 72.5 & 57.89 & 43.28 & 42.35 \\

  \cmidrule{1-19}
  \multirow{13}*{\raisebox{10.0em}{\rotatebox{90}{KV-size 2048}}} 
    &   StreamingLLM  & 20.21 & 38.11 & 28.47 & 39.22 & 38.22 & 16.87 & 26.69 & 20.83 & 26.97 & 65.0 & 85.11 & 39.93 & 5.0 & 52.5 & 59.55 & 45.66 & 38.02 \\
 &   ExpectedAttention  & 23.0 & 41.33 & 43.55 & 47.73 & 40.37 & 21.23 & 28.21 & 21.77 & 27.38 & 68.5 & 89.41 & 40.75 & 8.5 & 63.5 & 63.84 & 54.96 & 42.75 \\
 &  TOVA & 21.83 & 41.99 & 45.37 & 48.47 & 43.54 & 23.92 & 27.41 & 22.4 & 27.17 & 67.5 & 89.21 & 42.14 & 6.5 & 72.0 & 64.16 & 55.53 & 43.7 \\
  \cdashline{2-19}
 &   SnapKV                  & 22.66 & 41.71 & 46.74 & 48.86 & 43.68 & 23.76 & 27.09 & 22.39 & 27.28 & 62.0 & 88.3 & 41.45 & 6.0 & 72.0 & 63.64 & 53.8 & 43.21 \\
 &   +\think~(0.5)           & 20.06 & 40.36 & 48.02 & 45.55 & 36.94 & 22.3 & 24.23 & 22.33 & 25.74 & 47.5 & 84.41 & 33.97 & 6.14 & 71.5 & 56.41 & 55.49 & 40.06 \\
 &   +\think~(0.8)           & 6.08 & 4.94 & 15.32 & 13.56 & 5.43 & 6.54 & 17.19 & 15.0 & 16.77 & 0.0 & 31.5 & 7.76 & 2.01 & 15.21 & 32.81 & 33.7 & 13.99 \\
 \hc &   +\textbf{\our}~(0.5)    & 22.98 & 40.11 & 46.65 & 48.86 & 42.48 & 23.97 & 27.24 & 22.27 & 26.99 & 61.5 & 88.69 & 41.45 & 6.5 & 72.0 & 63.57 & 54.77 & 43.13 \\
 \hd &   +\textbf{\our}~(0.8)    & 23.65 & 41.91 & 46.59 & 47.13 & 41.33 & 21.84 & 25.99 & 22.58 & 26.81 & 59.0 & 88.04 & 39.62 & 5.0 & 72.5 & 62.74 & 48.69 & 42.09 \\

    \cdashline{2-19}
 &   PyramidKV        & 23.7 & 42.37 & 45.43 & 48.7 & 43.73 & 22.86 & 26.65 & 22.16 & 26.73 & 60.5 & 88.44 & 41.36 & 6.0 & 72.0 & 61.91 & 50.23 & 42.67 \\
 &   +\think~(0.5)    & 21.64 & 39.24 & 45.13 & 44.64 & 36.05 & 22.4 & 24.07 & 22.56 & 26.07 & 40.0 & 84.29 & 32.72 & 8.62 & 71.5 & 54.38 & 52.73 & 39.13 \\
 &   +\think~(0.8)    &  6.55 & 4.91 & 14.13 & 13.94 & 6.65 & 5.54 & 17.1 & 14.68 & 16.96 & 0.0 & 28.27 & 7.91 & 1.73 & 21.6 & 29.64 & 28.33 & 13.62\\
 \hc &   +\textbf{\our}~(0.5)    &  23.3 & 40.47 & 41.47 & 47.72 & 43.38 & 23.49 & 25.5 & 21.85 & 25.49 & 59.5 & 88.02 & 41.41 & 6.0 & 71.0 & 61.79 & 51.52 & 41.99 \\
 \hd &   +\textbf{\our}~(0.8)    & 21.93 & 41.48 & 45.24 & 46.2 & 41.5 & 22.38 & 23.08 & 21.74 & 25.83 & 55.5 & 86.79 & 39.72 & 5.0 & 72.5 & 59.55 & 44.35 & 40.8 \\
 \bottomrule
    \end{tabular}}
     \caption{Performance comparison on LLaMA-3-8B-Instruct at LongBench. \textbf{\our}~($\lambda$) and \think($\lambda$) denote the channel-wise key cache pruning ratio $\lambda$.}
     \label{tab:full_llama3-8b}
    \end{table*}

\begin{table*}[t!]
    \centering \scriptsize

\resizebox{\textwidth}{!}{\begin{tabular}
{C{1em}l@{\hspace{0.05ex}}C{3.8em}@{\hspace{0.05ex}}C{3.8em}@{\hspace{0.05ex}}C{3.8em}@{\hspace{0.05ex}}c@{\hspace{0.05ex}}c@{\hspace{0.05ex}}C{3.8em}@{\hspace{0.05ex}}c@{\hspace{0.05ex}}C{3.8em}@{\hspace{0.05ex}}c@{\hspace{0.05ex}}C{3.8em}@{\hspace{0.05ex}}c@{\hspace{0.05ex}}c@{\hspace{0.05ex}}c@{\hspace{0.6ex}}C{3.8em}@{\hspace{0.6ex}}C{3.8em}@{\hspace{0.6ex}}C{3.8em}@{\hspace{0.6ex}}c}
    \toprule
& \multirow{4}{*}{\textbf{Method}}& \multicolumn{3}{c}{\textbf{Single-Document QA}} & \multicolumn{3}{c}{\textbf{Multi-Document QA}}& \multicolumn{3}{c}{\textbf{Summarization}}& \multicolumn{3}{c}{\textbf{Few-shot Learning}}& \multicolumn{2}{c}{\textbf{Synthetic}} & \multicolumn{2}{c}{\textbf{Code}}&\multirow{4}{*}{\textbf{Avg.}} \\
& & \rotatebox[origin=c]{30}{\bf NrtvQA} & \rotatebox[origin=c]{30}{\bf Qasper} & \rotatebox[origin=c]{30}{\bf MF-en} & \rotatebox[origin=c]{30}{\bf HotpotQA} & \rotatebox[origin=c]{30}{\bf 2WikiMQA} & \rotatebox[origin=c]{30}{\bf Musique} & \rotatebox[origin=c]{30}{\bf GovReport} & \rotatebox[origin=c]{30}{\bf QMSum} & \rotatebox[origin=c]{30}{\bf MultiNews} & \rotatebox[origin=c]{30}{\bf TREC} & \rotatebox[origin=c]{30}{\bf TriviaQA} & \rotatebox[origin=c]{30}{\bf SAMSum} & \rotatebox[origin=c]{30}{\bf PCount} & \rotatebox[origin=c]{30}{~\bf PRe~~} & \rotatebox[origin=c]{30}{~\bf Lcc~~} & \rotatebox[origin=c]{30}{~\bf RB-P~} \\
\cmidrule{1-19}

 - & Vanilla & 30.84 & 47.4 & 56.07 & 59.3 & 50.23 & 32.12 & 34.81 & 24.84 & 27.15 & 72.5 & 81.27 & 44.47 & 11.25 & 100.0 & 64.7 & 58.77 & 49.73 \\
\cmidrule{1-19}
 \multirow{14}*{\raisebox{10.0em}{\rotatebox{90}{KV-size 128}}}
 &   StreamingLLM  & 14.2 & 22.37 & 23.51 & 34.93 & 30.51 & 9.4 & 17.76 & 20.2 & 18.75 & 24.5 & 81.4 & 35.1 & 4.5 & 99.0 & 59.89 & 56.18 & 34.51 \\
 &   ExpectedAttention  & 19.62 & 26.06 & 31.65 & 40.95 & 25.31 & 15.1 & 24.01 & 21.99 & 24.38 & 21.0 & 87.16 & 38.22 & 5.0 & 97.67 & 44.09 & 40.46 & 35.17 \\
 &  TOVA & 26.14 & 26.54 & 46.7 & 47.0 & 33.23 & 18.31 & 21.49 & 21.83 & 20.73 & 37.5 & 86.75 & 40.78 & 4.0 & 89.0 & 61.43 & 58.43 & 39.99 \\
  \cdashline{2-19}
   & AdaSnapKV & 20.3 & 24.27 & 39.46 & 49.14 & 36.34 & 17.8 & 18.1 & 21.04 & 20.26 & 25.0 & 84.45 & 39.12 & 4.5 & 96.5 & 62.11 & 61.76 & 38.76\\
 &   SnapKV &  14.81 & 22.58 & 39.17 & 45.51 & 33.78 & 10.46 & 17.54 & 20.27 & 18.84 & 26.5 & 85.27 & 38.17 & 5.5 & 94.0 & 61.63 & 57.79 & 36.99  \\
 &   +\think~(0.5)    & 15.52 & 20.21 & 37.06 & 41.55 & 32.17 & 12.43 & 16.7 & 20.26 & 18.22 & 11.5 & 74.44 & 33.91 & 3.5 & 94.5 & 51.45 & 49.12 & 33.28 \\
 &   +\think~(0.8)    & 11.24 & 9.91 & 21.63 & 23.92 & 10.32 & 5.27 & 13.04 & 13.95 & 13.92 & 0.0 & 47.82 & 11.94 & 1.38 & 80.89 & 32.36 & 31.77 & 20.59 \\
 \hc &   +\textbf{\our}~(0.5)    & 15.48 & 21.07 & 38.47 & 45.86 & 33.46 & 10.57 & 17.87 & 20.5 & 18.26 & 25.0 & 83.51 & 38.03 & 6.5 & 93.5 & 60.19 & 57.31 & 36.6 \\
 \hd &   +\textbf{\our}~(0.8)    & 13.09 & 15.89 & 25.61 & 34.04 & 23.21 & 8.12 & 12.2 & 14.45 & 12.25 & 16.0 & 66.85 & 25.26 & 3.0 & 71.0 & 37.63 & 44.72 & 26.46 \\
 \cdashline{2-19}
 &   PyramidKV        &  30.9 & 48.14 & 56.19 & 59.16 & 50.73 & 32.56 & 34.74 & 24.82 & 27.18 & 71.0 & 82.32 & 44.89 & 10.83 & 100.0 & 64.65 & 59.33 & 49.84 \\
 &   +\think~(0.5)    & 30.94 & 48.36 & 55.06 & 55.76 & 49.81 & 30.55 & 33.39 & 26.05 & 26.56 & 64.5 & 87.36 & 38.68 & 10.11 & 99.5 & 49.42 & 46.87 & 47.06 \\
 &   +\think~(0.8)    &  11.62 & 7.33 & 12.86 & 17.06 & 6.56 & 6.04 & 17.57 & 15.81 & 16.92 & 0.0 & 52.85 & 8.94 & 2.0 & 61.54 & 28.23 & 27.47 & 18.3\\
 \hc &   +\textbf{\our}~(0.5)    &  30.96 & 47.79 & 56.29 & 59.22 & 50.2 & 32.88 & 34.69 & 24.94 & 26.6 & 71.0 & 81.12 & 44.74 & 13.6 & 99.5 & 62.36 & 55.66 & 49.47 \\
 \hd &   +\textbf{\our}~(0.8)    & 31.46 & 47.78 & 56.8 & 58.88 & 49.68 & 33.01 & 33.07 & 25.32 & 26.28 & 70.0 & 79.88 & 43.23 & 12.38 & 99.0 & 59.15 & 59.05 & 49.06 \\
 \cmidrule{1-19}
 \multirow{13}*{\raisebox{10.0em}{\rotatebox{90}{KV-size 512}}}
 &   StreamingLLM  & 16.84 & 23.76 & 24.7 & 39.54 & 31.5 & 10.49 & 23.39 & 20.51 & 24.13 & 46.0 & 82.56 & 38.24 & 4.5 & 96.5 & 65.78 & 62.7 & 38.2 \\
 &   ExpectedAttention  & 24.6 & 37.46 & 35.47 & 46.79 & 40.86 & 19.36 & 27.5 & 22.13 & 26.28 & 49.5 & 90.42 & 40.63 & 4.5 & 90.5 & 59.19 & 51.2 & 41.65 \\
 &  TOVA & 30.63 & 39.32 & 54.32 & 51.82 & 43.1 & 27.17 & 25.3 & 22.48 & 24.62 & 58.5 & 82.64 & 44.29 & 6.75 & 99.5 & 65.53 & 61.08 & 46.07 \\
  \cdashline{2-19}
   & AdaSnapKV & 27.36 & 39.5 & 52.65 & 57.46 & 48.36 & 28.75 & 24.34 & 23.3 & 24.67 & 46.0 & 82.42 & 41.73 & 9.0 & 99.5 & 66.58 & 62.59 & 45.89\\
 &   SnapKV &  27.74 & 38.03 & 52.23 & 56.96 & 44.97 & 24.94 & 24.05 & 23.26 & 24.29 & 42.0 & 83.1 & 40.92 & 8.0 & 99.5 & 67.21 & 61.42 & 44.91  \\
 &   +\think~(0.5)    & 26.29 & 36.37 & 49.56 & 55.67 & 41.8 & 25.18 & 22.67 & 22.5 & 23.43 & 38.5 & 88.47 & 38.88 & 5.25 & 99.5 & 55.97 & 51.55 & 42.6 \\
 &   +\think~(0.8)    &  15.96 & 12.02 & 20.63 & 28.4 & 5.7 & 12.1 & 16.65 & 16.03 & 16.74 & 0.0 & 64.46 & 10.95 & 3.88 & 90.16 & 33.15 & 32.12 & 23.68\\
 \hd &   +\textbf{\our}~(0.8)    & 24.62 & 32.79 & 37.63 & 50.58 & 36.67 & 24.3 & 20.23 & 18.67 & 19.0 & 34.0 & 67.34 & 32.05 & 6.0 & 75.5 & 51.75 & 51.68 & 36.43 \\
 \cdashline{2-19}
 &   PyramidKV        &  30.9 & 48.14 & 56.19 & 59.16 & 50.73 & 32.56 & 34.74 & 24.82 & 27.09 & 71.0 & 82.32 & 44.89 & 10.83 & 100.0 & 64.65 & 59.33 & 49.83 \\
 &   +\think~(0.5)    & 30.94 & 48.36 & 55.06 & 55.76 & 49.81 & 30.55 & 33.39 & 26.05 & 26.45 & 64.5 & 87.36 & 38.68 & 10.11 & 99.5 & 49.42 & 46.87 & 47.05 \\
 &   +\think~(0.8)    &  11.62 & 7.33 & 12.86 & 17.06 & 6.56 & 6.04 & 17.57 & 15.81 & 16.8 & 0.0 & 52.85 & 8.94 & 2.0 & 61.54 & 28.23 & 27.47 & 18.29\\
 \hc &   +\textbf{\our}~(0.5)    &  30.8 & 47.6 & 56.74 & 59.47 & 50.2 & 33.59 & 34.45 & 25.03 & 26.67 & 71.0 & 82.1 & 44.94 & 14.1 & 99.5 & 62.41 & 56.06 & 49.67 \\
 \hd &   +\textbf{\our}~(0.8)    & 30.88 & 47.52 & 57.03 & 58.54 & 49.19 & 32.03 & 33.39 & 25.48 & 26.06 & 70.0 & 80.55 & 43.21 & 11.12 & 99.5 & 60.21 & 58.0 & 48.92 \\
 \cmidrule{1-19}
 \multirow{13}*{\raisebox{10.0em}{\rotatebox{90}{KV-size 1024}}} 
 &   StreamingLLM  & 17.96 & 29.67 & 30.09 & 41.9 & 33.6 & 12.12 & 26.0 & 20.6 & 25.74 & 53.0 & 85.93 & 40.21 & 5.0 & 91.0 & 66.58 & 62.56 & 40.12 \\
 &   ExpectedAttention  & 26.99 & 40.54 & 41.78 & 51.37 & 43.31 & 22.91 & 29.31 & 22.93 & 26.96 & 54.5 & 90.64 & 42.71 & 5.0 & 95.5 & 64.23 & 56.58 & 44.7 \\
 &  TOVA & 30.72 & 42.18 & 56.18 & 54.63 & 49.84 & 25.16 & 27.68 & 23.1 & 26.21 & 62.5 & 82.18 & 44.19 & 7.75 & 99.5 & 65.55 & 60.57 & 47.37 \\
  \cdashline{2-19}
   & AdaSnapKV & 30.19 & 44.59 & 54.31 & 58.31 & 48.01 & 29.7 & 27.07 & 23.39 & 26.27 & 57.5 & 78.69 & 43.19 & 9.5 & 100.0 & 66.74 & 62.6 & 47.5\\
 &   SnapKV &  30.07 & 43.95 & 55.24 & 57.89 & 48.15 & 28.09 & 26.72 & 23.04 & 25.91 & 58.0 & 78.3 & 41.79 & 10.06 & 99.5 & 66.71 & 60.64 & 47.13  \\
 &   +\think~(0.5)    & 30.34 & 42.13 & 50.49 & 54.82 & 47.24 & 26.57 & 25.27 & 23.12 & 25.23 & 49.5 & 85.55 & 38.52 & 7.56 & 99.5 & 54.59 & 51.42 & 44.49 \\
 &   +\think~(0.8)    &  14.68 & 10.65 & 16.96 & 25.68 & 7.06 & 9.75 & 17.73 & 16.01 & 17.13 & 0.0 & 60.02 & 10.47 & 4.5 & 85.42 & 33.49 & 30.72 & 22.52\\
 \hc &   +\textbf{\our}~(0.5)    & 30.23 & 44.91 & 54.68 & 58.91 & 47.57 & 29.89 & 26.81 & 23.29 & 26.05 & 56.5 & 79.05 & 41.95 & 9.56 & 99.5 & 66.19 & 60.3 & 47.21 \\
 \hd &   +\textbf{\our}~(0.8)    & 30.63 & 43.51 & 54.81 & 57.89 & 46.01 & 28.87 & 25.73 & 23.45 & 25.17 & 54.5 & 81.61 & 40.6 & 8.88 & 99.5 & 65.97 & 61.3 & 46.78 \\
 \cdashline{2-19}
 &   PyramidKV        &  30.9 & 48.14 & 56.32 & 59.16 & 50.73 & 32.56 & 34.74 & 24.82 & 26.91 & 71.0 & 82.32 & 44.9 & 10.83 & 100.0 & 64.12 & 59.33 & 49.8 \\
 &   +\think~(0.5)    & 30.94 & 48.36 & 55.22 & 55.76 & 49.81 & 30.55 & 33.39 & 26.05 & 26.15 & 64.5 & 87.36 & 38.71 & 10.11 & 99.5 & 49.08 & 46.87 & 47.02 \\
 &   +\think~(0.8)    &  11.62 & 7.33 & 13.03 & 17.13 & 5.97 & 6.04 & 17.57 & 15.81 & 16.56 & 0.0 & 52.85 & 8.95 & 2.0 & 61.54 & 27.74 & 27.47 & 18.23\\
 \hc &   +\textbf{\our}~(0.5)    &  31.36 & 47.91 & 56.71 & 59.14 & 50.11 & 33.38 & 34.9 & 25.21 & 25.79 & 70.5 & 79.37 & 44.68 & 13.1 & 99.5 & 60.68 & 55.88 & 49.26 \\
 \hd &   +\textbf{\our}~(0.8)    & 31.1 & 47.72 & 55.73 & 58.1 & 49.31 & 32.15 & 33.15 & 25.48 & 25.12 & 69.5 & 80.96 & 43.59 & 13.38 & 100.0 & 60.34 & 57.83 & 48.97 \\
  \cmidrule{1-19}
 \multirow{13}*{\raisebox{10.0em}{\rotatebox{90}{KV-size 2048}}} 
 &   StreamingLLM  & 19.92 & 39.44 & 33.49 & 45.13 & 43.4 & 16.55 & 28.46 & 21.15 & 26.53 & 59.0 & 88.43 & 41.79 & 4.5 & 92.0 & 67.32 & 65.08 & 43.26 \\
 &   ExpectedAttention  & 30.41 & 44.32 & 48.43 & 58.07 & 48.53 & 27.53 & 31.45 & 23.46 & 26.89 & 61.0 & 89.12 & 43.52 & 7.5 & 96.0 & 64.73 & 59.54 & 47.53 \\
 &  TOVA & 30.43 & 45.82 & 56.74 & 56.11 & 52.25 & 25.83 & 30.44 & 23.97 & 26.94 & 70.5 & 82.04 & 44.93 & 10.33 & 100.0 & 64.86 & 60.3 & 48.84 \\
  \cdashline{2-19}
   & AdaSnapKV & 30.29 & 46.43 & 54.4 & 59.29 & 50.35 & 30.32 & 29.52 & 24.19 & 26.82 & 69.0 & 80.97 & 43.34 & 10.03 & 100.0 & 64.62 & 61.17 & 48.8\\
 &   SnapKV &  31.48 & 46.32 & 54.91 & 58.28 & 50.42 & 29.45 & 29.95 & 24.01 & 26.88 & 64.5 & 81.59 & 43.02 & 10.33 & 99.5 & 64.96 & 60.58 & 48.51  \\
 &   +\think~(0.5)    & 31.9 & 44.43 & 54.34 & 55.17 & 45.31 & 30.32 & 27.82 & 24.09 & 26.45 & 56.5 & 86.94 & 38.41 & 8.2 & 99.5 & 51.15 & 50.9 & 45.71 \\
 &   +\think~(0.8)    &  14.41 & 5.87 & 10.47 & 21.73 & 5.2 & 8.92 & 18.17 & 15.79 & 17.31 & 0.0 & 59.83 & 9.48 & 2.0 & 77.77 & 31.76 & 31.12 & 20.61\\
 \hc &   +\textbf{\our}~(0.5)    & 31.68 & 48.27 & 55.07 & 59.06 & 49.59 & 31.96 & 30.01 & 24.31 & 27.22 & 64.5 & 81.12 & 43.22 & 10.83 & 100.0 & 64.74 & 59.73 & 48.83 \\
 \hd &   +\textbf{\our}~(0.8)    & 31.81 & 46.45 & 56.2 & 57.82 & 46.31 & 32.03 & 28.69 & 24.07 & 26.45 & 62.0 & 82.45 & 42.13 & 10.63 & 100.0 & 65.09 & 60.4 & 48.28 \\
 \cdashline{2-19}
 &   PyramidKV        &  29.98 & 47.38 & 55.54 & 56.59 & 50.78 & 26.76 & 28.93 & 23.74 & 26.58 & 56.5 & 90.64 & 42.69 & 12.0 & 99.5 & 67.13 & 63.77 & 48.66 \\
 &   +\think~(0.5)    & 30.03 & 46.45 & 53.54 & 55.27 & 47.27 & 26.83 & 27.65 & 24.04 & 26.35 & 52.0 & 84.38 & 36.27 & 9.93 & 99.5 & 50.66 & 47.87 & 44.88 \\
 &   +\think~(0.8)    &  13.32 & 8.22 & 14.99 & 22.32 & 6.37 & 8.8 & 17.86 & 16.63 & 16.45 & 0.0 & 57.64 & 9.72 & 2.09 & 81.56 & 28.89 & 29.25 & 20.88\\
 \hc &   +\textbf{\our}~(0.5)    &  22.13 & 46.05 & 56.08 & 57.04 & 50.11 & 28.03 & 28.03 & 23.95 & 26.4 & 54.5 & 79.35 & 42.56 & 11.5 & 99.5 & 60.64 & 57.25 & 46.44 \\
 \hd &   +\textbf{\our}~(0.8)    & 20.87 & 44.67 & 53.79 & 55.64 & 48.57 & 27.47 & 26.05 & 24.04 & 25.36 & 55.0 & 78.33 & 39.78 & 11.44 & 98.5 & 60.86 & 56.25 & 45.41 \\
 \bottomrule
    \end{tabular}}
     \caption{Performance comparison on LLaMA-3.1-8B-Instruct at LongBench.}
     \label{tab:full_llama3.1-8b}
    \end{table*}

\begin{table*}[t!]
    \centering \scriptsize

\resizebox{\textwidth}{!}{\begin{tabular}
{C{1em}l@{\hspace{0.05ex}}C{3.8em}@{\hspace{0.05ex}}C{3.8em}@{\hspace{0.05ex}}C{3.8em}@{\hspace{0.05ex}}c@{\hspace{0.05ex}}c@{\hspace{0.05ex}}C{3.8em}@{\hspace{0.05ex}}c@{\hspace{0.05ex}}C{3.8em}@{\hspace{0.05ex}}c@{\hspace{0.05ex}}C{3.8em}@{\hspace{0.05ex}}c@{\hspace{0.05ex}}c@{\hspace{0.05ex}}c@{\hspace{0.6ex}}C{3.8em}@{\hspace{0.6ex}}C{3.8em}@{\hspace{0.6ex}}C{3.8em}@{\hspace{0.6ex}}c}
    \toprule
& \multirow{4}{*}{\textbf{Method}}& \multicolumn{3}{c}{\textbf{Single-Document QA}} & \multicolumn{3}{c}{\textbf{Multi-Document QA}}& \multicolumn{3}{c}{\textbf{Summarization}}& \multicolumn{3}{c}{\textbf{Few-shot Learning}}& \multicolumn{2}{c}{\textbf{Synthetic}} & \multicolumn{2}{c}{\textbf{Code}}&\multirow{4}{*}{\textbf{Avg.}} \\
& & \rotatebox[origin=c]{30}{\bf NrtvQA} & \rotatebox[origin=c]{30}{\bf Qasper} & \rotatebox[origin=c]{30}{\bf MF-en} & \rotatebox[origin=c]{30}{\bf HotpotQA} & \rotatebox[origin=c]{30}{\bf 2WikiMQA} & \rotatebox[origin=c]{30}{\bf Musique} & \rotatebox[origin=c]{30}{\bf GovReport} & \rotatebox[origin=c]{30}{\bf QMSum} & \rotatebox[origin=c]{30}{\bf MultiNews} & \rotatebox[origin=c]{30}{\bf TREC} & \rotatebox[origin=c]{30}{\bf TriviaQA} & \rotatebox[origin=c]{30}{\bf SAMSum} & \rotatebox[origin=c]{30}{\bf PCount} & \rotatebox[origin=c]{30}{~\bf PRe~~} & \rotatebox[origin=c]{30}{~\bf Lcc~~} & \rotatebox[origin=c]{30}{~\bf RB-P~} \\
\cmidrule{1-19}

\multicolumn{19}{c}{\textbf{LLaMA-3.1-70B-Instruct}}\\
 - & Vanilla & 36.42 & 49.85 & 55.65 & 64.4 & 68.55 & 46.9 & 35.28 & 24.23 & 26.74 & 77.0 & 94.45 & 46.83 & 20.0 & 98.5 & 35.74 & 46.95 & 51.72 \\
\cmidrule{1-19}
 \multirow{10}*{\raisebox{10.0em}{\rotatebox{90}{KV-size 128}}}
 &   StreamingLLM  & 19.55 & 21.96 & 25.95 & 39.65 & 39.61 & 18.13 & 18.21 & 18.98 & 18.94 & 5.5 & 91.61 & 37.87 & 10.5 & 97.5 & 59.35 & 52.39 & 35.98 \\
 &   ExpectedAttention  & 20.92 & 25.17 & 29.53 & 32.91 & 35.18 & 7.08 & 23.17 & 15.5 & 24.47 & 20.0 & 92.53 & 38.29 & 13.5 & 67.5 & 38.39 & 40.08 & 32.76 \\
 &  TOVA & 32.92 & 34.58 & 48.6 & 57.88 & 56.65 & 38.8 & 20.65 & 21.19 & 18.87 & 33.0 & 94.02 & 43.09 & 7.5 & 93.5 & 62.76 & 60.25 & 45.27 \\
  \cdashline{2-19}
 &   SnapKV &  21.76 & 23.32 & 36.57 & 50.53 & 44.15 & 22.23 & 18.2 & 19.09 & 18.92 & 20.5 & 92.61 & 38.95 & 9.5 & 98.5 & 56.55 & 55.16 & 39.16  \\
 &   +\think~(0.5)    & 20.4 & 20.6 & 35.23 & 47.79 & 41.39 & 21.6 & 17.61 & 18.34 & 18.64 & 5.0 & 91.4 & 35.5 & 11.0 & 96.75 & 53.23 & 50.44 & 36.56 \\
 &   +\think~(0.8)    & 13.09 & 9.1 & 26.23 & 36.04 & 22.66 & 10.4 & 15.09 & 13.06 & 14.29 & 0.0 & 56.46 & 10.67 & 0.0 & 19.17 & 35.57 & 33.3 & 19.7 \\
 \hc &   +\textbf{\our}~(0.5)    & 20.01 & 22.88 & 37.19 & 50.54 & 43.07 & 19.73 & 18.1 & 18.96 & 18.99 & 21.0 & 93.06 & 39.05 & 9.0 & 98.5 & 53.57 & 54.24 & 38.62 \\
 \hd &   +\textbf{\our}~(0.8)    & 19.04 & 19.74 & 33.94 & 49.59 & 39.51 & 19.06 & 17.11 & 16.8 & 17.62 & 3.5 & 93.29 & 32.85 & 8.5 & 96.0 & 50.71 & 48.28 & 35.35 \\
 \cdashline{2-19}
 &   PyramidKV        &  36.53 & 49.06 & 55.67 & 65.39 & 67.96 & 46.6 & 35.25 & 24.25 & 26.95 & 77.5 & 94.35 & 47.11 & 21.0 & 98.5 & 36.2 & 46.77 & 51.82 \\
 \hc &   +\textbf{\our}~(0.5)    & 35.99 & 48.51 & 55.18 & 65.02 & 67.8 & 46.84 & 35.01 & 24.14 & 26.87 & 77.0 & 93.85 & 46.5 & 19.0 & 98.5 & 35.37 & 45.39 & 51.31  \\
 \cmidrule{1-19}
 \multirow{8}*{\raisebox{10.0em}{\rotatebox{90}{KV-size 512}}}
 &   StreamingLLM  & 20.08 & 28.43 & 27.65 & 45.26 & 43.04 & 22.5 & 24.35 & 19.64 & 24.2 & 47.5 & 92.26 & 40.67 & 11.0 & 97.5 & 63.09 & 55.94 & 41.44 \\
 &  TOVA &  32.54 & 43.81 & 50.85 & 54.86 & 62.61 & 24.81 & 17.48 & 20.13 & 24.07 & 61.5 & 94.1 & 47.36 & 3.5 & 16.5 & 52.99 & 62.44 & 41.85 \\
  \cdashline{2-19}
 &   SnapKV &  33.7 & 44.46 & 49.81 & 63.7 & 64.26 & 41.93 & 24.81 & 21.87 & 24.29 & 58.0 & 93.95 & 45.21 & 16.0 & 98.5 & 46.84 & 58.94 & 49.14  \\
 &   +\think~(0.5)    & 33.54 & 40.38 & 51.09 & 60.45 & 60.9 & 38.23 & 24.17 & 21.98 & 23.71 & 40.0 & 92.14 & 41.78 & 15.0 & 98.5 & 61.99 & 59.7 & 47.72 \\
 &   +\think~(0.8)    &  17.39 & 5.6 & 25.45 & 41.4 & 22.2 & 14.28 & 18.63 & 13.48 & 18.66 & 0.0 & 25.97 & 7.57 & 10.0 & 87.5 & 33.89 & 32.15 & 23.39\\
 \hc &   +\textbf{\our}~(0.5)    & 33.79 & 43.77 & 49.88 & 63.6 & 64.53 & 41.56 & 24.74 & 22.29 & 24.21 & 57.0 & 93.95 & 44.3 & 16.5 & 98.5 & 45.62 & 58.13 & 48.9 \\
 \hd &   +\textbf{\our}~(0.8)    & 32.08 & 41.45 & 49.5 & 59.84 & 56.91 & 36.55 & 22.99 & 21.97 & 23.02 & 38.0 & 93.2 & 39.29 & 15.0 & 98.0 & 56.77 & 56.54 & 46.32 \\
 \cdashline{2-19}
 &   PyramidKV        &  36.53 & 49.06 & 55.67 & 65.39 & 67.96 & 46.6 & 35.25 & 24.25 & 26.77 & 77.5 & 94.35 & 47.11 & 21.0 & 98.5 & 36.2 & 46.77 & 51.81 \\
  \cmidrule{1-19}
 \multirow{7}*{\raisebox{10.0em}{\rotatebox{90}{KV-size 1024}}} 
 &   StreamingLLM  & 23.1 & 32.35 & 29.8 & 51.12 & 47.22 & 22.36 & 26.4 & 20.12 & 25.9 & 60.5 & 93.72 & 41.95 & 14.0 & 96.5 & 60.28 & 59.95 & 44.08 \\
 &  TOVA & 27.31 & 47.6 & 55.04 & 61.25 & 66.88 & 37.54 & 26.12 & 21.12 & 26.04 & 70.0 & 94.1 & 47.3 & 15.0 & 98.5 & 42.65 & 58.25 & 49.67 \\
  \cdashline{2-19}
 &   SnapKV &  35.17 & 48.76 & 52.53 & 65.03 & 66.14 & 44.55 & 28.01 & 22.35 & 26.17 & 65.5 & 93.95 & 45.03 & 15.5 & 98.5 & 40.57 & 55.82 & 50.22  \\
 &   +\think~(0.5)    & 36.05 & 46.36 & 52.3 & 62.73 & 62.43 & 40.67 & 26.98 & 23.17 & 25.17 & 53.0 & 93.14 & 42.39 & 13.5 & 99.0 & 63.86 & 59.65 & 50.02 \\
 \hc &   +\textbf{\our}~(0.5)    & 35.26 & 47.63 & 52.34 & 64.76 & 66.51 & 44.68 & 27.88 & 23.19 & 26.18 & 66.0 & 93.95 & 45.54 & 14.5 & 98.5 & 40.12 & 55.21 & 50.14 \\
 \hd &   +\textbf{\our}~(0.8)    & 35.11 & 45.39 & 51.37 & 59.55 & 60.45 & 40.1 & 24.67 & 22.79 & 24.87 & 52.5 & 92.95 & 36.9 & 12.5 & 98.5 & 54.81 & 55.4 & 47.99 \\
 &   PyramidKV        &  36.53 & 49.06 & 55.56 & 65.39 & 67.96 & 46.6 & 35.25 & 24.25 & 26.64 & 77.5 & 94.35 & 47.12 & 21.0 & 98.5 & 36.54 & 46.77 & 51.81 \\
   \cmidrule{1-19}
 \multirow{7}*{\raisebox{10.0em}{\rotatebox{90}{KV-size 2048}}} 
 &   StreamingLLM  & 25.53 & 41.15 & 38.29 & 52.91 & 53.74 & 26.76 & 29.14 & 20.82 & 26.34 & 66.5 & 93.3 & 43.55 & 17.5 & 97.0 & 47.63 & 60.9 & 46.32 \\
 &  TOVA & 34.03 & 48.95 & 55.37 & 63.53 & 67.43 & 46.93 & 31.0 & 23.17 & 26.68 & 76.0 & 94.35 & 47.0 & 16.5 & 98.5 & 37.87 & 53.94 & 51.33 \\
  \cdashline{2-19}
 &   SnapKV &  36.49 & 50.02 & 53.43 & 65.58 & 65.28 & 46.99 & 30.75 & 23.49 & 26.35 & 70.5 & 94.45 & 46.1 & 17.5 & 98.5 & 37.22 & 53.41 & 51.0  \\
 &   +\think~(0.5)    & 36.49 & 48.46 & 52.92 & 65.15 & 66.19 & 46.28 & 30.82 & 23.83 & 26.73 & 70.5 & 93.95 & 46.41 & 18.0 & 98.5 & 37.09 & 52.31 & 50.85 \\
 \hc &   +\textbf{\our}~(0.5)    & 35.66 & 50.33 & 51.34 & 62.69 & 62.64 & 41.71 & 28.99 & 23.74 & 25.93 & 60.5 & 93.14 & 41.99 & 16.5 & 99.0 & 63.66 & 59.72 & 51.1 \\
 \hd &   +\textbf{\our}~(0.8)    & 35.24 & 48.11 & 51.69 & 59.3 & 59.89 & 39.86 & 26.44 & 23.09 & 25.46 & 56.0 & 93.2 & 35.72 & 16.0 & 98.0 & 50.07 & 53.62 & 48.23 \\
 &   PyramidKV        &  37.11 & 48.66 & 55.56 & 64.38 & 67.06 & 46.67 & 30.09 & 23.38 & 26.51 & 67.0 & 92.87 & 46.18 & 18.0 & 98.5 & 36.52 & 53.64 & 50.76 \\
 \bottomrule
    \end{tabular}}
     \caption{Performance comparison on LLaMA-3.1-70B-Instruct at LongBench.}
     \label{tab:full_llama3.1-70b}
    \end{table*}

\begin{table*}[t!]
    \centering \scriptsize

\resizebox{\textwidth}{!}{\begin{tabular}
{C{1em}l@{\hspace{0.05ex}}C{3.8em}@{\hspace{0.05ex}}C{3.8em}@{\hspace{0.05ex}}C{3.8em}@{\hspace{0.05ex}}c@{\hspace{0.05ex}}c@{\hspace{0.05ex}}C{3.8em}@{\hspace{0.05ex}}c@{\hspace{0.05ex}}C{3.8em}@{\hspace{0.05ex}}c@{\hspace{0.05ex}}C{3.8em}@{\hspace{0.05ex}}c@{\hspace{0.05ex}}c@{\hspace{0.05ex}}c@{\hspace{0.6ex}}C{3.8em}@{\hspace{0.6ex}}C{3.8em}@{\hspace{0.6ex}}C{3.8em}@{\hspace{0.6ex}}c}
    \toprule
& \multirow{4}{*}{\textbf{Method}}& \multicolumn{3}{c}{\textbf{Single-Document QA}} & \multicolumn{3}{c}{\textbf{Multi-Document QA}}& \multicolumn{3}{c}{\textbf{Summarization}}& \multicolumn{3}{c}{\textbf{Few-shot Learning}}& \multicolumn{2}{c}{\textbf{Synthetic}} & \multicolumn{2}{c}{\textbf{Code}}&\multirow{4}{*}{\textbf{Avg.}} \\
& & \rotatebox[origin=c]{30}{\bf NrtvQA} & \rotatebox[origin=c]{30}{\bf Qasper} & \rotatebox[origin=c]{30}{\bf MF-en} & \rotatebox[origin=c]{30}{\bf HotpotQA} & \rotatebox[origin=c]{30}{\bf 2WikiMQA} & \rotatebox[origin=c]{30}{\bf Musique} & \rotatebox[origin=c]{30}{\bf GovReport} & \rotatebox[origin=c]{30}{\bf QMSum} & \rotatebox[origin=c]{30}{\bf MultiNews} & \rotatebox[origin=c]{30}{\bf TREC} & \rotatebox[origin=c]{30}{\bf TriviaQA} & \rotatebox[origin=c]{30}{\bf SAMSum} & \rotatebox[origin=c]{30}{\bf PCount} & \rotatebox[origin=c]{30}{~\bf PRe~~} & \rotatebox[origin=c]{30}{~\bf Lcc~~} & \rotatebox[origin=c]{30}{~\bf RB-P~} \\
\cmidrule{1-19}

\multicolumn{19}{c}{\textbf{Qwen-3-8B}}\\
 - & Vanilla & 29.07 & 44.26 & 55.57 & 62.3 & 48.11 & 35.81 & 33.59 & 24.52 & 24.9 & 69.0 & 88.9 & 41.04 & 9.0 & 91.43 & 68.99 & 67.93 & 49.65 \\
\cmidrule{1-19}
 \multirow{12}*{\raisebox{10.0em}{\rotatebox{90}{KV-size 128}}}
 &   StreamingLLM  & 14.81 & 19.68 & 24.18 & 28.9 & 29.81 & 9.55 & 15.78 & 18.72 & 15.58 & 17.75 & 43.86 & 34.02 & 3.0 & 40.46 & 63.21 & 60.35 & 27.48 \\
 &   ExpectedAttention  &  16.9 & 27.23 & 29.24 & 25.32 & 14.29 & 9.95 & 23.55 & 20.31 & 21.48 & 8.5 & 77.06 & 35.23 & 4.27 & 12.33 & 46.07 & 41.4 & 25.82\\
 &  TOVA & 18.09 & 25.8 & 39.71 & 44.34 & 34.02 & 17.83 & 17.11 & 19.33 & 14.55 & 17.0 & 88.03 & 38.51 & 7.5 & 84.92 & 62.42 & 59.21 & 36.77 \\
  \cdashline{2-19}
 &   SnapKV &  17.12 & 23.54 & 33.8 & 40.24 & 34.32 & 15.47 & 16.15 & 19.12 & 15.69 & 21.0 & 72.55 & 36.93 & 3.5 & 74.3 & 63.7 & 59.03 & 34.15  \\
 &   +\think~(0.5)    & 16.15 & 23.52 & 30.33 & 38.71 & 31.45 & 16.99 & 15.25 & 19.19 & 14.43 & 9.5 & 68.64 & 29.54 & 1.5 & 72.62 & 56.51 & 53.93 & 31.14 \\
 &   +\think~(0.8)    & 11.51 & 14.43 & 18.33 & 26.1 & 20.52 & 7.59 & 12.43 & 17.43 & 11.34 & 0.5 & 27.87 & 9.13 & 3.0 & 43.5 & 29.13 & 28.46 & 17.58 \\
 \hc &   +\textbf{\our}~(0.5)    & 16.82 & 23.08 & 33.76 & 40.25 & 34.49 & 15.61 & 15.99 & 18.99 & 15.3 & 19.5 & 72.26 & 36.61 & 3.0 & 74.23 & 62.99 & 58.48 & 33.84 \\
 \hd &   +\textbf{\our}~(0.8)    & 17.16 & 23.03 & 32.66 & 38.41 & 33.53 & 14.7 & 15.69 & 18.92 & 15.55 & 18.5 & 64.76 & 35.5 & 1.0 & 70.79 & 62.24 & 58.54 & 32.56 \\
 \cdashline{2-19}
 &   PyramidKV        &  29.76 & 44.26 & 56.27 & 61.52 & 48.17 & 33.64 & 33.54 & 24.42 & 24.61 & 68.0 & 88.52 & 41.65 & 10.0 & 91.92 & 67.0 & 66.66 & 49.37 \\
 &   +\think~(0.5)    & 24.74 & 42.12 & 51.07 & 60.22 & 44.99 & 31.28 & 32.47 & 23.91 & 24.03 & 69.0 & 86.55 & 28.49 & 8.0 & 99.75 & 61.04 & 60.23 & 46.74 \\
 \hc &   +\textbf{\our}~(0.5)    &  29.12 & 43.5 & 55.57 & 62.14 & 47.95 & 33.49 & 33.39 & 24.34 & 24.09 & 68.0 & 89.07 & 41.08 & 10.0 & 92.67 & 66.21 & 65.86 & 49.15 \\
 \hd &   +\textbf{\our}~(0.8)    & 28.36 & 44.03 & 53.46 & 59.63 & 49.67 & 31.46 & 33.08 & 23.82 & 24.22 & 68.0 & 89.6 & 39.84 & 10.0 & 95.6 & 66.03 & 64.27 & 48.82 \\
 \cmidrule{1-19}
  \multirow{12}*{\raisebox{10.0em}{\rotatebox{90}{KV-size 512}}}
 &   StreamingLLM  & 16.91 & 23.14 & 26.92 & 32.25 & 32.57 & 10.41 & 22.37 & 19.71 & 21.29 & 45.0 & 62.68 & 36.52 & 7.0 & 34.58 & 67.59 & 63.55 & 32.66 \\
 &   ExpectedAttention  & 21.17 & 31.86 & 36.85 & 44.82 & 37.1 & 20.29 & 28.9 & 21.23 & 24.11 & 45.0 & 85.71 & 38.66 & 3.26 & 21.33 & 56.83 & 53.56 & 35.67 \\
 &  TOVA & 22.3 & 37.04 & 48.71 & 54.34 & 45.15 & 23.84 & 22.8 & 20.8 & 20.63 & 51.0 & 88.88 & 42.25 & 4.5 & 98.06 & 68.85 & 66.06 & 44.7 \\
  \cdashline{2-19}
 &   SnapKV &  25.11 & 34.04 & 47.47 & 55.54 & 40.39 & 26.09 & 22.83 & 21.32 & 21.2 & 48.5 & 88.2 & 38.69 & 7.58 & 97.31 & 68.56 & 67.64 & 44.4  \\
 &   +\think~(0.5)    & 22.26 & 32.85 & 45.24 & 54.57 & 38.56 & 27.18 & 20.81 & 20.85 & 19.09 & 34.0 & 86.6 & 31.57 & 4.5 & 99.5 & 61.68 & 60.95 & 41.26 \\
 &   +\think~(0.8)    & 10.25 & 17.31 & 24.36 & 29.42 & 20.44 & 11.71 & 16.61 & 17.66 & 14.04 & 0.0 & 50.87 & 10.05 & 4.5 & 77.25 & 30.43 & 33.04 & 23.0 \\
 \hc &   +\textbf{\our}~(0.5)    & 24.72 & 33.56 & 46.53 & 54.93 & 41.6 & 26.29 & 22.75 & 20.92 & 21.16 & 49.0 & 89.3 & 38.22 & 8.03 & 96.78 & 67.98 & 67.25 & 44.31 \\
 \hd &   +\textbf{\our}~(0.8)    & 24.6 & 32.77 & 45.96 & 56.33 & 40.4 & 23.76 & 22.47 & 20.75 & 20.93 & 42.5 & 87.3 & 37.15 & 7.02 & 97.83 & 67.51 & 66.52 & 43.36 \\
  \cdashline{2-19}
 &   PyramidKV        &  29.76 & 44.26 & 56.27 & 61.52 & 48.17 & 33.64 & 33.54 & 24.42 & 24.28 & 68.0 & 88.52 & 41.65 & 10.0 & 91.92 & 67.0 & 66.66 & 49.35 \\
 &   +\think~(0.5)    & 24.74 & 42.12 & 51.07 & 60.22 & 44.99 & 31.28 & 32.47 & 23.91 & 23.82 & 69.5 & 86.39 & 28.22 & 7.5 & 99.75 & 61.13 & 60.12 & 46.7 \\
 \hc &   +\textbf{\our}~(0.5)    &  29.03 & 43.77 & 55.1 & 62.07 & 48.21 & 33.94 & 33.47 & 24.4 & 24.17 & 69.0 & 88.57 & 41.1 & 9.5 & 91.92 & 66.75 & 66.21 & 49.2 \\
 \hd &   +\textbf{\our}~(0.8)    & 28.7 & 43.97 & 53.29 & 60.98 & 48.45 & 31.44 & 32.63 & 23.66 & 24.25 & 67.0 & 89.43 & 40.03 & 9.5 & 95.1 & 66.01 & 64.77 & 48.7 \\
  \cmidrule{1-19}
  \multirow{11}*{\raisebox{10.0em}{\rotatebox{90}{KV-size 1024}}} 
 &   StreamingLLM  & 19.31 & 25.15 & 29.14 & 33.38 & 33.8 & 11.9 & 25.51 & 20.71 & 23.55 & 53.5 & 71.95 & 37.15 & 8.5 & 31.2 & 68.17 & 65.42 & 34.9 \\
 &   ExpectedAttention  &  24.01 & 35.71 & 42.89 & 50.51 & 42.38 & 24.08 & 30.44 & 21.7 & 24.71 & 63.0 & 86.36 & 39.64 & 4.36 & 29.07 & 63.3 & 59.55 & 40.11\\
 &  TOVA & 25.0 & 39.69 & 49.96 & 58.53 & 45.63 & 29.51 & 26.39 & 21.41 & 23.19 & 62.5 & 88.25 & 42.52 & 7.66 & 95.72 & 69.18 & 67.44 & 47.04 \\
  \cdashline{2-19}
 &   SnapKV &  25.59 & 39.64 & 52.09 & 56.63 & 44.85 & 32.69 & 26.23 & 22.04 & 23.17 & 61.5 & 89.18 & 39.64 & 8.6 & 96.88 & 69.69 & 69.15 & 47.35  \\
 &   +\think~(0.5)    & 23.51 & 36.86 & 49.17 & 57.85 & 42.75 & 30.57 & 23.84 & 22.04 & 22.05 & 51.5 & 86.59 & 31.04 & 5.0 & 100.0 & 61.96 & 61.05 & 44.11 \\
 &   +\think~(0.8)    & 10.81 & 15.51 & 24.45 & 29.39 & 19.26 & 13.58 & 17.51 & 18.24 & 14.6 & 0.0 & 42.82 & 9.45 & 3.5 & 66.42 & 28.65 & 32.29 & 21.65 \\
 \hc &   +\textbf{\our}~(0.5)    & 25.71 & 39.83 & 52.44 & 55.94 & 45.77 & 32.51 & 26.25 & 22.2 & 23.2 & 60.0 & 89.2 & 39.11 & 7.59 & 97.54 & 68.66 & 68.98 & 47.18 \\
 \cdashline{2-19}
 &   PyramidKV        &  29.76 & 44.26 & 56.05 & 61.52 & 48.17 & 33.64 & 33.54 & 24.42 & 23.88 & 68.0 & 88.52 & 41.61 & 10.0 & 91.92 & 66.83 & 66.66 & 49.3 \\
 &   +\think~(0.5)    & 25.17 & 42.4 & 49.81 & 61.18 & 46.02 & 31.23 & 32.72 & 23.82 & 23.48 & 69.5 & 86.39 & 28.22 & 7.5 & 99.75 & 61.69 & 60.12 & 46.81 \\
 \hc &   +\textbf{\our}~(0.5)    &  29.32 & 43.79 & 54.92 & 62.15 & 48.89 & 33.4 & 33.25 & 24.34 & 23.79 & 68.5 & 88.54 & 41.28 & 10.0 & 92.29 & 66.64 & 66.16 & 49.2 \\
 \hd &   +\textbf{\our}~(0.8)    & 28.16 & 43.57 & 53.83 & 60.06 & 46.95 & 31.48 & 33.07 & 24.07 & 23.62 & 68.0 & 89.1 & 39.93 & 8.5 & 94.6 & 65.76 & 64.84 & 48.47 \\
 \cmidrule{1-19}
  \multirow{11}*{\raisebox{10.0em}{\rotatebox{90}{KV-size 2048}}} 
 &   StreamingLLM  & 22.55 & 31.89 & 35.35 & 39.99 & 39.72 & 16.92 & 28.83 & 21.35 & 24.57 & 62.0 & 80.83 & 38.22 & 7.5 & 36.92 & 67.1 & 65.52 & 38.7 \\
 &   ExpectedAttention  &  26.73 & 40.61 & 48.65 & 54.65 & 43.82 & 28.51 & 32.08 & 22.71 & 24.86 & 67.17 & 87.4 & 40.79 & 6.53 & 49.12 & 66.46 & 63.74 & 43.99\\
 &  TOVA & 27.2 & 42.68 & 51.96 & 59.65 & 47.8 & 32.41 & 29.54 & 22.55 & 24.47 & 68.0 & 89.0 & 42.45 & 10.1 & 95.27 & 67.7 & 66.47 & 48.58 \\
  \cdashline{2-19}
 &   SnapKV &  28.89 & 42.56 & 53.93 & 60.47 & 47.73 & 32.65 & 29.68 & 22.85 & 24.57 & 67.0 & 89.77 & 40.67 & 6.85 & 96.83 & 68.05 & 67.19 & 48.73  \\
 &   +\think~(0.5)    & 25.77 & 40.45 & 50.64 & 60.34 & 43.55 & 32.7 & 27.49 & 22.63 & 23.81 & 62.5 & 86.51 & 30.53 & 6.0 & 100.0 & 62.3 & 62.6 & 46.11 \\
 &   +\think~(0.8)    & 9.39 & 11.95 & 21.63 & 26.22 & 18.33 & 11.71 & 17.62 & 17.58 & 14.73 & 0.0 & 33.65 & 9.6 & 3.0 & 52.5 & 26.89 & 29.85 & 19.04 \\
 \hc &   +\textbf{\our}~(0.5)    & 28.11 & 43.24 & 53.07 & 61.07 & 49.42 & 34.31 & 29.37 & 22.73 & 24.28 & 67.5 & 89.43 & 39.8 & 5.82 & 97.21 & 68.12 & 68.47 & 48.87 \\
 \cdashline{2-19}
 &   PyramidKV        &  25.88 & 39.26 & 52.05 & 57.87 & 42.73 & 29.2 & 27.37 & 22.58 & 24.09 & 60.0 & 89.66 & 40.17 & 7.1 & 97.18 & 67.5 & 65.81 & 46.78 \\
 &   +\think~(0.5)    & 23.57 & 38.32 & 51.62 & 56.82 & 38.67 & 27.5 & 25.58 & 22.47 & 23.42 & 55.0 & 86.82 & 30.94 & 6.0 & 99.75 & 61.38 & 60.34 & 44.26 \\
 \hc &   +\textbf{\our}~(0.5)    &  25.88 & 40.06 & 53.1 & 57.91 & 42.7 & 28.85 & 27.37 & 22.59 & 24.2 & 62.5 & 88.83 & 39.68 & 8.0 & 97.85 & 66.95 & 65.9 & 47.02 \\
 \hd &   +\textbf{\our}~(0.8)    & 23.77 & 39.99 & 50.94 & 57.31 & 42.23 & 24.97 & 26.9 & 22.57 & 23.72 & 55.0 & 89.6 & 38.05 & 7.0 & 98.42 & 65.87 & 64.71 & 45.69 \\
 \bottomrule
    \end{tabular}}
     \caption{Performance comparison on Qwen3-8B at LongBench.}
     \label{tab:full_qwen3-8b}
    \end{table*}

\begin{table*}[t!]
    \centering \scriptsize

\resizebox{\textwidth}{!}{\begin{tabular}
{C{1em}l@{\hspace{0.05ex}}C{3.8em}@{\hspace{0.05ex}}C{3.8em}@{\hspace{0.05ex}}C{3.8em}@{\hspace{0.05ex}}c@{\hspace{0.05ex}}c@{\hspace{0.05ex}}C{3.8em}@{\hspace{0.05ex}}c@{\hspace{0.05ex}}C{3.8em}@{\hspace{0.05ex}}c@{\hspace{0.05ex}}C{3.8em}@{\hspace{0.05ex}}c@{\hspace{0.05ex}}c@{\hspace{0.05ex}}c@{\hspace{0.6ex}}C{3.8em}@{\hspace{0.6ex}}C{3.8em}@{\hspace{0.6ex}}C{3.8em}@{\hspace{0.6ex}}c}
    \toprule
& \multirow{4}{*}{\textbf{Method}}& \multicolumn{3}{c}{\textbf{Single-Document QA}} & \multicolumn{3}{c}{\textbf{Multi-Document QA}}& \multicolumn{3}{c}{\textbf{Summarization}}& \multicolumn{3}{c}{\textbf{Few-shot Learning}}& \multicolumn{2}{c}{\textbf{Synthetic}} & \multicolumn{2}{c}{\textbf{Code}}&\multirow{4}{*}{\textbf{Avg.}} \\
& & \rotatebox[origin=c]{30}{\bf NrtvQA} & \rotatebox[origin=c]{30}{\bf Qasper} & \rotatebox[origin=c]{30}{\bf MF-en} & \rotatebox[origin=c]{30}{\bf HotpotQA} & \rotatebox[origin=c]{30}{\bf 2WikiMQA} & \rotatebox[origin=c]{30}{\bf Musique} & \rotatebox[origin=c]{30}{\bf GovReport} & \rotatebox[origin=c]{30}{\bf QMSum} & \rotatebox[origin=c]{30}{\bf MultiNews} & \rotatebox[origin=c]{30}{\bf TREC} & \rotatebox[origin=c]{30}{\bf TriviaQA} & \rotatebox[origin=c]{30}{\bf SAMSum} & \rotatebox[origin=c]{30}{\bf PCount} & \rotatebox[origin=c]{30}{~\bf PRe~~} & \rotatebox[origin=c]{30}{~\bf Lcc~~} & \rotatebox[origin=c]{30}{~\bf RB-P~} \\
\cmidrule{1-19}

\multicolumn{19}{c}{\textbf{Qwen-3-32B}}\\
 \multirow{8}*{\raisebox{10.0em}{\rotatebox{90}{KV-size 128}}}
 &   StreamingLLM  & 17.04 & 21.12 & 26.96 & 30.13 & 37.41 & 9.69 & 15.18 & 19.17 & 16.59 & 17.5 & 21.37 & 29.11 & 3.0 & 85.01 & 25.86 & 28.1 & 25.2 \\
 &  TOVA & 30.32 & 31.87 & 42.71 & 55.13 & 48.72 & 28.31 & 17.68 & 20.31 & 16.41 & 21.5 & 71.92 & 32.0 & 5.5 & 91.0 & 35.25 & 36.03 & 36.54 \\
  \cdashline{2-19}
 &   SnapKV &  19.95 & 24.41 & 35.94 & 39.32 & 39.29 & 11.23 & 16.39 & 19.46 & 16.53 & 18.25 & 35.29 & 31.86 & 4.5 & 91.62 & 40.0 & 44.29 & 30.52  \\
 &   +\think~(0.5)    & 20.79 & 23.39 & 36.97 & 33.74 & 35.55 & 11.15 & 15.4 & 19.71 & 15.74 & 4.0 & 31.86 & 33.12 & 4.5 & 89.03 & 41.18 & 43.81 & 28.75 \\
 &   +\think~(0.8)    & 13.79 & 18.77 & 20.3 & 25.04 & 22.29 & 8.88 & 11.24 & 17.8 & 12.1 & 0.0 & 21.62 & 13.15 & 2.5 & 70.9 & 26.54 & 28.88 & 19.61 \\
 \hc &   +\textbf{\our}~(0.5)    & 20.57 & 24.09 & 35.54 & 38.06 & 38.71 & 10.54 & 16.46 & 19.45 & 16.52 & 18.5 & 32.18 & 31.86 & 4.5 & 91.57 & 40.66 & 44.06 & 30.2 \\
 \hd &   +\textbf{\our}~(0.8)    & 21.78 & 22.85 & 34.93 & 38.07 & 37.8 & 12.36 & 16.4 & 19.22 & 16.28 & 17.0 & 29.39 & 31.48 & 3.5 & 86.38 & 42.51 & 43.12 & 29.57 \\
 &   PyramidKV        &  31.44 & 47.85 & 51.49 & 53.25 & 55.54 & 28.19 & 33.19 & 24.14 & 25.39 & 71.0 & 77.32 & 39.09 & 16.5 & 99.75 & 24.4 & 30.59 & 44.32 \\
 \cmidrule{1-19}
 \multirow{7}*{\raisebox{10.0em}{\rotatebox{90}{KV-size 512}}}
 &   StreamingLLM  & 17.91 & 25.73 & 29.19 & 31.66 & 39.0 & 11.66 & 21.19 & 19.14 & 22.31 & 46.5 & 27.46 & 32.05 & 6.5 & 77.11 & 27.56 & 29.59 & 29.04 \\
 &  TOVA & 31.83 & 40.26 & 48.49 & 54.42 & 54.98 & 31.49 & 23.49 & 21.75 & 22.35 & 62.5 & 78.01 & 37.9 & 11.0 & 98.18 & 28.21 & 32.68 & 42.35 \\
  \cdashline{2-19}
 &   SnapKV &  30.06 & 39.74 & 47.38 & 55.96 & 50.59 & 28.89 & 23.35 & 21.61 & 22.29 & 40.0 & 72.23 & 36.36 & 11.0 & 93.07 & 28.24 & 37.69 & 39.9  \\
 &   +\think~(0.5)    & 26.8 & 38.3 & 47.0 & 53.64 & 42.53 & 26.07 & 21.81 & 22.14 & 20.74 & 30.5 & 77.94 & 37.41 & 11.5 & 98.05 & 30.5 & 36.47 & 38.84 \\
 &   +\think~(0.8)    & 14.58 & 21.64 & 23.01 & 28.87 & 17.27 & 13.34 & 14.75 & 18.1 & 15.12 & 0.0 & 33.46 & 10.87 & 7.0 & 85.66 & 22.56 & 22.71 & 21.81 \\
 \hc &   +\textbf{\our}~(0.5)    & 30.6 & 38.81 & 46.44 & 55.66 & 49.99 & 28.92 & 23.56 & 21.65 & 22.54 & 39.0 & 70.17 & 34.76 & 11.0 & 93.31 & 27.35 & 37.7 & 39.47 \\
 &   PyramidKV        &  31.44 & 47.85 & 51.49 & 53.25 & 55.54 & 28.19 & 33.19 & 24.14 & 25.09 & 71.0 & 77.32 & 39.09 & 16.5 & 99.75 & 24.4 & 30.59 & 44.3 \\
  \cmidrule{1-19}
 \multirow{8}*{\raisebox{10.0em}{\rotatebox{90}{KV-size 1024}}} 
 &   StreamingLLM  & 20.66 & 27.99 & 30.63 & 35.26 & 38.14 & 12.78 & 23.67 & 20.13 & 24.05 & 56.5 & 37.11 & 34.17 & 8.0 & 63.12 & 28.85 & 32.81 & 30.87 \\
 &  TOVA & 31.03 & 44.82 & 50.94 & 55.29 & 57.49 & 30.37 & 26.25 & 22.43 & 24.25 & 64.5 & 77.86 & 38.98 & 15.5 & 98.54 & 24.59 & 29.21 & 43.25 \\
  \cdashline{2-19}
 &   SnapKV &  31.17 & 44.08 & 48.35 & 56.13 & 53.94 & 30.03 & 26.76 & 22.58 & 24.54 & 50.25 & 79.74 & 36.79 & 14.0 & 95.29 & 23.97 & 33.39 & 41.94  \\
 &   +\think~(0.5)    & 28.24 & 41.85 & 47.89 & 54.1 & 45.95 & 27.42 & 24.57 & 22.92 & 23.6 & 45.5 & 82.39 & 38.09 & 12.0 & 99.75 & 26.13 & 36.3 & 41.04 \\
 &   +\think~(0.8)    & 15.01 & 22.06 & 20.19 & 26.13 & 15.92 & 11.32 & 16.26 & 17.51 & 15.45 & 0.0 & 29.65 & 7.53 & 7.0 & 87.57 & 20.93 & 20.67 & 20.83 \\
 \hc &   +\textbf{\our}~(0.5)    & 30.62 & 44.09 & 47.06 & 56.38 & 52.26 & 30.51 & 27.0 & 22.33 & 24.64 & 51.0 & 79.55 & 35.87 & 15.5 & 96.32 & 24.26 & 32.95 & 41.9 \\
 \hd &   +\textbf{\our}~(0.8)    & 30.18 & 43.72 & 47.09 & 56.42 & 50.13 & 30.06 & 26.26 & 22.74 & 24.17 & 46.17 & 76.16 & -1 & -1 & -1 & -1 & -1 & 41.19 \\
 &   PyramidKV        &  31.44 & 47.85 & 51.3 & 53.25 & 55.7 & 28.19 & 33.19 & 24.14 & 24.82 & 71.0 & 77.65 & 39.06 & 16.5 & 99.75 & 23.71 & 30.59 & 44.26 \\
   \cmidrule{1-19}
 \multirow{7}*{\raisebox{10.0em}{\rotatebox{90}{KV-size 2048}}} 
 &   StreamingLLM  & 23.1 & 35.85 & 35.58 & 37.33 & 44.36 & 16.88 & 27.19 & 21.03 & 24.82 & 60.5 & 42.44 & 36.5 & 13.5 & 62.0 & 28.88 & 31.62 & 33.85 \\
 &  TOVA & 32.22 & 46.91 & 51.48 & 55.72 & 55.69 & 30.93 & 29.35 & 23.62 & 25.08 & 67.5 & 77.7 & 38.79 & 16.0 & 99.67 & 24.09 & 30.04 & 44.05 \\
  \cdashline{2-19}
 &   SnapKV &  32.36 & 45.84 & 51.04 & 54.21 & 55.28 & 31.22 & 29.48 & 23.63 & 24.96 & 62.5 & 79.03 & 38.09 & 14.0 & 98.1 & 23.75 & 33.19 & 43.54  \\
 &   +\think~(0.5)    & 28.95 & 43.23 & 48.36 & 54.22 & 45.77 & 29.25 & 28.16 & 23.56 & 24.72 & 60.17 & 84.44 & 37.95 & 11.0 & 100.0 & 24.42 & 34.04 & 42.39 \\
 &   +\think~(0.8)    & 14.06 & 21.21 & 17.2 & 23.84 & 15.96 & 9.55 & 16.75 & 17.33 & 15.29 & 0.5 & 30.04 & 4.53 & 9.0 & 89.79 & 18.22 & 19.38 & 20.17 \\
 \hc &   +\textbf{\our}~(0.5)    & 32.29 & 45.76 & 50.09 & 55.42 & 54.04 & 31.44 & 29.58 & 23.66 & 25.18 & 61.5 & 77.75 & 36.97 & 13.0 & 98.32 & 25.0 & 33.2 & 43.33 \\
 &   PyramidKV        &  29.62 & 43.39 & 50.28 & 55.9 & 51.83 & 28.31 & 26.45 & 22.21 & 24.39 & 47.67 & 76.77 & 36.73 & 16.0 & 98.35 & 24.38 & 38.59 & 41.93 \\
 \bottomrule
    \end{tabular}}
     \caption{Performance comparison on Qwen3-32B at LongBench.}
     \label{tab:full_qwen3-32b}
    \end{table*}

\subsection{RULER}
\label{app:ex_ruler}

To further assess \our{}'s robustness under extreme long-context settings, we evaluate its performance on the RULER benchmark with 8K and 16K input lengths under various cache budgets (20\% and 50\%). The results are reported in Table~\ref{tab:app_full_ruler}.

Together, these results reinforce the compatibility of our method with diverse LLM architectures and its potential as a plug-and-play component for long-context optimization.

\begin{table*}[t]
\centering
\scriptsize
\setlength{\tabcolsep}{4.5pt}{
\begin{tabular}{clc*{13}{>{\centering\arraybackslash}p{0.6cm}}}
\toprule
& \textbf{Method} & \textbf{Niah1} & \textbf{Niah2} & \textbf{Niah3} & \textbf{MKey1} & \textbf{MKey2} & \textbf{MKey3} & \textbf{MValue} & \textbf{MQuery} & \textbf{VT} & \textbf{CWE} & \textbf{FWE} & \textbf{QA1} & \textbf{QA2} & \textbf{Avg.} \\
\midrule
\multicolumn{16}{c}{\textbf{16K}}\\
& Vanilla    & 100.0 & 100.0 & 100.0 & 99.6 & 100 & 99.2 & 99.1 & 99.0 & 99.8 & 88.9 & 90.0 & 81.0 & 57.2 & 93.36 \\
\midrule
\multirow{13}*{\raisebox{10.0em}{\rotatebox{90}{20\% KV cache}}} & StreamingLLM  & 18.8 & 17.4 & 19.0 & 20.2 & 20.0 & 18.4 & 18.25 & 18.2 & 32.84 & 0.18 & 81.33 & 31.4 & 33.6 & 25.35\\
& ExpectedAttention  &  99.2 & 42.0 & 3.4 & 33.8 & 57.0 & 0.8 & 9.35 & 21.1 & 66.12 & 54.46 & 70.6 & 72.0 & 48.2 & 44.46\\
& TOVA      &  100.0 & 100.0 & 97.8 & 99.4 & 96.8 & 0.4 & 98.9 & 99.25 & 99.76 & 54.04 & 90.8 & 77.4 & 54.6 & 82.24\\
\cdashline{2-16}
& SnapKV        & 100.0 & 100.0 & 10.0 & 99.8 & 97.2 & 63.2 & 97.7 & 99.45 & 97.36 & 53.92 & 85.73 & 80.8 & 57.2 & 80.18 \\
& +\think(0.5)  & 96.6 & 99.6 & 9.4 & 99.0 & 92.2 & 55.4 & 98.55 & 98.25 & 94.84 & 29.12 & 88.87 & 76.0 & 50.6 & 76.03 \\
& +\think(0.8)  & 0.0 & 0.0 & 0.0 & 0.0 & 0.0 & 0.0 & 0.05 & 0.0 & 0.0 & 0.32 & 0.0 & 18.8 & 20.2 & 3.03\\
\hc &   +\textbf{\our}~(0.5)    & 100.0 & 100.0 & 10.2 & 99.4 & 96.6 & 62.8 & 98.05 & 99.45 & 97.64 & 53.8 & 86.2 & 80.8 & 56.0 & 80.07 \\
\hd &   +\textbf{\our}~(0.8)    & 100.0 & 99.8 & 9.6 & 99.2 & 94.2 & 49.4 & 98.1 & 98.75 & 96.64 & 41.12 & 87.07 & 80.0 & 53.8 & 77.51\\
\cdashline{2-16}
& PyramidKV  & 100.0 & 100.0 & 5.0 & 99.8 & 98.2 & 55.0 & 98.6 & 99.35 & 98.6 & 16.88 & 87.0 & 80.0 & 57.2 & 76.59\\
& +\think(0.5)  & 97.2 & 100.0 & 4.8 & 99.4 & 93.0 & 49.2 & 98.7 & 98.75 & 96.16 & 8.46 & 88.33 & 76.2 & 52.4 & 74.05 \\
& +\think(0.8)  & 0.0 & 0.0 & 0.0 & 0.0 & 0.0 & 0.0 & 0.0 & 0.0 & 0.0 & 0.24 & 0.0 & 14.8 & 19.4 & 2.65\\
\hc &   +\textbf{\our}~(0.5)    & 99.2 & 99.2 & 5.2 & 99.4 & 97.6 & 54.4 & 97.95 & 98.7 & 98.16 & 16.84 & 86.27 & 79.6 & 56.8 & 76.1\\
\hd &   +\textbf{\our}~(0.8)    & 99.4 & 98.8 & 5.2 & 99.2 & 94.4 & 44.2 & 97.1 & 97.7 & 95.24 & 12.08 & 86.2 & 78.4 & 54.0 & 73.99\\

\midrule
\multirow{13}*{\raisebox{10.0em}{\rotatebox{90}{50\% KV cache}}} & StreamingLLM  & 47.0 & 45.4 & 49.4 & 51.2 & 48.6 & 48.0 & 48.1 & 48.0 & 68.56 & 10.84 & 85.13 & 82.6 & 43.4 & 52.02\\
& ExpectedAttention  & 100.0 & 93.0 & 18.0 & 93.4 & 98.0 & 42.6 & 72.55 & 77.15 & 97.96 & 81.52 & 83.87 & 80.0 & 54.6 & 76.36 \\
& TOVA      & 100.0 & 100.0 & 100.0 & 99.8 & 99.8 & 48.6 & 98.85 & 98.9 & 99.8 & 90.6 & 91.87 & 80.6 & 56.6 & 89.65 \\
\cdashline{2-16}
& SnapKV        & 100.0 & 100.0 & 72.0 & 99.6 & 100.0 & 97.8 & 98.5 & 99.15 & 99.6 & 84.0 & 90.4 & 81.8 & 57.2 & 90.77 \\
& +\think(0.5)  & 97.4 & 99.8 & 69.4 & 99.4 & 98.4 & 96.0 & 99.0 & 97.9 & 98.2 & 69.06 & 91.67 & 78.2 & 51.4 & 88.14 \\
& +\think(0.8)  & 0.0 & 0.0 & 0.0 & 0.0 & 0.0 & 0.0 & 0.05 & 0.0 & 0.0 & 1.62 & 0.07 & 10.4 & 14.6 & 2.06 \\
\hc &   +\textbf{\our}~(0.5)    & 100.0 & 99.6 & 71.8 & 98.6 & 99.8 & 97.8 & 98.65 & 99.0 & 99.68 & 84.2 & 88.27 & 81.6 & 56.8 & 90.45 \\
\hd &   +\textbf{\our}~(0.8)    & 99.0 & 95.0 & 70.0 & 98.2 & 96.8 & 94.2 & 98.65 & 98.6 & 99.24 & 75.26 & 91.27 & 80.6 & 55.2 & 88.62 \\
\cdashline{2-16}
& PyramidKV  & 100.0 & 100.0 & 48.6 & 99.8 & 100.0 & 95.2 & 99.2 & 99.1 & 99.8 & 52.88 & 90.13 & 81.0 & 57.4 & 86.39\\
& +\think(0.5)  & 98.2 & 99.6 & 47.4 & 99.6 & 98.8 & 92.4 & 99.15 & 97.8 & 98.64 & 32.9 & 92.2 & 77.4 & 52.2 & 83.56 \\
& +\think(0.8)  & 0.0 & 0.0 & 0.0 & 0.0 & 0.0 & 0.0 & 0.05 & 0.0 & 0.0 & 0.74 & 0.0 & 9.6 & 13.4 & 1.83\\
\hc &   +\textbf{\our}~(0.8)    & 99.8 & 100.0 & 47.4 & 99.6 & 99.8 & 95.6 & 99.2 & 99.05 & 99.8 & 52.8 & 90.27 & 80.8 & 57.2 & 86.26\\
\hd &   +\textbf{\our}~(0.8)    & 100.0 & 100.0 & 47.8 & 99.8 & 99.8 & 93.4 & 99.15 & 99.15 & 99.28 & 47.38 & 90.33 & 80.2 & 54.0 & 85.41\\

\midrule
\midrule
\multicolumn{16}{c}{\textbf{8K}}\\
& Vanilla    & 100.0 & 100.0 & 100.0 & 100.0 & 99.8 & 99.2 & 99.9 & 99.6 & 99.88 & 97.6 & 87.93 & 82.8 & 62.2 & 94.53\\
\midrule
\multirow{13}*{\raisebox{10.0em}{\rotatebox{90}{20\% KV cache}}} & StreamingLLM  & 18.8 & 18.8 & 20.8 & 20.2 & 18.4 & 18.0 & 18.1 & 17.25 & 33.6 & 9.68 & 82.6 & 32.4 & 45.2 & 27.22\\
& ExpectedAttention  &  98.8 & 62.0 & 0.0 & 56.2 & 66.2 & 0.4 & 16.4 & 35.5 & 63.12 & 60.22 & 68.53 & 65.6 & 54.2 & 49.78\\
& TOVA      &  100.0 & 99.8 & 93.4 & 100.0 & 96.2 & 0.4 & 99.4 & 99.4 & 99.56 & 44.32 & 67.67 & 74.8 & 56.8 & 79.37\\
\cdashline{2-16}
& SnapKV        & 100.0 & 99.2 & 2.6 & 100.0 & 97.4 & 36.0 & 96.15 & 99.6 & 94.76 & 62.04 & 70.73 & 81.2 & 61.8 & 77.04 \\
& +\think(0.5)  & 93.0 & 97.0 & 2.6 & 99.4 & 88.0 & 31.2 & 96.25 & 99.5 & 90.76 & 45.66 & 68.07 & 77.6 & 56.0 & 72.7 \\
& +\think(0.8)  & 0.0 & 0.0 & 0.0 & 0.0 & 0.0 & 0.0 & 0.0 & 0.0 & 0.0 & 0.16 & 0.0 & 26.4 & 25.4 & 4.0\\
\hc &   +\textbf{\our}~(0.5)    & 100.0 & 99.2 & 2.6 & 100.0 & 97.0 & 34.6 & 96.15 & 99.65 & 95.48 & 61.68 & 70.73 & 81.2 & 61.0 & 76.87 \\
\hd &   +\textbf{\our}~(0.8)    &  93.0 & 92.0 & 2.6 & 95.8 & 88.4 & 25.0 & 89.6 & 92.05 & 85.44 & 47.82 & 66.4 & 76.8 & 51.2 & 69.7
\\
\cdashline{2-16}
& PyramidKV  & 100.0 & 99.8 & 2.4 & 100.0 & 98.2 & 27.8 & 98.4 & 99.65 & 94.24 & 32.44 & 66.67 & 81.2 & 62.6 & 74.11\\
& +\think(0.5)  & 95.0 & 98.2 & 2.4 & 99.6 & 88.2 & 25.2 & 97.9 & 99.55 & 91.04 & 19.02 & 63.27 & 77.8 & 56.2 & 70.26 \\
& +\think(0.8)  & 0.0 & 0.0 & 0.0 & 0.0 & 0.0 & 0.0 & 0.0 & 0.0 & 0.0 & 0.16 & 0.0 & 26.2 & 24.6 & 3.92\\
\hc &   +\textbf{\our}~(0.5)    & 100.0 & 99.6 & 2.4 & 99.8 & 97.8 & 28.0 & 98.1 & 99.65 & 93.84 & 31.64 & 66.13 & 79.0 & 61.8 & 73.67 \\
\hd &   +\textbf{\our}~(0.8)    & 97.2 & 96.6 & 2.4 & 99.2 & 93.6 & 20.4 & 94.3 & 97.4 & 88.76 & 21.06 & 63.67 & 74.8 & 54.4 & 69.52\\

\midrule
\multirow{13}*{\raisebox{10.0em}{\rotatebox{90}{50\% KV cache}}} & StreamingLLM  & 47.0 & 49.4 & 55.4 & 54.4 & 52.0 & 50.0 & 51.2 & 49.65 & 71.08 & 26.5 & 85.6 & 33.4 & 53.4 & 52.23\\
& ExpectedAttention  & 99.8 & 91.2 & 10.0 & 96.8 & 95.4 & 35.4 & 73.5 & 78.8 & 94.36 & 95.2 & 82.4 & 78.6 & 59.2 & 76.2 \\
& TOVA      &  100.0 & 100.0 & 100.0 & 100.0 & 99.8 & 53.0 & 99.8 & 99.65 & 99.88 & 87.28 & 78.07 & 82.2 & 62.0 & 89.36\\
\cdashline{2-16}
& SnapKV        & 100.0 & 100.0 & 48.0 & 100.0 & 99.6 & 92.8 & 99.0 & 99.7 & 98.8 & 92.48 & 83.27 & 82.6 & 62.6 & 89.14 \\
& +\think(0.5)  & 97.0 & 99.0 & 43.2 & 99.6 & 96.2 & 89.8 & 99.15 & 99.7 & 97.28 & 86.0 & 81.27 & 78.4 & 57.8 & 86.49 \\
& +\think(0.8)  & 0.0 & 0.0 & 0.0 & 0.0 & 0.0 & 0.0 & 0.05 & 0.0 & 0.0 & 3.26 & 0.0 & 19.2 & 21.0 & 3.35\\
\hc &   +\textbf{\our}~(0.5)    & 100.0 & 100.0 & 47.8 & 100.0 & 99.6 & 93.2 & 98.8 & 99.55 & 98.92 & 92.48 & 83.07 & 82.2 & 61.4 & 89.0 \\
\hd &   +\textbf{\our}~(0.8)    & 100.0 & 99.8 & 43.0 & 100.0 & 99.2 & 85.6 & 97.6 & 99.7 & 97.56 & 88.5 & 82.27 & 79.8 & 57.8 & 86.99\\
\cdashline{2-16}
& PyramidKV  & 100.0 & 100.0 & 24.6 & 100.0 & 99.8 & 87.4 & 99.8 & 99.6 & 99.0 & 69.54 & 79.73 & 82.8 & 61.6 & 84.91\\
& +\think(0.5)  & 98.0 & 99.6 & 21.8 & 99.4 & 98.8 & 85.8 & 99.55 & 99.25 & 97.92 & 54.9 & 77.93 & 77.4 & 57.2 & 82.12 \\
& +\think(0.8)  & 0.0 & 0.0 & 0.0 & 0.0 & 0.0 & 0.0 & 0.0 & 0.0 & 0.0 & 1.24 & 0.0 & 17.4 & 20.6 & 3.02\\
\hc &   +\textbf{\our}~(0.5)    & 99.2 & 99.4 & 25.0 & 99.8 & 99.4 & 87.2 & 99.35 & 99.25 & 98.6 & 69.14 & 79.8 & 82.8 & 60.6 & 84.58 \\
\hd &   +\textbf{\our}~(0.8)    & 99.0 & 100.0 & 24.0 & 99.8 & 99.6 & 84.2 & 98.75 & 99.25 & 97.64 & 65.26 & 78.47 & 80.4 & 58.4 & 83.44\\
\bottomrule
\end{tabular}
}
\caption{RULER evaluation results on the LLaMA3.1-8B-Instruct model with \our{} under a 20\% and 50 \% KV cache budget with 8K and 16K input length.
}
\label{tab:app_full_ruler}
\end{table*}

\end{document}